\documentclass[11pt]{article}

\usepackage[margin=1in]{geometry}
\usepackage[T1]{fontenc}
\usepackage[utf8]{inputenc}
\usepackage{microtype}
\usepackage{graphicx}
\usepackage{subcaption}
\usepackage{booktabs}
\usepackage{amsmath}
\usepackage{amssymb}
\usepackage{mathtools}
\usepackage{amsthm}
\usepackage{multirow}
\usepackage{array}
\usepackage{algorithm}
\usepackage{algorithmic}
\usepackage{xcolor}
\usepackage[authoryear,round]{natbib}
\usepackage{hyperref}
\usepackage[capitalize,noabbrev]{cleveref}
\hypersetup{colorlinks=true, linkcolor=blue!70!black, citecolor=blue!70!black, urlcolor=blue!70!black}

\theoremstyle{plain}

\theoremstyle{definition}

\theoremstyle{remark}

\newcommand{\E}{\mathbb{E}}

\newcommand{\N}{\mathcal{N}}

\newcommand{\Rcal}{\mathcal{R}}

\newcommand{\RRk}[1]{\text{RR}@{#1}}
\newcommand{\DA}{\text{DA}}
\newcommand{\OP}{\text{OP}}
\newcommand{\IIS}{\text{IIS}}

\newcommand{\OPTIMAL}{\textsc{Optimal}}
\newcommand{\INFEASIBLE}{\textsc{Infeasible}}

\newcommand{\GRPO}{\textsc{GRPO}}
\newcommand{\DAPO}{\textsc{DAPO}}
\newcommand{\SFT}{\textsc{SFT}}
\newcommand{\PRM}{\textsc{PRM}}
\newcommand{\RAG}{\textsc{RAG}}

\newcommand{\ORLoopBench}{\textsc{ORLoopBench}}
\newcommand{\ORDebug}{\textsc{OR-Debug-Bench}}
\newcommand{\ORBias}{\textsc{OR-Bias-Bench}}

\newcommand{\Qwen}{\textsc{Qwen3-8B}}

\newcommand{\CR}{\text{CR}}
\newcommand{\Qopt}{Q^*}

\newcommand{\figcolwidth}{\columnwidth}
\newcommand{\figfullwidth}{\textwidth}
\newcommand{\tabcolwidth}{\columnwidth}
\newcommand{\tabfullwidth}{\textwidth}

\newcommand{\eqnref}[1]{Eq.~\eqref{eq:#1}}

\renewcommand{\figcolwidth}{0.72\textwidth}
\renewcommand{\figfullwidth}{0.86\textwidth}
\renewcommand{\tabcolwidth}{0.72\textwidth}
\renewcommand{\tabfullwidth}{0.92\textwidth}

\title{\Large \ORLoopBench{}: Solver-in-the-Loop Benchmarks for Self-Correction and Behavioral Rationality in Operations Research}

\author{
  Ruicheng Ao\thanks{E-mail: \texttt{aorc@mit.edu}, \texttt{dslevi@mit.edu}, \texttt{xinshang.w@alibaba-inc.com}.}
  \and
  David Simchi-Levi
  \and
  Xinshang Wang
}
\date{}

\begin{document}
\maketitle

\begin{abstract}
Operations Research practitioners debug infeasible models through an iterative process: inspecting Irreducible Infeasible Subsystems (\IIS{}), identifying constraint conflicts, and repairing formulations until feasibility is restored. Existing LLM benchmarks mostly treat OR as one-shot translation from problem descriptions to solver code, omitting this diagnostic loop. We formalize infeasible-model repair as a solver-in-the-loop Markov Decision Process in which each action triggers solver re-execution and \IIS{} recomputation, yielding deterministic, verifiable feedback. We introduce \textbf{\ORLoopBench{}}, a benchmark suite with two components: \textbf{\ORDebug{}} releases 5,362 LP/MILP repair instances, while \textbf{\ORBias{}} evaluates closed-form operational decision rationality across inventory settings. Solver-verified RLVR training enables an 8B model to surpass frontier APIs on LP repair (95.3\% vs 92.4\% \RRk{5}), improves diagnostic behavior, and transfers to MILP repair. The same evaluation exposes semantic drift in whole-model code regeneration: feasible regenerated MILPs can solve the wrong problem. Process-level evaluation with solver oracles enables targeted training for reliable OR self-correction.
\end{abstract}


\section{Introduction}
\label{sec:intro}

\subsection{From Translation to Debugging}

When a linear program returns \INFEASIBLE{}, the real work begins. An analyst must examine the Irreducible Infeasible Subsystem (\IIS{}, the minimal subset of constraints that cannot be simultaneously satisfied), diagnose the root cause, and systematically repair the formulation. This iterative debugging loop is where OR expertise manifests.

Yet existing benchmarks evaluate LLMs on Operations Research as one-shot translation: given a problem description, generate solver code. This paradigm ignores the debugging process central to real OR practice. Unlike generic error messages, \IIS{} provides a \textbf{minimal certificate of infeasibility}, enabling targeted, interpretable repairs. This structured feedback is what should enable targeted self-correction.

\begin{figure}[t]
    \centering
    \includegraphics[width=\figcolwidth]{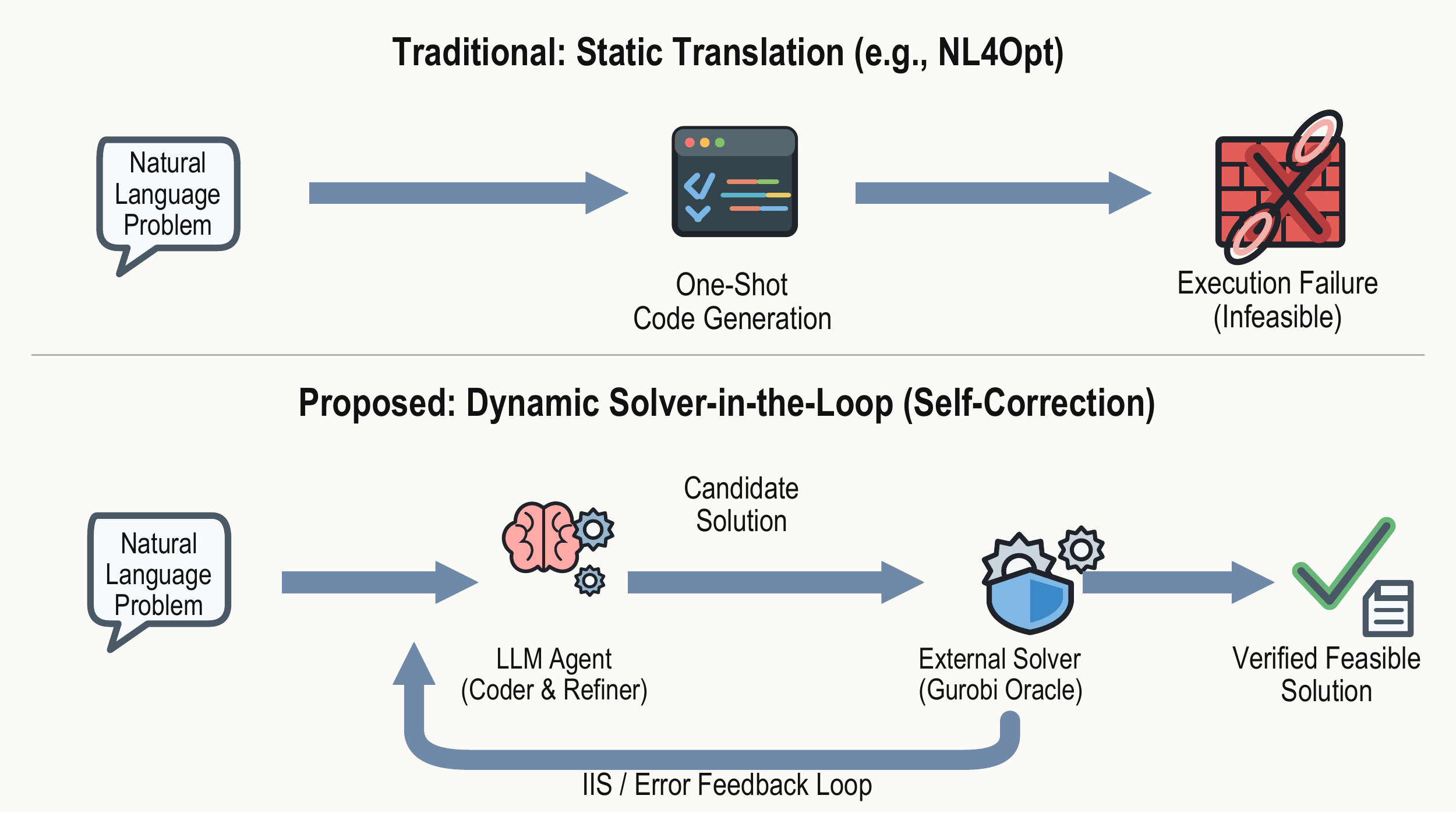}
    \caption{Evaluation paradigms compared. \textbf{Top}: Static translation benchmarks evaluate one-shot code generation with no execution feedback. \textbf{Bottom}: Our solver-in-the-loop approach enables iterative self-correction through \IIS{} feedback.}
    \label{fig:comparison}
\end{figure}

\subsection{Self-Correction in Structured Domains}

Recent work showed that 64.5\% of LLM errors result when models fail to self-correct \citep{correctbench2025}. However, CorrectBench focuses on general programming tasks and omits the OR domain, where structured self-correction is uniquely suited. First, solvers provide \textbf{deterministic feedback}: precise, verifiable signals such as \IIS{}, slack values, and objective bounds. Second, \textbf{verifiable ground truth} enables mathematical checking of optimal solutions. Third, the \textbf{interpretable process} means the diagnostic reasoning chain admits structured evaluation.

This combination (deterministic oracle, verifiable outcomes, structured process) makes OR a natural testbed for studying self-correction. The solver-in-the-loop paradigm forces agents to reason from \IIS{} feedback, enabling systematic hypothesis refinement rather than blind trial-and-error.

\subsection{Behavioral Rationality in Operations}

While debugging addresses \emph{upstream} formulation errors, operational decisions face a distinct \emph{downstream} challenge. Concurrent work on AIM-Bench \citep{aimbench2025} revealed systematic behavioral biases in LLM inventory managers: a ``pull-to-center'' tendency where models over-order when optimal quantity is low and under-order when it is high. This bias persists across model scales, raising concerns about deploying LLMs in high-stakes operations management.

\subsection{Contributions}

We make four contributions:

\begin{enumerate}
    \item \textbf{A solver-in-the-loop MDP for OR debugging}: We formalize infeasible-model repair as a sequential decision problem in which each action modifies the formulation, triggers solver re-execution, and receives updated \IIS{} feedback. This shifts evaluation from static NL-to-code translation to the diagnostic loop used in practice.

    \item \textbf{Solver-verified training for small models}: We adapt RLVR to OR repair with rewards for feasibility recovery, objective preservation, diagnostic accuracy, and faithful use of \IIS{} evidence. An 8B model trained with \GRPO{} surpasses the strongest frontier API in LP repair: 95.3\% vs 92.4\% \RRk{5}. In the core 26-model evaluation, it also improves \DA{} by +14.6 pp.

    \item \textbf{\ORLoopBench{}}: \ORLoopBench{} consists of two controlled components. \ORDebug{} releases 5,362 LP/MILP repair instances spanning LP error types A--I and MILP repair settings. \ORBias{} evaluates operational decision rationality in newsvendor and EOQ inventory settings with ID/OOD splits. The benchmark files are available at \url{https://github.com/Archer222arc/ORLoopBench}.

    \item \textbf{Benchmark findings}: \ORDebug{} evaluates iterative infeasibility repair, MILP transfer, and semantic drift in feasible-but-wrong regenerated code; \ORBias{} includes one-shot and feedback-based decision protocols. The MDP repair framework transfers to MILP, reaching 87.1\% \RRk{5} versus 71.0\% for the best API baseline.
\end{enumerate}

\begin{table}[t]
\caption{Key results summary.}
\label{tab:preview}
\centering
\small
\begin{tabular}{@{}lll@{}}
\toprule
Finding & Metric & Result \\
\midrule
8B surpasses frontier APIs & \RRk{5} & 95.3\% vs 92.4\% \\
Diagnostic accuracy gain & \DA{} & 62.4\% vs 47.8\% \\
Efficiency improvement & Steps & 2.25 vs 3.15 \\
OOD generalization & Bias drift & -9.6\% (best OOD) \\
Bias reduction & Bias diff & 20.0\%$\rightarrow$10.4\% \\
\bottomrule
\end{tabular}
\end{table}

\section{Benchmark Setup}
\label{sec:setup}

\begin{figure*}[t]
    \centering
    \includegraphics[width=\figfullwidth]{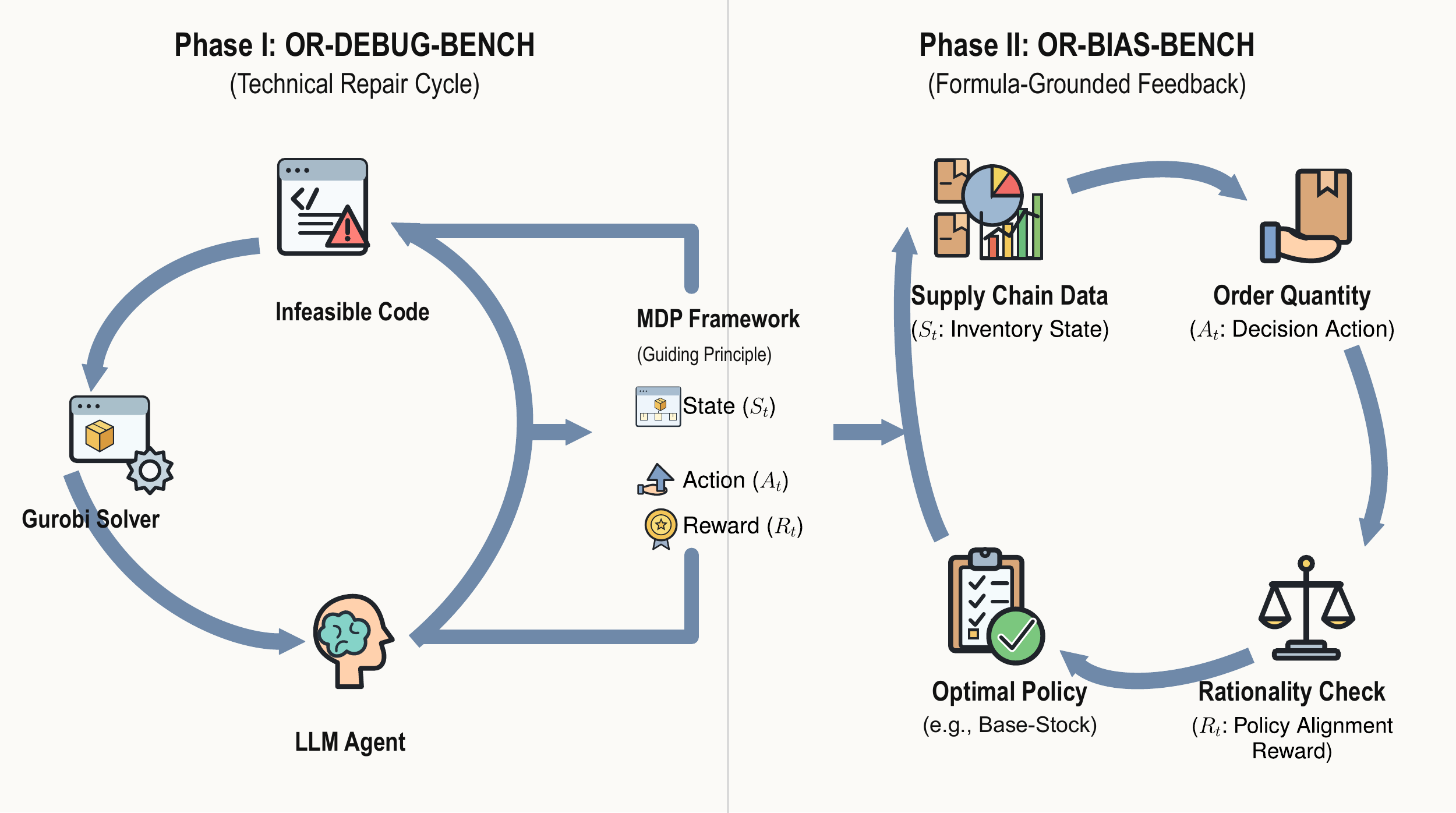}
    \caption{Two-phase benchmark framework. \textbf{Phase I (\ORDebug{})}: Iterative debugging where the agent receives Gurobi \IIS{} feedback and repairs infeasible code. \textbf{Phase II (\ORBias{})}: Inventory decision-making verified against closed-form optimal policies.}
    \label{fig:two-stage}
\end{figure*}

\subsection{\ORDebug{}: Data Organization \& Metrics}

\begin{table}[t]
\caption{\ORLoopBench{} benchmark positioning. \ORDebug{} evaluates iterative debugging with deterministic oracle feedback.}
\label{tab:positioning}
\centering
\small
\resizebox{\tabcolwidth}{!}{
\begin{tabular}{@{}lccccc@{}}
\toprule
Benchmark & Year & Task & Oracle & Multi-step & Self-Corr. \\
\midrule
NL4Opt~\citep{nl4opt2022} & 2022 & NL$\rightarrow$LP & -- & -- & -- \\
OptiBench~\citep{optibench2024} & 2024 & Formulation & -- & -- & -- \\
MAMO~\citep{mamo2024} & 2024 & Complex LP & -- & -- & -- \\
ORLM~\citep{orlm2025} & 2025 & Formulation & -- & -- & -- \\
AIM-Bench~\citep{aimbench2025} & 2025 & Inventory & Closed-form & -- & -- \\
SWE-bench~\citep{swebench2024} & 2024 & Code Debug & Unit Tests & $\checkmark$ & Limited \\
CorrectBench~\citep{correctbench2025} & 2025 & Self-Corr. & -- & -- & General \\
\midrule
\textbf{\ORDebug{}} & \textbf{2026} & \textbf{Debugging} & \textbf{Gurobi IIS} & $\checkmark$ & $\checkmark$ \\
\textbf{\ORBias{}} & \textbf{2026} & \textbf{Decision} & \textbf{Closed-form} & -- & $\checkmark$ \\
\bottomrule
\end{tabular}
}
\end{table}

\ORLoopBench{} organizes our benchmark suite into two components: \ORDebug{} for upstream solver-guided model repair and \ORBias{} for downstream closed-form operational decisions.

\begin{figure}[t]
    \centering
    \includegraphics[width=\figcolwidth]{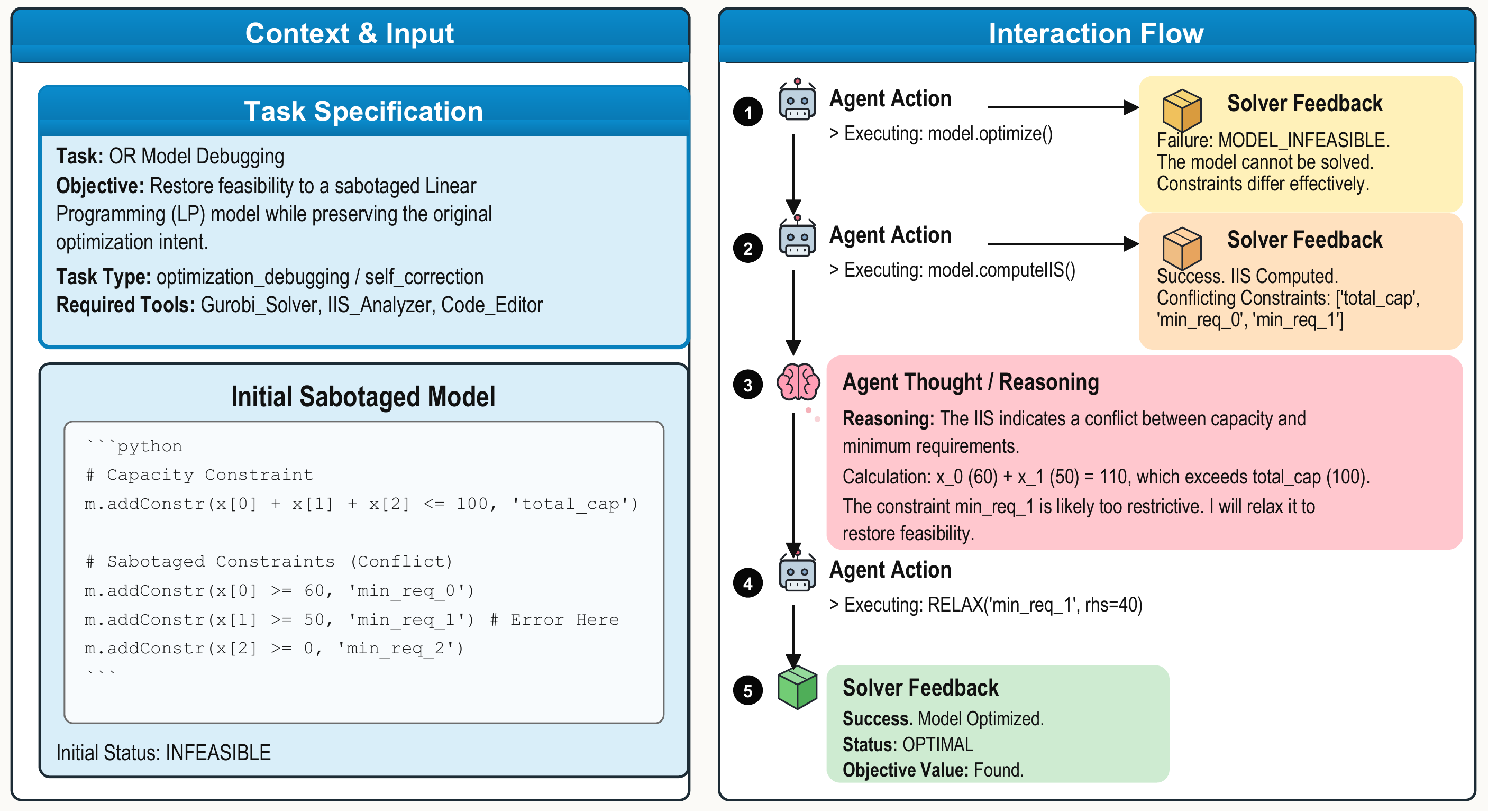}
    \caption{Example \ORDebug{} episode. \textbf{Left}: The agent receives a sabotaged LP where minimum requirements ($60+50+0=110$) exceed capacity ($100$). \textbf{Right}: The agent (1) attempts optimization, (2) computes \IIS{}, (3) reasons about the conflict, (4) relaxes the key constraint, and (5) achieves \OPTIMAL{} status in 2 repair steps.}
    \label{fig:setting-example}
\end{figure}

\ORDebug{} turns infeasibility repair into an interactive task rather than a one-shot code-generation task. Each instance provides a natural-language problem description and sabotaged Gurobi code that returns \INFEASIBLE{}. The hidden ground truth contains the original feasible code, the sabotaged constraint, and the intended repair. During evaluation, Gurobi 11.0 computes an \IIS{} after each attempted repair, giving the agent a deterministic certificate of the current conflict. Figure~\ref{fig:setting-example} illustrates a complete episode where the agent diagnoses an \IIS{} conflict and repairs the model in two steps.

\begin{table}[t]
\caption{\ORDebug{} benchmark statistics and evaluation protocols.}
\label{tab:debug_stats}
\centering
\small
\begin{tabular}{@{}ll@{}}
\toprule
Attribute & Value \\
\midrule
\multicolumn{2}{@{}l@{}}{\textit{Dataset Structure}} \\
Released Instances & 5,362 \\
Scope & LP/MILP infeasibility repair \\
LP Error Types & 9 (A--I) \\
MILP Error Types & 8 \\
\IIS{} Range & 1--11 constraints \\
Difficulty Levels & Easy / Medium / Hard \\
\midrule
\multicolumn{2}{@{}l@{}}{\textit{Reported Protocols}} \\
LP Repair & 450 instances (50 per error type) \\
MILP Repair & 10 domains $\times$ 8 error types, 5 repeats \\
\midrule
\multicolumn{2}{@{}l@{}}{\textit{Training Artifacts}} \\
Training (SFT) & 696 trajectories \\
Training (RL) & 300 prompts \\
\midrule
Max Steps per Episode & 50 \\
\bottomrule
\end{tabular}
\end{table}

\textbf{Data Organization.}
The released \ORDebug{} file contains 5,362 LP/MILP repair instances; the JSON field names in the public repository are storage labels, not separate benchmark definitions. Training uses 996 generation artifacts produced during benchmark construction. The LP repair evaluation uses 450 instances (50 per error type A--I), and the MILP repair evaluation uses repeated runs across 10 domains and 8 error types. For \ORBias{}, training uses 900 samples across three curriculum stages with \CR{} range [0.3, 0.7], while evaluation covers broader ranges: ID [0.05, 0.95] and OOD [0.10, 0.89].

\textbf{Metrics.}
We report three complementary quantities. \RRk{k} measures the percentage of instances repaired to \OPTIMAL{} within $k$ repair steps in a single solver-interaction episode. \DA{} measures whether the agent identified the true conflicting constraints, rather than reaching feasibility by accident. \OP{} measures whether the repaired model preserves the original objective value.

\subsection{\ORBias{}: Data Organization \& Metrics}

\ORBias{} evaluates whether models make operational decisions that agree with closed-form optima. It contains 2,000 newsvendor instances (1,000 ID + 1,000 OOD), where the optimal order quantity is $\Qopt = F^{-1}(\CR)$, and 300 EOQ instances, where $Q^*=\sqrt{2DK/h}$. The same evaluation framework supports both one-shot decisions and a multi-turn feedback protocol in which the model may revise its quantity after seeing formula-grounded cost feedback.

\begin{table}[t]
\caption{\ORBias{} data splits used in experiments. The benchmark contains 2,000 instances (1,000 ID + 1,000 OOD); we evaluate on stratified subsets.}
\label{tab:bias_stats}
\centering
\small
\begin{tabular}{@{}lccc@{}}
\toprule
Attribute & Train & ID Eval & OOD Eval \\
\midrule
Samples & 900 & 400 & 200 \\
\CR{} Range & [0.3, 0.7] & [0.05, 0.95] & [0.10, 0.89] \\
Demand Dist. & $\N(\mu, \sigma)$ & $\N(\mu, \sigma)$ & $\N(\mu, \sigma)$ \\
$\mu$ Range & [50, 200] & [50, 200] & [50, 200] \\
$\sigma$ Range & [10, 50] & [10, 50] & [10, 50] \\
\bottomrule
\end{tabular}
\end{table}

\textbf{Metrics.}
Rationality measures valid response percentage.
Bias Diff = $|\E[Q/\Qopt | \CR>0.5] - \E[Q/\Qopt | \CR<0.5]|$ measures pull-to-center.
ID$\rightarrow$OOD $\Delta$ measures generalization.

\begin{table}[t]
\caption{\ORLoopBench{} components. \ORDebug{} targets upstream model repair through iterative solver interaction, while \ORBias{} evaluates downstream decision-making against analytical ground truth.}
\label{tab:our_benchmarks}
\centering
\small
\resizebox{\tabcolwidth}{!}{
\begin{tabular}{@{}lcc@{}}
\toprule
Aspect & \ORDebug{} & \ORBias{} \\
\midrule
Domain & Mathematical Programming & Operations Management \\
Task & Debug infeasible code & Inventory decision \\
Oracle & Gurobi \IIS{} feedback & Closed-form $\Qopt = F^{-1}(\CR)$ \\
Interaction & Multi-step solver episode & Single-shot + feedback diagnostic \\
LLM Challenge & Error parsing, code repair & Cognitive bias mitigation \\
Key Metrics & \RRk{k}, \DA{}, \OP{} & Rationality, Bias Diff \\
\bottomrule
\end{tabular}
}
\end{table}

\subsection{Evaluation Protocol}

For \ORDebug{}, an episode alternates between model actions and solver feedback: the agent sees the current infeasible model and \IIS{}, edits or queries the formulation, and receives the updated solver status. The loop stops at \OPTIMAL{} or the step budget. For \ORBias{}, the agent outputs an order quantity $Q$ and, in the feedback protocol, may revise it after receiving closed-form cost feedback.

We evaluate 26 models on the LP \ORDebug{} repair task and report MILP repair results on random instances with repeated runs. \ORBias{} results cover the newsvendor and EOQ inventory settings across frontier APIs and local variants.

\section{Benchmark Construction}
\label{sec:construction}

\subsection{Saboteur-based Problem Generation}

We design a ``saboteur'' pipeline that injects controlled errors into valid linear programs while maintaining verifiable ground truth. Each generated error must: (1) produce a verifiable \INFEASIBLE{} status, (2) yield a non-empty \IIS{} containing the sabotaged constraint, and (3) have a unique ground-truth fix restoring \OPTIMAL{} status.

\begin{table}[t]
\caption{Error type taxonomy for \ORDebug{}.}
\label{tab:error_types}
\centering
\small
\begin{tabular}{@{}llc@{}}
\toprule
Type & Name & Difficulty \\
\midrule
A & Direction Flip & Hard \\
B & Variable Type Error & Easy \\
C & Coefficient Modification & Easy \\
D & Contradicting Constraint & Hard \\
E & Multi-Constraint Conflict & Hard \\
F & Hidden Dependency & Hard \\
G & Cascading Conflict & Hard \\
H & IIS-Incomplete & Medium \\
I & Optimal Selection & Medium \\
\bottomrule
\end{tabular}
\end{table}

\textbf{Generation Pipeline.}
The pipeline operates in four stages: (1) source selection from a feasible LP pool, (2) sabotage application using type-specific corruption, (3) verification via Gurobi \IIS{} computation, and (4) oracle labeling for evaluation. Of the initial candidate pool, 87\% pass all validation checks on first generation; the remainder require at most two iterations. Full algorithmic details appear in Appendix~\ref{app:saboteur}.

\textbf{Robust Injection.}
Reliable error injection uses adaptive methods: Type A (direction flip) uses slack-based constraint selection to improve success from 30\% to 95\%; Type C (upper bound conflict) uses a 4-tier fallback strategy achieving 72\% success. Complete algorithms appear in Appendix~\ref{app:robust_injection}.

\begin{table}[t]
\caption{Difficulty calibration for \ORDebug{}. Levels are defined by baseline API model performance.}
\label{tab:difficulty_calibration}
\centering
\small
\begin{tabular}{@{}lcc@{}}
\toprule
Level & Error Types & Baseline RR@5 \\
\midrule
Easy & B, C & $\geq$85\% \\
Medium & H, I & 70--85\% \\
Hard & A, D, E, F, G & $<$70\% \\
\bottomrule
\end{tabular}
\end{table}

\subsection{Anti-Pattern Measures}

Three mechanisms prevent pattern-matching shortcuts. First, \textbf{randomized naming} in Types G--I uses UUID-based identifiers to prevent models from exploiting semantic name patterns. Second, \textbf{hidden dependencies} (Type F) create scenarios where the \IIS{} reveals a symptom constraint while the root cause lies elsewhere. Third, \textbf{cascading conflicts} (Type G) require multi-step reasoning: fixing the primary conflict reveals a secondary one. Appendix~\ref{app:anti_gaming} provides the construction details.

\subsection{Newsvendor Problem Generation}

For \ORBias{}, we generate newsvendor scenarios with controlled critical ratios:
\begin{equation}
    \Qopt = \mu + \sigma \cdot \Phi^{-1}(\CR), \quad \CR = \frac{p-c}{p-s}
\label{eq:newsvendor}
\end{equation}
where $\Phi^{-1}$ is the standard normal inverse CDF, $\mu$ is mean demand, and $\sigma$ is standard deviation.

\textbf{Stratified Sampling.}
We stratify by \CR{} buckets: ID covers [0.05, 0.95] with 100+ samples per bucket; OOD covers [0.10, 0.89] testing intermediate-to-extreme values. The four evaluation difficulty levels are detailed in Appendix~\ref{app:bias_curriculum}.

\begin{table}[t]
\caption{Evaluation difficulty levels for \ORBias{}. Each level targets specific bias phenomena with controlled \CR{} ranges and prompt complexity.}
\label{tab:bias_curriculum}
\centering
\small
\resizebox{\tabcolwidth}{!}{
\begin{tabular}{@{}lccl@{}}
\toprule
Level & \CR{} Range & Prompt Style & Target Phenomenon \\
\midrule
L1 & $[0.4, 0.6]$ & Clean & Foundation (neutral) \\
L2 & $[0.05, 0.2) \cup (0.8, 0.95]$ & Clean & Bias trigger (extreme) \\
L3 & $[0.3, 0.7]$ & + Distractors & Robustness test \\
L4 & $[0.1, 0.9]$ & + Censored & Expert inference \\
\bottomrule
\end{tabular}
}
\end{table}

\subsection{Interaction Model}

The central methodological step is to make solver feedback part of the state transition. Instead of asking a model to regenerate code from scratch, \ORDebug{} exposes a small set of repair actions and reruns the solver after every action. The next state is therefore not sampled or judged heuristically; it is the deterministic result of Gurobi execution and \IIS{} recomputation.

\textbf{\ORDebug{} state and actions.}
The state contains the natural-language problem, current code, solver status, current \IIS{}, action history, and step index. Actions fall into three groups: diagnostic queries such as \textsc{Get\_IIS} and \textsc{Check\_Slack}; repair actions such as \textsc{Relax}, \textsc{Drop}, and \textsc{Rewrite}; and \textsc{Submit} for episode termination. Full state/action specifications appear in Appendix~\ref{app:state}.

\textbf{Reward.}
The reward combines three goals:
\begin{equation}
\Rcal = 0.5 R_{\text{outcome}} + 0.3 R_{\text{diagnosis}} + 0.2 R_{\text{efficiency}}.
\end{equation}
Outcome rewards check whether the repaired model reaches \OPTIMAL{}; diagnostic rewards check whether the agent identified the ground-truth conflicting constraints; efficiency rewards favor shorter repair sequences. This design makes success, explanation quality, and repair cost visible separately.

\textbf{\ORBias{} interaction.}
The bias task is single-step in its one-shot form: the model chooses an order quantity and the oracle compares it with the closed-form optimum in \eqnref{newsvendor}. The feedback variant repeats this decision after reporting the realized cost and optimal cost.

\textbf{Diagnostic Accuracy (\DA{}).}
We measure alignment between diagnosed constraints and ground truth:
\begin{equation}
    \DA = \frac{|\text{diagnosed} \cap \IIS_{\text{GT}}|}{|\IIS_{\text{GT}}|}
\end{equation}
DA measures coverage of the true conflict set; false-positive diagnoses and off-target edits are tracked separately through precision-style diagnosis checks, the faithfulness penalty, and \OP{}. High \RRk{5} with low \DA{} indicates ``lucky'' solutions that fix problems without understanding root causes.

\section{Training Methods}
\label{sec:training}

\begin{figure*}[t]
\centering
\includegraphics[width=0.82\textwidth]{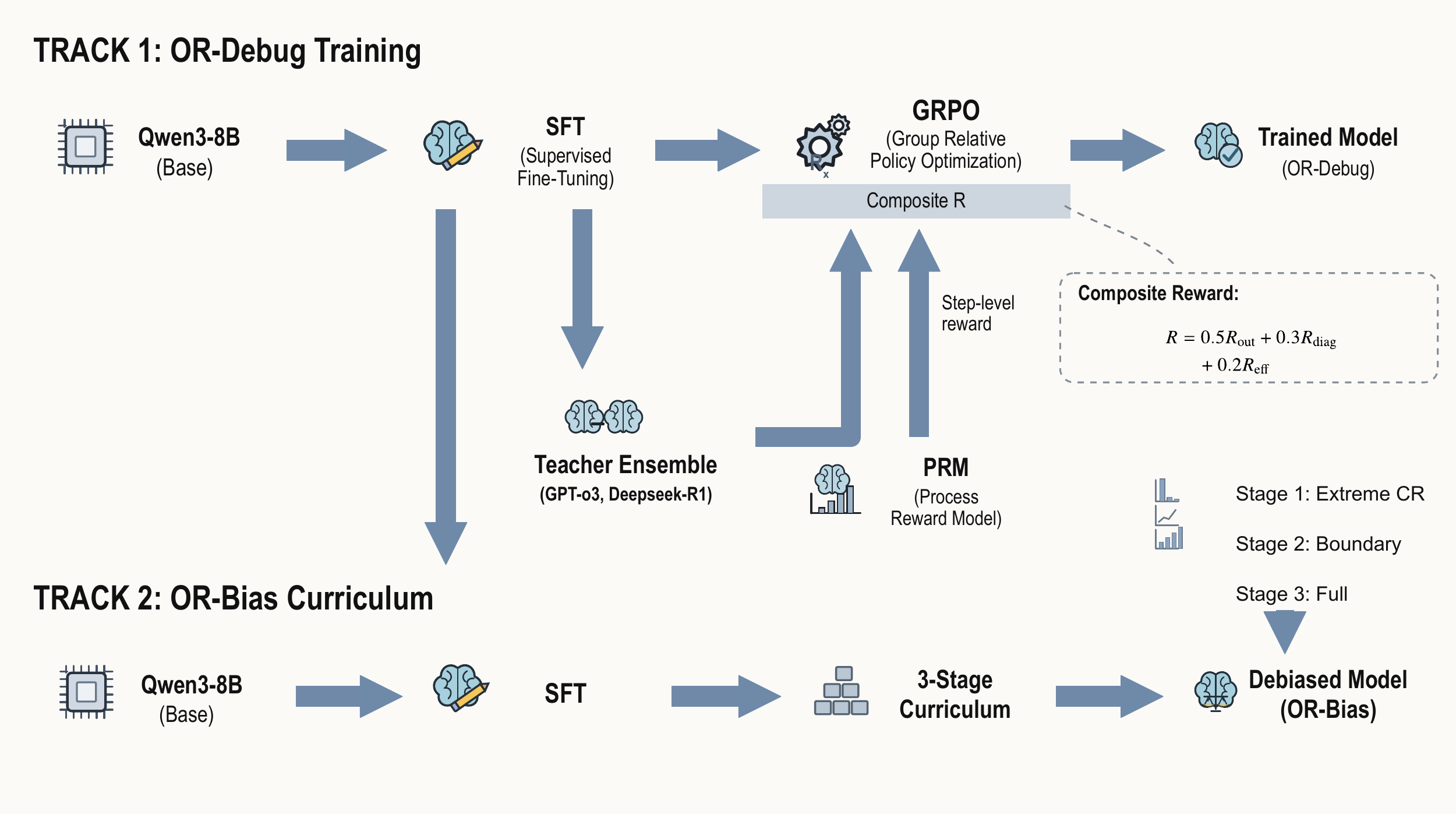}
\caption{Training pipeline overview. Track~1 trains the \ORDebug{} model: SFT on teacher trajectories followed by \GRPO{} with composite reward and optional \PRM{} supervision. Track~2 trains the \ORBias{} model: SFT on rational responses followed by a three-stage curriculum (Extreme~$\rightarrow$~Boundary~$\rightarrow$~Full) that targets pull-to-center bias.}
\label{fig:training_pipeline}
\end{figure*}

Figure~\ref{fig:training_pipeline} illustrates the two training tracks. Both start from Qwen3-8B. The debugging track first teaches the model the action format and basic \IIS{} interpretation through supervised fine-tuning, then uses solver-verified reinforcement learning to optimize repair success. The bias track uses supervised examples followed by a curriculum designed to counter pull-to-center behavior.

\subsection{Foundation Model Selection}

We selected Qwen3-8B-Instruct based on a pilot study (100 samples) showing +41.9\% post-SFT improvement headroom.

\begin{table}[t]
\caption{Pilot study: foundation model screening on \ORDebug{} validation.}
\label{tab:pilot_brief}
\centering
\small
\begin{tabular}{@{}lcccc@{}}
\toprule
Model & Base \RRk{5} & +\SFT{} \RRk{5} & $\Delta$ \\
\midrule
\textbf{Qwen3-8B} & \textbf{51.2\%} & \textbf{93.1\%} & \textbf{+41.9\%} \\
\bottomrule
\end{tabular}
\end{table}

Qwen3-8B achieved 93.1\% RR@5 after SFT with efficient token usage (2,100 tokens/episode). See Appendix~\ref{app:model_selection} for methodology.

\subsection{Supervised Trajectories}

We collect successful debugging trajectories from three teacher models (GPT-5.2-chat 40\%, o4-mini 35\%, DeepSeek-R1 25\%), chosen for diverse reasoning styles. We retain trajectories that solve the instance within five steps and diagnose at least half of the ground-truth conflict. This filtering yields 696 of 1,247 trajectories (55.8\% acceptance), averaging 2.3 steps with 68\% diagnostic accuracy. For \ORBias{}, we collect 500 rational responses. Training examples are shown in Appendix~\ref{app:training_examples}.

\subsection{\GRPO{} Training with Composite Reward}

Following DeepSeek-R1~\citep{deepseekr1}, we use Group Relative Policy Optimization with KL removal ($\beta=0$), asymmetric clipping $[0.2, 0.28]$, and LoRA ($r=16$, $\alpha=32$).

\textbf{Composite Reward.}
The reward balances outcome verification, diagnostic quality, and efficiency:
\begin{equation}
R = 0.5 \cdot R_{\text{outcome}} + 0.3 \cdot R_{\text{diagnosis}} + 0.2 \cdot R_{\text{efficiency}}
\label{eq:composite_reward}
\end{equation}
where $R_{\text{outcome}}=+100$ for \OPTIMAL{} and $-50$ otherwise, $R_{\text{diagnosis}}=\DA \cdot 100$, and $R_{\text{efficiency}}=-1$ per step. The diagnostic term addresses trial-and-error repairs that restore feasibility without identifying the root cause. A faithfulness penalty ($-20$) discourages repairs targeting non-\IIS{} constraints; without this penalty, models often achieve \OPTIMAL{} through indirect fixes that mask the root cause. Training converges after 4 epochs with \RRk{5}=95.0\%. Full training curves appear in Appendix~\ref{app:grpo_curves}.

\subsection{Process Supervision}

Outcome-based rewards provide sparse feedback: many wrong actions only reveal their cost after several solver calls. We therefore train a process reward model (\PRM{}) to score individual steps. A step receives the highest label if it reaches \OPTIMAL{} or shrinks the \IIS{}, a medium label if it identifies a ground-truth conflict, a small label for useful diagnostic actions, and zero otherwise. The \PRM{} achieves AUC-ROC of 0.94 on held-out labels and improves \DA{} by +4.7 pp over SFT (68.0\%$\rightarrow$72.7\%), while the curriculum variant achieves higher \RRk{5}. Details appear in Appendix~\ref{app:prm}.

\subsection{Curriculum Learning for Bias Mitigation}

For \ORBias{}, we use a three-stage curriculum targeting the pull-to-center bias:

\begin{table}[t]
\caption{Three-stage curriculum for \ORBias{}.}
\label{tab:curriculum}
\centering
\small
\begin{tabular}{@{}llll@{}}
\toprule
Stage & Focus & Samples & \CR{} Distribution \\
\midrule
1 & Direction Learning & 200 & Extreme (0.1, 0.9) \\
2 & Boundary Refinement & 300 & Near-boundary \\
3 & Full Distribution & 400 & [0.2, 0.8] \\
\bottomrule
\end{tabular}
\end{table}

\textbf{Stage 1 (Direction Learning)} uses extreme \CR{} values (0.1, 0.9) to teach whether quantity should move up or down. \textbf{Stage 2 (Boundary Refinement)} uses near-boundary values ([0.15, 0.25] and [0.75, 0.85]) to refine magnitude estimation. \textbf{Stage 3 (Full Distribution)} covers [0.2, 0.8] to consolidate learning.

This staged approach achieves 48\% bias reduction (20.0\%$\rightarrow$10.4\%) on OOD scenarios. Among the trained local variants, curriculum training is the only approach with improved OOD bias relative to ID ($-$9.6\% drift). Analysis appears in Appendix~\ref{app:curriculum}.

\section{Experiments}
\label{sec:experiments}

\subsection{Experimental Setup}

We evaluate 26 models: 4 local \Qwen{} variants (\SFT{}, \GRPO{}, \DAPO{}, Curriculum) and 22 API models spanning Claude, GPT, o-series, DeepSeek, Gemini, Qwen API, Llama, and kimi-k2. Experiments run on 2$\times$A100 80GB with SGLang inference (TP=2, concurrency=16).

\subsection{Main Results}

\begin{table}[t]
\caption{\ORDebug{} LP results on representative models. The test set contains 450 instances, selected as 50 held-out problems from each error type A--I. Full results for all 26 models appear in Appendix~\ref{app:full_results}.}
\label{tab:main_results}
\centering
\small
\begin{tabular}{@{}lcccc@{}}
\toprule
Model & RR & \RRk{5} & \DA{} & Steps \\
\midrule
\textbf{\Qwen{}-\GRPO{}} & 100\% & \textbf{95.3\%} & \textbf{62.4\%} & \textbf{2.25} \\
\Qwen{}-Curriculum & 100\% & 94.0\% & 61.7\% & 2.22 \\
\Qwen{}-\DAPO{} & 100\% & 93.8\% & 60.4\% & 2.31 \\
\Qwen{}-\SFT{} & 99.8\% & 93.1\% & 60.8\% & 2.34 \\
\midrule
o4-mini & 97.8\% & \textbf{86.2\%} & 47.8\% & \textbf{3.15} \\
claude-sonnet-4 & \textbf{100\%} & 86.2\% & 50.1\% & 3.71 \\
o1 & 99.8\% & 82.9\% & 47.8\% & 3.78 \\
gpt-5.2-chat & 99.8\% & 81.8\% & 40.9\% & 3.72 \\
gemini-2.5-flash & 84.2\% & 70.7\% & 19.2\% & 3.23 \\
Llama-3.3-70B & 93.8\% & 60.9\% & 46.9\% & 4.81 \\
DeepSeek-V3.2 & 99.3\% & 58.9\% & 44.8\% & 4.86 \\
DeepSeek-R1 & 99.1\% & 56.7\% & 34.5\% & 5.08 \\
\bottomrule
\end{tabular}
\end{table}

Table~\ref{tab:main_results} reports the core 26-model evaluation, including diagnostic accuracy and step efficiency. \Qwen{}-\GRPO{} reaches 95.3\% \RRk{5}, 62.4\% \DA{}, and 2.25 steps on average. Against the frontier summary in Appendix~\ref{app:frontier_summary}, the strongest LP API is Claude Sonnet 4.6 at 92.4\% \RRk{5}, so \Qwen{}-\GRPO{} remains ahead by +2.9 pp. In the core evaluation, the trained model also improves \DA{} by +14.6 pp and uses fewer repair steps than o4-mini, the step-efficient top-\RRk{5} core API baseline (2.25 vs 3.15).

The trained models use a ``diagnose once, repair correctly'' pattern: 1.3 diagnostic actions per episode vs 2.1 for API models, then targeted repairs. This reflects a different reasoning strategy: systematic elimination rather than trial-and-error.

\textbf{Per-Error-Type Performance.}
Domain-specific training provides larger gains on harder problems (A, D--G): +9.6\% average (94.4\% vs 84.8\%). Easy types (B, C) show smaller gains (+3.0\%) as baselines already exceed 95\%. Medium types (H, I) improve by +14.0\% (95.0\% vs 81.0\%). Full breakdown appears in Appendix~\ref{app:per_error}.

\textbf{Cost-Performance Trade-off.}
Local deployment avoids per-call API charges in our setup. Training cost ($\sim$8 GPU-hours on 2$\times$A100) amortizes across high-volume evaluation, and production cost depends on utilization and infrastructure.

\subsection{MILP Repair and Regeneration Checks}

Using random MILP evaluation instances across 10 domains and 8 error types with five repeated runs, the LP-trained model transfers zero-shot at 78.8\% \RRk{5}; MILP-specific training reaches 87.1\%. The best API baseline, Claude Sonnet 4.6, reaches 71.0\%. Thus, mixed-integer structure reduces absolute performance, but the MDP repair interface and solver-verified training remain effective without architectural changes.

\textbf{Semantic drift in code regeneration.}
OptiMUS-style regeneration can produce executable and feasible models that no longer encode the intended problem. On the MILP semantic-drift evaluation, GPT-5.4 reaches \OPTIMAL{} solver status in 90\% of cases, but only 28.2\% preserve the correct objective; Claude Sonnet 4.6 reaches 85\% \OPTIMAL{} but 22.4\% correct objective, and Gemini 3.1 Pro reaches 3\% \OPTIMAL{} and 0.8\% correct objective. Constraint-level repair avoids this failure mode by preserving the objective and editing only targeted constraints.

\subsection{Bias Evaluation Results}

\begin{table}[t]
\caption{\ORBias{} results (400 ID / 200 OOD samples). Bias = difference from rational ordering.}
\label{tab:bias_results}
\centering
\small
\resizebox{\tabcolwidth}{!}{
\begin{tabular}{@{}lcccc@{}}
\toprule
Model & ID Bias & OOD Bias & $\Delta$ & Status \\
\midrule
claude-haiku-4.5 & \textbf{0.0\%} & 3.6\% & +3.6\% & Best ID \\
\Qwen{}-OM-\SFT{} & 4.9\% & 11.5\% & +6.6\% & OK \\
o4-mini & 6.7\% & 7.7\% & +1.0\% & OK \\
\textbf{\Qwen{}-Curriculum} & 20.0\% & \textbf{10.4\%} & \textbf{-9.6\%} & \textbf{Best OOD} \\
gpt-4.1 & 11.8\% & 15.4\% & +3.6\% & OK \\
Llama-3.3-70B & 19.2\% & 12.5\% & -6.7\% & OK \\
gpt-5-mini & 1.2\% & 53.3\% & +52.1\% & Degraded \\
\bottomrule
\end{tabular}
}
\end{table}

Curriculum training achieves the best OOD generalization with \textbf{$-$9.6\%} drift (20.0\%$\rightarrow$10.4\%), the only trained model with substantial OOD improvement. Several API models show increased OOD bias; gpt-5-mini degrades sharply from 1.2\% to 53.3\%, while o4-mini remains comparatively stable (+1.0\%). While claude-haiku-4.5 achieves lowest ID bias (0\%), curriculum training shows \emph{improved} OOD performance ($-$9.6\% drift vs $-$6.7\% for Llama-3.3-70B).

\ORBias{} includes both newsvendor and EOQ inventory decisions. On the 300-instance EOQ setting and a five-round multi-turn protocol with closed-form error feedback, DeepSeek-R1 reduces bias from 56.9\% to 0.5\% in 2.6 rounds on average, while GPT-5.2 corrects immediately; other models remain problem-dependent, indicating that solver- or formula-grounded feedback improves some but not all operational decision biases.

\subsection{Inference Scaling Analysis}

\begin{figure}[t]
    \centering
    \includegraphics[width=\figcolwidth]{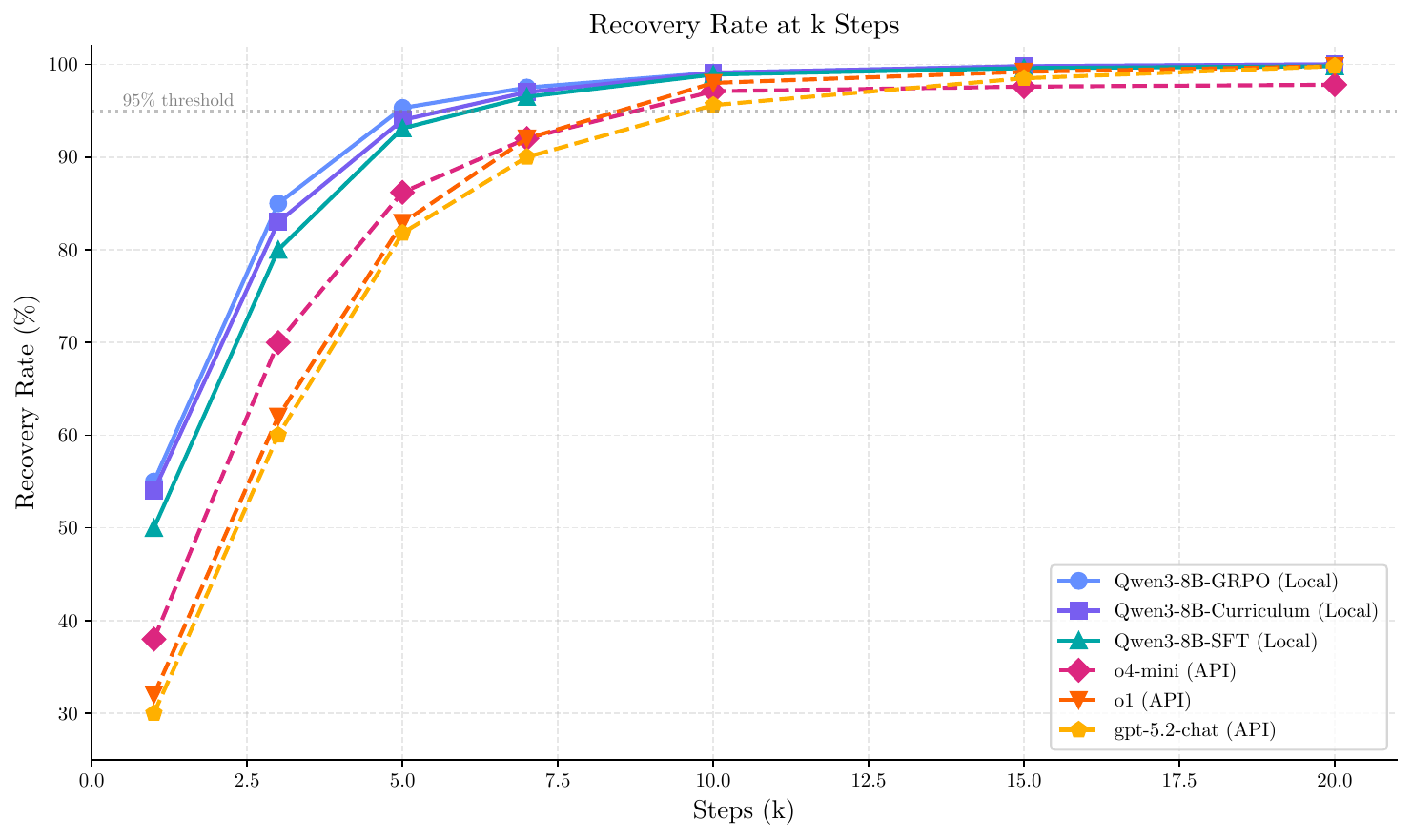}
    \caption{Recovery Rate vs. repair-step budget $k$ in the core 26-model evaluation. \Qwen{}-\GRPO{} reaches 95.3\% \RRk{5}; Appendix~\ref{app:frontier_summary} reports the frontier API summary.}
    \label{fig:rrk_curves}
\end{figure}

Figure~\ref{fig:rrk_curves} shows \RRk{k} scaling. \Qwen{}-\GRPO{} at $k=3$ (92.1\%) already surpasses o4-mini at $k=10$ (90.7\%). Token efficiency is 2.8$\times$ better: 2,109 tokens per success vs 5,976 for o4-mini. Hard problems show steeper scaling: +26.8\% from $k=1$ to $k=5$ vs +9.8\% for Easy problems. Full analysis appears in Appendix~\ref{app:scaling}.

\subsection{Ablation Study}

\begin{table}[t]
\caption{Ablation study on \ORDebug{} (200-sample validation set).}
\label{tab:ablation_summary}
\centering
\small
\begin{tabular}{@{}lccc@{}}
\toprule
Configuration & \RRk{5} & \DA{} & $\Delta$\RRk{5} \\
\midrule
\SFT{} Baseline & 91.5\% & 68.0\% & -- \\
+ \GRPO{} & 92.0\% & 66.0\% & +0.5\% \\
+ Curriculum & 95.0\% & 68.0\% & +3.5\% \\
+ \PRM{} & 92.0\% & 72.7\% & +0.5\% \\
Curriculum + \GRPO{} (best) & \textbf{95.3\%} & 62.4\% & \textbf{+3.8\%} \\
\bottomrule
\end{tabular}
\end{table}

Three ablation results stand out: (1) Curriculum pre-training provides +3.5\% \RRk{5}, the largest single improvement; (2) relative to SFT, \PRM{} raises \DA{} from 68.0\% to 72.7\% with similar \RRk{5} (91.5\% to 92.0\%), while relative to curriculum it prioritizes \DA{} over \RRk{5}; (3) Curriculum synergizes with \GRPO{} (+3.8\% over SFT), providing favorable initialization for hard problems.

\section{Related Work}
\label{sec:related}

\textbf{OR Benchmarks and Formulation Systems.}
Table~\ref{tab:positioning} compares \ORLoopBench{} with existing OR benchmarks. NL4Opt~\citep{nl4opt2022} initiated natural-language-to-LP evaluation; later work expanded prompt-based and learning-based formulation systems~\citep{chainofexperts2024,optimus2024,optibench2024,orlm2025,sirl2025,mind2026}. Related benchmarks and agents cover OR reasoning, LP formulation, workflow execution, and dynamic programming~\citep{llmopt2024,mamo2024,orqa2025,pilotbench2026,dpbench2025}. These efforts primarily evaluate static, one-shot formulation. \ORDebug{} instead starts from an already-built infeasible model, exposes live \IIS{} feedback after each repair, and evaluates iterative constraint-level correction under objective-preservation checks. A broader comparison appears in Appendix~\ref{app:related_full}.

\textbf{Formulation Training vs Debugging.}
Recent systems use solvers, search, or localized data construction to improve formulation quality~\citep{sirl2025,autoformulation2025,optitree2025,mind2026}. OptiChat studies natural-language interaction with optimization models, including infeasibility diagnosis through solver tools~\citep{optichat2024,optichat2025}. A supply-chain-focused predecessor, OptiRepair~\citep{ao2026optirepair}, studies closed-loop diagnosis and repair for multi-echelon inventory models. These works are close in spirit but differ in scope: \ORDebug{} provides a controlled benchmark, MDP action interface, objective-preservation checks, and solver-verified training/evaluation for iterative constraint-level repair. This distinction matters because one-shot formulation, interactive diagnosis, and localized repair expose different failure modes and require different evaluation protocols.

\textbf{Self-Correction and Verifiable Rewards.}
General self-correction work studies iterative refinement, reasoning bootstrapping, code debugging, and RL-trained correction~\citep{selfrefine2023,star2022,swebench2024,correctbench2025,rlselfcorrect2024}. These settings typically rely on self-generated feedback or sampled tests. In contrast, solver feedback is deterministic: \IIS{} computation identifies infeasibility certificates and supports automatically checkable progress signals. Our \GRPO{} and \PRM{} training build on RLVR and process-supervision methods~\citep{ppo2017,instructgpt2022,dpo2023,grpo2024,deepseekr1,lightman2024let,mathshepherd2024,biprm2025,pavs2025}, but adapt them to OR debugging where each repair step can be verified by the solver.

\textbf{Behavioral Rationality in LLM Decisions.}
Recent OM work examines LLMs and predictive models as decision inputs in demand simulation, pricing, service-system selection, and inventory decisions~\citep{zhang2025predicting,ao2026pricinginfo,ao2026servicesystems,aimbench2025}. \ORBias{} builds on the pull-to-center phenomenon documented in LLM inventory choices and classical human newsvendor bias~\citep{schweitzer2000decision,kremer2010demand}. It adds explicit ID/OOD splits and formula-grounded feedback protocols, showing that curriculum learning can reduce OOD decision bias rather than only improve in-distribution fit.

\section{Discussion}
\label{sec:discussion}

\ORLoopBench{} is organized around two complementary failure modes. \ORDebug{} targets upstream formulation repair: models must act on Gurobi \IIS{} feedback and reason about constraint conflicts, where API models rely more on trial-and-error. \ORBias{} targets downstream decision quality: models must produce closed-form inventory decisions that remain rational under ID/OOD shifts, where gpt-5-mini's 1.2\% ID bias and 53.3\% OOD bias reveal brittle heuristics. The training response differs accordingly: \GRPO{} improves \ORDebug{} through solver-verified rewards, while curriculum training reduces \ORBias{} OOD bias by staging exposure to extreme \CR{} values.

Three design choices prevent shortcut solutions. \OP{} penalizes feasible repairs that alter the objective; the faithfulness penalty discourages edits outside the current \IIS{}; and \DA{} separates root-cause diagnosis from lucky feasibility restoration. These checks matter most in MILP code regeneration, where a model can produce executable code that solves to \OPTIMAL{} while changing objective coefficients or constraint semantics. Constraint-level repair narrows the action space and preserves the original objective.

Standard prompting remains insufficient because OR debugging requires procedural knowledge: iterative solver interaction, domain-specific diagnostics, and state-dependent repair strategy. Zero-Shot CoT reaches 23.0\% \RRk{5}; three-shot ICL reaches 38.7\%, still 54 points below SFT. Appendix~\ref{app:prompting_baselines} analyzes these prompting baselines.

The remaining failures point to the limits of the current interface. Type H--I failures often involve large \IIS{} sets where one repair exposes another conflict; false positives occur when constraints play similar semantic roles; and repair-magnitude errors suggest hybrid approaches in which the model identifies the faulty constraint and an optimizer computes the smallest valid adjustment.

\textbf{Limitations.} \ORDebug{} covers infeasible LP/MILP repair with Gurobi-verifiable feedback; nonlinear, stochastic, robust, and multi-objective formulations require different certificates. \IIS{} is a minimal infeasible subset rather than a complete causal explanation, and multiple \IIS{} sets may exist. \ORBias{} covers closed-form newsvendor and EOQ decisions; its multi-turn protocol is an upper-bound diagnostic, not a full production simulation. Deployment remains human-in-the-loop and requires audit logs, data governance, solver/version reproducibility, and practitioner validation.

\section{Conclusion}
\label{sec:conclusion}

We introduced a solver-in-the-loop MDP for OR model repair and \ORLoopBench{}, a benchmark suite consisting of \ORDebug{} and \ORBias{} that evaluates iterative self-correction and behavioral rationality rather than one-shot formulation. Experiments on 12,000+ samples across 26 models show that solver-verified training enables 8B Qwen models to reach 95.3\% \RRk{5} on LP repair, ahead of the strongest frontier API summary result (92.4\%), while also improving \DA{} by +14.6 pp and using fewer repair steps than o4-mini, the step-efficient top-\RRk{5} core API baseline (2.25 vs 3.15). The same evaluation separates diagnostic correctness from trial-and-error success, shows transfer beyond LP through MILP repair (87.1\% \RRk{5} with MILP-specific training and 78.8\% zero-shot transfer), and demonstrates that localized constraint repair avoids semantic drift in feasible-but-wrong regenerated code. On the decision side, curriculum learning is the only approach showing improved OOD performance ($-$9.6\% bias drift).

Although our tasks are controlled, the \textit{Action $\to$ Feedback $\to$ Plan Update} loop captures a core requirement for reliable agent deployment. Domain-specific training within deterministic verification loops offers a concrete path toward automated decision-making in operations. Natural extensions include broader MINLP and stochastic debugging, multi-period operations, \RAG{} integration with OR knowledge bases, and practitioner-in-the-loop validation.

\section*{Impact Statement}

This paper presents work whose goal is to advance the field of Machine Learning. There are many potential societal consequences of our work, none of which we feel must be specifically highlighted here.

\bibliographystyle{plainnat}
\bibliography{references}

\begin{thebibliography}{54}
\providecommand{\natexlab}[1]{#1}
\providecommand{\url}[1]{\texttt{#1}}
\expandafter\ifx\csname urlstyle\endcsname\relax
  \providecommand{\doi}[1]{doi: #1}\else
  \providecommand{\doi}{doi: \begingroup \urlstyle{rm}\Url}\fi

\bibitem[AhmadiTeshnizi et~al.(2024)AhmadiTeshnizi, Gao, and
  Udell]{optimus2024}
Ali AhmadiTeshnizi, Wenzhi Gao, and Madeleine Udell.
\newblock {Optimus}: Scalable optimization modeling with {(MI)LP} solvers and
  large language models.
\newblock \emph{arXiv preprint arXiv:2402.10172}, 2024.

\bibitem[Ao et~al.(2026{\natexlab{a}})Ao, Chen, Gao, Li, and
  Simchi-Levi]{ao2026servicesystems}
Ruicheng Ao, Hongyu Chen, Siyang Gao, Hanwei Li, and David Simchi-Levi.
\newblock Designing service systems from textual evidence.
\newblock \emph{arXiv preprint arXiv:2603.10400}, 2026{\natexlab{a}}.
\newblock \doi{10.48550/arXiv.2603.10400}.

\bibitem[Ao et~al.(2026{\natexlab{b}})Ao, Gao, and
  Simchi-Levi]{ao2026multiagentplanning}
Ruicheng Ao, Siyang Gao, and David Simchi-Levi.
\newblock On the reliability limits of {LLM}-based multi-agent planning.
\newblock \emph{arXiv preprint arXiv:2603.26993}, 2026{\natexlab{b}}.
\newblock \doi{10.48550/arXiv.2603.26993}.

\bibitem[Ao et~al.(2026{\natexlab{c}})Ao, Jiang, and
  Simchi-Levi]{ao2026pricinginfo}
Ruicheng Ao, Jiashuo Jiang, and David Simchi-Levi.
\newblock The value of information in resource-constrained pricing.
\newblock \emph{arXiv preprint arXiv:2603.24974}, 2026{\natexlab{c}}.
\newblock \doi{10.48550/arXiv.2603.24974}.

\bibitem[Ao et~al.(2026{\natexlab{d}})Ao, Luo, Simchi-Levi, and
  Wang]{ao2026llminference}
Ruicheng Ao, Gan Luo, David Simchi-Levi, and Xinshang Wang.
\newblock Optimizing {LLM} inference: Fluid-guided online scheduling with
  memory constraints.
\newblock \emph{arXiv preprint arXiv:2504.11320}, 2026{\natexlab{d}}.

\bibitem[Ao et~al.(2026{\natexlab{e}})Ao, Min, Zhu, Yin, and
  Wang]{pilotbench2026}
Ruicheng Ao, Zeping Min, Tingyu Zhu, Wotao Yin, and Xinshang Wang.
\newblock {PILOT-Bench}: Probabilistic interaction for {LLM} operations in
  tool-driven scenarios.
\newblock In \emph{International Conference on Learning Representations},
  2026{\natexlab{e}}.
\newblock URL \url{https://openreview.net/forum?id=KTZ56LG7jZ}.

\bibitem[Ao et~al.(2026{\natexlab{f}})Ao, Simchi-Levi, and
  Wang]{ao2026optirepair}
Ruicheng Ao, David Simchi-Levi, and Xinshang Wang.
\newblock {OptiRepair}: Closed-loop diagnosis and repair of supply chain
  optimization models with {LLM} agents.
\newblock \emph{arXiv preprint arXiv:2602.19439}, 2026{\natexlab{f}}.
\newblock \doi{10.48550/arXiv.2602.19439}.

\bibitem[Astorga et~al.(2025)Astorga, Liu, Xiao, and van~der
  Schaar]{autoformulation2025}
Nicolas Astorga, Tennison Liu, Yuanzhang Xiao, and Mihaela van~der Schaar.
\newblock Autoformulation of mathematical optimization models using {LLMs}.
\newblock In \emph{International Conference on Machine Learning}, 2025.
\newblock arXiv:2411.01679.

\bibitem[Bertsimas and Margaritis(2024)]{bertsimas2024robust}
Dimitris Bertsimas and Georgios Margaritis.
\newblock Robust and adaptive optimization under a large language model lens.
\newblock \emph{arXiv preprint arXiv:2501.00568}, 2024.

\bibitem[Chen et~al.(2024)Chen, Constante-Flores, and Li]{optichat2024}
Hao Chen, Gonzalo~E. Constante-Flores, and Can Li.
\newblock Diagnosing infeasible optimization problems using large language
  models.
\newblock \emph{INFOR: Information Systems and Operational Research},
  62\penalty0 (4):\penalty0 573--587, 2024.
\newblock \doi{10.1080/03155986.2024.2385189}.

\bibitem[Chen et~al.(2025{\natexlab{a}})Chen, Constante-Flores, Mantri,
  Kompalli, Ahluwalia, and Li]{optichat2025}
Hao Chen, Gonzalo~Esteban Constante-Flores, Krishna Sri~Ipsit Mantri,
  Sai~Madhukiran Kompalli, Akshdeep~Singh Ahluwalia, and Can Li.
\newblock {OptiChat}: Bridging optimization models and practitioners with large
  language models.
\newblock \emph{arXiv preprint arXiv:2501.08406}, 2025{\natexlab{a}}.

\bibitem[Chen et~al.(2021)Chen, Tworek, Jun, Yuan, Pinto, Kaplan, Edwards,
  Burda, Joseph, Brockman, et~al.]{chen2021humaneval}
Mark Chen, Jerry Tworek, Heewoo Jun, Qiming Yuan, Henrique Ponde de~Oliveira
  Pinto, Jared Kaplan, Harri Edwards, Yuri Burda, Nicholas Joseph, Greg
  Brockman, et~al.
\newblock Evaluating large language models trained on code.
\newblock \emph{arXiv preprint arXiv:2107.03374}, 2021.

\bibitem[Chen et~al.(2025{\natexlab{b}})Chen, Xia, Shao, Ge, and Ye]{sirl2025}
Yitian Chen, Jingfan Xia, Siyu Shao, Dongdong Ge, and Yinyu Ye.
\newblock Solver-informed {RL}: Grounding large language models for authentic
  optimization modeling.
\newblock \emph{arXiv preprint arXiv:2505.11792}, 2025{\natexlab{b}}.

\bibitem[Chinneck(2008)]{chinneck2008}
John~W. Chinneck.
\newblock \emph{Feasibility and Infeasibility in Optimization: Algorithms and
  Computational Methods}, volume 118 of \emph{International Series in
  Operations Research \& Management Science}.
\newblock Springer, New York, NY, 2008.
\newblock \doi{10.1007/978-0-387-74932-7}.

\bibitem[Chu et~al.(2025)Chu, Zhai, Yang, Tong, Xie, Schuurmans, Le, Levine,
  and Ma]{sftrl2025}
Tianzhe Chu, Yuexiang Zhai, Jihan Yang, Shengbang Tong, Saining Xie, Dale
  Schuurmans, Quoc~V Le, Sergey Levine, and Yi~Ma.
\newblock {SFT} memorizes, {RL} generalizes: A comparative study of foundation
  model post-training.
\newblock \emph{arXiv preprint arXiv:2501.17161}, 2025.

\bibitem[{DeepSeek-AI}(2025)]{deepseekr1}
{DeepSeek-AI}.
\newblock {DeepSeek-R1}: Incentivizing reasoning capability in {LLMs} via
  reinforcement learning.
\newblock \emph{arXiv preprint arXiv:2501.12948}, 2025.

\bibitem[Ding et~al.(2025)Ding, Tan, Zhang, and Chen]{orr12025}
Zezhen Ding, Zhen Tan, Jiheng Zhang, and Tianlong Chen.
\newblock {OR-R1}: Automating modeling and solving of operations research
  optimization problem via test-time reinforcement learning.
\newblock \emph{arXiv preprint arXiv:2511.09092}, 2025.

\bibitem[Greenberg(1983)]{greenberg1983}
Harvey~J. Greenberg.
\newblock A functional description of {ANALYZE}: A computer-assisted analysis
  system for linear programming models.
\newblock \emph{ACM Transactions on Mathematical Software}, 9\penalty0
  (1):\penalty0 18--56, 1983.
\newblock \doi{10.1145/356022.356024}.

\bibitem[{Gurobi Optimization, LLC}(2024)]{gurobi2024}
{Gurobi Optimization, LLC}.
\newblock \emph{Gurobi Optimizer Reference Manual}, 2024.
\newblock URL \url{https://www.gurobi.com/documentation/}.

\bibitem[Huang et~al.(2025{\natexlab{a}})]{orlm2025}
Chenyu Huang et~al.
\newblock {ORLM}: A customizable framework in training large models for
  automated optimization modeling.
\newblock \emph{Operations Research}, 2025{\natexlab{a}}.

\bibitem[Huang et~al.(2025{\natexlab{b}})Huang, Shen, Hu, Gao, and
  Wang]{mamo2024}
Xuhan Huang, Qingning Shen, Yan Hu, Anningzhe Gao, and Benyou Wang.
\newblock {LLM}s for mathematical modeling: Towards bridging the gap between
  natural and mathematical languages.
\newblock In \emph{Findings of the Association for Computational Linguistics:
  NAACL 2025}, pages 2678--2710, Albuquerque, New Mexico, 2025{\natexlab{b}}.
  Association for Computational Linguistics.
\newblock \doi{10.18653/v1/2025.findings-naacl.146}.
\newblock URL \url{https://aclanthology.org/2025.findings-naacl.146/}.
\newblock Introduces the MAMO benchmark; arXiv:2405.13144.

\bibitem[Jiang et~al.(2024)Jiang, Shu, Qian, Lu, Zhou, Zhou, and
  Yu]{llmopt2024}
Caigao Jiang, Xiang Shu, Hong Qian, Xingyu Lu, Jun Zhou, Aimin Zhou, and Yang
  Yu.
\newblock {LLMOPT}: Learning to define and solve general optimization problems
  from scratch.
\newblock \emph{arXiv preprint arXiv:2410.13213}, 2024.

\bibitem[Jimenez et~al.(2024)Jimenez, Yang, Wettig, Yao, Pei, Press, and
  Narasimhan]{swebench2024}
Carlos~E Jimenez, John Yang, Alexander Wettig, Shunyu Yao, Kexin Pei, Ofir
  Press, and Karthik Narasimhan.
\newblock {SWE-bench}: Can language models resolve real-world {GitHub} issues?
\newblock In \emph{International Conference on Learning Representations}, 2024.
\newblock arXiv:2310.06770.

\bibitem[Kremer et~al.(2010)Kremer, Minner, and
  Van~Wassenhove]{kremer2010demand}
Mirko Kremer, Stefan Minner, and Luk~N Van~Wassenhove.
\newblock Do random errors explain newsvendor behavior?
\newblock \emph{Manufacturing \& Service Operations Management}, 12\penalty0
  (4):\penalty0 673--681, 2010.
\newblock \doi{10.1287/msom.1100.0294}.

\bibitem[Kumar et~al.(2024)Kumar, Zhuang, Agarwal, Su, Co-Reyes, Singh, Baumli,
  Iqbal, Bishop, Roelofs, et~al.]{rlselfcorrect2024}
Aviral Kumar, Vincent Zhuang, Rishabh Agarwal, Yi~Su, John~D Co-Reyes, Avi
  Singh, Kate Baumli, Shariq Iqbal, Colton Bishop, Rebecca Roelofs, et~al.
\newblock Training language models to self-correct via reinforcement learning.
\newblock \emph{arXiv preprint arXiv:2409.12917}, 2024.

\bibitem[Lambert et~al.(2025)Lambert, Morrison, Pyatkin, Huang, Ivison,
  Brahman, Miranda, Liu, Dziri, Lyu, Gu, Malik, Graf, Hwang, Yang, Le~Bras,
  Tafjord, Wilhelm, Soldaini, Smith, Wang, Dasigi, and Hajishirzi]{tulu3rlvr}
Nathan Lambert, Jacob Morrison, Valentina Pyatkin, Shengyi Huang, Hamish
  Ivison, Faeze Brahman, Lester James~V. Miranda, Alisa Liu, Nouha Dziri, Shane
  Lyu, Yuling Gu, Saumya Malik, Victoria Graf, Jena~D. Hwang, Jiangjiang Yang,
  Ronan Le~Bras, Oyvind Tafjord, Chris Wilhelm, Luca Soldaini, Noah~A. Smith,
  Yizhong Wang, Pradeep Dasigi, and Hannaneh Hajishirzi.
\newblock {T\"ulu} 3: Pushing frontiers in open language model post-training.
\newblock \emph{arXiv preprint arXiv:2411.15124}, 2025.

\bibitem[Li et~al.(2025)Li, Ping, Chen, Qi, Wang, Luo, and Zhang]{agentgit2025}
Yang Li, Siqi Ping, Xiyu Chen, Xiaojian Qi, Zigan Wang, Ye~Luo, and Xiaowei
  Zhang.
\newblock {AgentGit}: A version control framework for reliable and scalable
  {LLM}-powered multi-agent systems.
\newblock \emph{arXiv preprint arXiv:2511.00628}, 2025.

\bibitem[Li et~al.(2022)Li, Choi, Chung, Kushman, Schrittwieser, Leblond,
  Eccles, Keeling, Gimeno, Dal~Lago, et~al.]{alphacode2022}
Yujia Li, David Choi, Junyoung Chung, Nate Kushman, Julian Schrittwieser,
  R{\'e}mi Leblond, Tom Eccles, James Keeling, Felix Gimeno, Agustin Dal~Lago,
  et~al.
\newblock Competition-level code generation with {AlphaCode}.
\newblock \emph{Science}, 378\penalty0 (6624):\penalty0 1092--1097, 2022.

\bibitem[Lightman et~al.(2024)Lightman, Kosaraju, Burda, Edwards, Baker, Lee,
  Leike, Schulman, Sutskever, and Cobbe]{lightman2024let}
Hunter Lightman, Vineet Kosaraju, Yuri Burda, Harrison Edwards, Bowen Baker,
  Teddy Lee, Jan Leike, John Schulman, Ilya Sutskever, and Karl Cobbe.
\newblock Let's verify step by step.
\newblock In \emph{International Conference on Learning Representations}, 2024.
\newblock arXiv:2305.20050.

\bibitem[Liu et~al.(2025)Liu, Wang, Cai, Han, Kuang, and Hao]{optitree2025}
Haoyang Liu, Jie Wang, Yuyang Cai, Xiongwei Han, Yufei Kuang, and Jianye Hao.
\newblock {OptiTree}: Hierarchical thoughts generation with tree search for
  {LLM} optimization modeling.
\newblock In \emph{Advances in Neural Information Processing Systems}, 2025.
\newblock arXiv:2510.22192.

\bibitem[Liu et~al.(2026)Liu, Wu, Kuang, Han, Zhong, Feng, and Lu]{mind2026}
Weiting Liu, Han Wu, Yufei Kuang, Xiongwei Han, Tao Zhong, Jianfeng Feng, and
  Wenlian Lu.
\newblock Automated optimization modeling via a localizable error-driven
  perspective.
\newblock \emph{arXiv preprint arXiv:2602.11164}, 2026.

\bibitem[Lu et~al.(2025)Lu, Xie, Wu, Ren, Chen, and Wen]{optmath2025}
Hongliang Lu, Zhonglin Xie, Yaoyu Wu, Can Ren, Yuxuan Chen, and Zaiwen Wen.
\newblock {OptMATH}: A scalable bidirectional data synthesis framework for
  optimization modeling.
\newblock \emph{arXiv preprint arXiv:2502.11102}, 2025.

\bibitem[Madaan et~al.(2023)Madaan, Tandon, Gupta, Hallinan, Gao, Wiegreffe,
  Alon, Dziri, Prabhumoye, Yang, Gupta, Majumder, Hermann, Welleck,
  Yazdanbakhsh, and Clark]{selfrefine2023}
Aman Madaan, Niket Tandon, Prakhar Gupta, Skyler Hallinan, Luyu Gao, Sarah
  Wiegreffe, Uri Alon, Nouha Dziri, Shrimai Prabhumoye, Yiming Yang, Shashank
  Gupta, Bodhisattwa~Prasad Majumder, Katherine Hermann, Sean Welleck, Amir
  Yazdanbakhsh, and Peter Clark.
\newblock Self-refine: Iterative refinement with self-feedback.
\newblock In \emph{Advances in Neural Information Processing Systems},
  volume~36, 2023.
\newblock arXiv:2303.17651.

\bibitem[Mostajabdaveh et~al.(2025)Mostajabdaveh, Yu, Dash, Ramamonjison,
  Byusa, Carenini, Zhou, and Zhang]{orqa2025}
Mahdi Mostajabdaveh, Timothy Tin~Long Yu, Samarendra Chandan~Bindu Dash, Rindra
  Ramamonjison, Jabo~Serge Byusa, Giuseppe Carenini, Zirui Zhou, and Yong
  Zhang.
\newblock Evaluating {LLM} reasoning in the operations research domain with
  {ORQA}.
\newblock In \emph{Proceedings of the AAAI Conference on Artificial
  Intelligence}, volume~39, pages 24902--24910, 2025.
\newblock \doi{10.1609/aaai.v39i23.34673}.
\newblock URL \url{https://doi.org/10.1609/aaai.v39i23.34673}.

\bibitem[Ouyang et~al.(2022)Ouyang, Wu, Jiang, Almeida, Wainwright, Mishkin,
  Zhang, Agarwal, Slama, Ray, et~al.]{instructgpt2022}
Long Ouyang, Jeffrey Wu, Xu~Jiang, Diogo Almeida, Carroll Wainwright, Pamela
  Mishkin, Chong Zhang, Sandhini Agarwal, Katarina Slama, Alex Ray, et~al.
\newblock Training language models to follow instructions with human feedback.
\newblock \emph{Advances in Neural Information Processing Systems},
  35:\penalty0 27730--27744, 2022.

\bibitem[Rafailov et~al.(2023)Rafailov, Sharma, Mitchell, Ermon, Manning, and
  Finn]{dpo2023}
Rafael Rafailov, Archit Sharma, Eric Mitchell, Stefano Ermon, Christopher~D
  Manning, and Chelsea Finn.
\newblock Direct preference optimization: Your language model is secretly a
  reward model.
\newblock \emph{arXiv preprint arXiv:2305.18290}, 2023.

\bibitem[Ramamonjison et~al.(2023)Ramamonjison, Yu, Li, Li, Carenini, Ghaddar,
  He, Mostajabdaveh, Banitalebi-Dehkordi, Zhou, and Zhang]{nl4opt2022}
Rindranirina Ramamonjison, Timothy Yu, Raymond Li, Haley Li, Giuseppe Carenini,
  Bissan Ghaddar, Shiqi He, Mahdi Mostajabdaveh, Amin Banitalebi-Dehkordi,
  Zirui Zhou, and Yong Zhang.
\newblock {NL4Opt} competition: Formulating optimization problems based on
  their natural language descriptions.
\newblock In \emph{Proceedings of the NeurIPS 2022 Competitions Track}, volume
  220 of \emph{Proceedings of Machine Learning Research}, pages 189--203. PMLR,
  2023.
\newblock arXiv:2303.08233.

\bibitem[Schulman et~al.(2017)Schulman, Wolski, Dhariwal, Radford, and
  Klimov]{ppo2017}
John Schulman, Filip Wolski, Prafulla Dhariwal, Alec Radford, and Oleg Klimov.
\newblock Proximal policy optimization algorithms.
\newblock \emph{arXiv preprint arXiv:1707.06347}, 2017.

\bibitem[Schweitzer and Cachon(2000)]{schweitzer2000decision}
Maurice~E Schweitzer and G{\'e}rard~P Cachon.
\newblock Decision bias in the newsvendor problem with a known demand
  distribution: Experimental evidence.
\newblock \emph{Management Science}, 46\penalty0 (3):\penalty0 404--420, 2000.
\newblock \doi{10.1287/mnsc.46.3.404.12070}.

\bibitem[Setlur et~al.(2025)Setlur, Nagpal, Fisch, Geng, Eisenstein, Agarwal,
  Agarwal, Berant, and Kumar]{pavs2025}
Amrith Setlur, Chirag Nagpal, Adam Fisch, Xinyang Geng, Jacob Eisenstein,
  Rishabh Agarwal, Alekh Agarwal, Jonathan Berant, and Aviral Kumar.
\newblock Rewarding progress: Scaling automated process verifiers for {LLM}
  reasoning.
\newblock In \emph{International Conference on Learning Representations}, 2025.
\newblock URL \url{https://openreview.net/forum?id=A6Y7AqlzLW}.
\newblock Spotlight.

\bibitem[Shao et~al.(2024)Shao, Wang, Zhu, Xu, Song, Bi, Zhang, Zhang, Li, Wu,
  et~al.]{grpo2024}
Zhihong Shao, Peiyi Wang, Qihao Zhu, Runxin Xu, Junxiao Song, Xiao Bi, Haowei
  Zhang, Mingchuan Zhang, YK~Li, Y~Wu, et~al.
\newblock {DeepSeekMath}: Pushing the limits of mathematical reasoning in open
  language models.
\newblock \emph{arXiv preprint arXiv:2402.03300}, 2024.
\newblock Introduces GRPO algorithm.

\bibitem[Thind et~al.(2025)Thind, Sun, Liang, and Yang]{optimai2025}
Raghav Thind, Youran Sun, Ling Liang, and Haizhao Yang.
\newblock {OptimAI}: Optimization from natural language using {LLM}-powered
  {AI} agents.
\newblock \emph{arXiv preprint arXiv:2504.16918}, 2025.

\bibitem[Tie et~al.(2025)Tie, Yuan, Zhao, Hu, Gu, Zhang, Zhang, Wu, Tu, Jin,
  Wen, Chen, Zhou, and Sun]{correctbench2025}
Guiyao Tie, Zenghui Yuan, Zeli Zhao, Chaoran Hu, Tianhe Gu, Ruihang Zhang,
  Sizhe Zhang, Junran Wu, Xiaoyue Tu, Ming Jin, Qingsong Wen, Lixing Chen, Pan
  Zhou, and Lichao Sun.
\newblock Can {LLMs} correct themselves? a benchmark of self-correction in
  {LLMs}.
\newblock In \emph{Advances in Neural Information Processing Systems},
  volume~38, 2025.
\newblock NeurIPS 2025 Datasets and Benchmarks Track. arXiv:2510.16062.

\bibitem[Wang et~al.(2024)Wang, Li, Shao, Xu, Dai, Li, Chen, Wu, and
  Sui]{mathshepherd2024}
Peiyi Wang, Lei Li, Zhihong Shao, Runxin Xu, Damai Dai, Yifei Li, Deli Chen,
  Yu~Wu, and Zhifang Sui.
\newblock {Math-Shepherd}: Verify and reinforce {LLMs} step-by-step without
  human annotations.
\newblock In \emph{Proceedings of the 62nd Annual Meeting of the Association
  for Computational Linguistics (Volume 1: Long Papers)}, pages 9426--9439,
  2024.
\newblock arXiv:2312.08935.

\bibitem[Wei et~al.(2022)Wei, Wang, Schuurmans, Bosma, Xia, Chi, Le, Zhou,
  et~al.]{cot2022}
Jason Wei, Xuezhi Wang, Dale Schuurmans, Maarten Bosma, Fei Xia, Ed~Chi, Quoc~V
  Le, Denny Zhou, et~al.
\newblock Chain-of-thought prompting elicits reasoning in large language
  models.
\newblock \emph{Advances in Neural Information Processing Systems},
  35:\penalty0 24824--24837, 2022.
\newblock arXiv:2201.11903.

\bibitem[Xiao et~al.(2024)Xiao, Zhang, Wu, Xu, Wang, Han, Fu, Zhong, Zeng,
  Song, et~al.]{chainofexperts2024}
Ziyang Xiao, Dongxiang Zhang, Yangjun Wu, Lilin Xu, Yuan~Jessica Wang, Xiongwei
  Han, Xiaojin Fu, Tao Zhong, Jia Zeng, Mingli Song, et~al.
\newblock Chain-of-experts: When {LLMs} meet complex operations research
  problems.
\newblock In \emph{International Conference on Learning Representations}, 2024.
\newblock OpenReview:HobyL1B9CZ.

\bibitem[Yang et~al.(2024)Yang, Wang, Huang, Guo, Shi, Han, Feng, Song, Liang,
  and Tang]{optibench2024}
Zhicheng Yang, Yiwei Wang, Yinya Huang, Zhijiang Guo, Wei Shi, Xiongwei Han,
  Liang Feng, Linqi Song, Xiaodan Liang, and Jing Tang.
\newblock {OptiBench} meets {ReSocratic}: Measure and improve {LLMs} for
  optimization modeling.
\newblock \emph{arXiv preprint arXiv:2407.09887}, 2024.

\bibitem[Yu et~al.(2025)Yu, Zhang, Zhu, Yuan, Zuo, Yue, Fan, Liu, Liu, Liu,
  et~al.]{dapo2025}
Qiying Yu, Zheng Zhang, Ruofei Zhu, Yufeng Yuan, Xiaochen Zuo, Yu~Yue, Tiantian
  Fan, Gaohong Liu, Lingjun Liu, Xin Liu, et~al.
\newblock {DAPO}: An open-source {LLM} reinforcement learning system at scale.
\newblock \emph{arXiv preprint arXiv:2503.14476}, 2025.

\bibitem[Zelikman et~al.(2022)Zelikman, Wu, Mu, and Goodman]{star2022}
Eric Zelikman, Yuhuai Wu, Jesse Mu, and Noah Goodman.
\newblock {STaR}: Bootstrapping reasoning with reasoning.
\newblock In \emph{Advances in Neural Information Processing Systems},
  volume~35, pages 15476--15488, 2022.

\bibitem[Zhang and Luo(2025)]{orllmagent2025}
Bowen Zhang and Pengcheng Luo.
\newblock {OR-LLM-Agent}: Automating modeling and solving of operations
  research optimization problem with reasoning large language model.
\newblock \emph{arXiv preprint arXiv:2503.10009}, 2025.

\bibitem[Zhang et~al.(2025{\natexlab{a}})]{biprm2025}
Lingyin Zhang et~al.
\newblock The bidirectional process reward model.
\newblock \emph{arXiv preprint arXiv:2508.01682}, 2025{\natexlab{a}}.

\bibitem[Zhang et~al.(2025{\natexlab{b}})Zhang, Zhang, and
  Zhao]{zhang2025predicting}
Runze Zhang, Xiaowei Zhang, and Mingyang Zhao.
\newblock Predicting effects, missing distributions: Evaluating {LLMs} as human
  behavior simulators in operations management.
\newblock \emph{arXiv preprint arXiv:2510.03310}, 2025{\natexlab{b}}.

\bibitem[Zhao et~al.(2025)]{aimbench2025}
Xuhua Zhao et~al.
\newblock {AIM-Bench}: Evaluating decision-making biases of agentic {LLM} as
  inventory manager.
\newblock \emph{arXiv preprint arXiv:2508.11416}, 2025.

\bibitem[Zhou et~al.(2025)Zhou, Yang, Xin, Chen, He, and Ge]{dpbench2025}
Chenyu Zhou, Jingyuan Yang, Linwei Xin, Yitian Chen, Ziyan He, and Dongdong Ge.
\newblock Auto-formulating dynamic programming problems with large language
  models.
\newblock \emph{arXiv preprint arXiv:2507.11737}, 2025.

\end{thebibliography}

\appendix

\section{Additional Related Work}
\label{app:related_full}

\textbf{OR Benchmarks.}
Table~\ref{tab:positioning} compares the two \ORLoopBench{} components, \ORDebug{} and \ORBias{}, with existing OR benchmarks. NL4Opt~\citep{nl4opt2022} introduced the task of translating natural language to linear program formulations. Subsequent work explored both \emph{prompt-based} approaches using in-context learning~\citep{chainofexperts2024,optimus2024,bertsimas2024robust,autoformulation2025,optitree2025} and \emph{learning-based} methods fine-tuning on domain data~\citep{optibench2024,orlm2025,sirl2025,mind2026}. Other recent efforts include ORQA~\citep{orqa2025} for OR reasoning evaluation, LLMOPT~\citep{llmopt2024}, and agent-based systems~\citep{orllmagent2025,optimai2025}. OptMATH~\citep{optmath2025} proposed bidirectional data synthesis for scalable training data generation. MAMO~\citep{mamo2024} further contributed benchmarks spanning easy and complex LP instances with solver integration. PILOT-Bench~\citep{pilotbench2026} evaluates LLM workflow execution under probabilistic tool failures and varying instruction quality. However, existing benchmarks largely evaluate \textbf{static, one-shot} formulation accuracy: a model produces code, and evaluation checks correctness without iterative refinement or solver feedback. Recent work has extended formulation benchmarks to dynamic programming: DP-Bench~\citep{dpbench2025} introduces 132 textbook-level DP problems and shows that even SOTA models struggle with stochastic transitions, achieving only 59.1\% accuracy. Our work addresses a complementary gap: evaluating how models respond to solver feedback and iteratively correct their formulations, focusing on LP infeasibility debugging rather than one-shot formulation.

\textbf{Formulation Training vs Debugging.}
Several recent systems use solvers or search to improve NL-to-model formulation. SIRL~\citep{sirl2025} trains formulation models with solver-informed rewards; Autoformulation~\citep{autoformulation2025} and OptiTree~\citep{optitree2025} search over hierarchical formulation choices; MIND~\citep{mind2026} constructs localized error-driven training data for automated modeling. OptiChat introduced natural-language interaction for infeasible optimization models using GPT-4, Gurobi \IIS{} feedback, prompt engineering, and practitioner studies~\citep{optichat2024}; its extended system broadens this interface to interpretation, retrieval, sensitivity, what-if, and counterfactual queries over 24 real-world-style optimization models~\citep{optichat2025}. Earlier expert-system work such as ANALYZE studied computer-assisted analysis of LP models before modern LLM interfaces~\citep{greenberg1983}. A supply-chain-focused predecessor, OptiRepair~\citep{ao2026optirepair}, studies closed-loop diagnosis and repair for multi-echelon inventory models. These works are closely related but differ in scope: \ORDebug{} provides a controlled benchmark, MDP action interface, objective-preservation checks, and solver-verified training/evaluation for iterative constraint-level repair.

\textbf{Self-Correction in LLMs.}
While Chain-of-Thought prompting~\citep{cot2022} improved LLM reasoning through step-by-step decomposition, it does not address error correction when intermediate steps fail. CorrectBench~\citep{correctbench2025} provided the first systematic study of LLM self-correction, documenting that 64.5\% of errors stem from correction failures rather than initial mistakes. However, CorrectBench focuses on general programming tasks and explicitly excludes the OR domain, where feedback is deterministic and mathematically verifiable. SWE-bench~\citep{swebench2024} evaluates code debugging through unit test feedback, but software testing differs from mathematical optimization verification in a key respect: unit tests sample behavior while solvers provide certified infeasibility certificates. Self-Refine~\citep{selfrefine2023} demonstrated iterative refinement with self-generated feedback, and STaR~\citep{star2022} showed that models can bootstrap reasoning by learning from their own correct solutions. \citet{rlselfcorrect2024} further demonstrated that RL can explicitly train models for iterative self-correction. However, these approaches rely on heuristic quality signals rather than formal verification. Our benchmarks use Gurobi's \IIS{} computation as a noise-free oracle, enabling precise evaluation of diagnostic reasoning. For multi-agent debugging scenarios, AgentGit~\citep{agentgit2025} provides version control abstractions that could complement our single-agent evaluation framework, while reliability analyses of delegated LLM planning characterize limits from communication and information compression~\citep{ao2026multiagentplanning}.

\textbf{RLVR and Process Supervision.}
Reinforcement Learning with Verifiable Rewards (RLVR) trains reasoning models with automatically checkable reward signals. Building on Proximal Policy Optimization (PPO)~\citep{ppo2017} and instruction tuning with human feedback~\citep{instructgpt2022}, alternative approaches include Direct Preference Optimization (DPO)~\citep{dpo2023} for offline alignment and Group Relative Policy Optimization (\GRPO{})~\citep{grpo2024} for online training. DeepSeek-R1~\citep{deepseekr1} demonstrated that \GRPO{} with outcome verification can induce sophisticated reasoning without explicit chain-of-thought supervision. DAPO~\citep{dapo2025} further improved RLVR systems with KL-penalty removal and asymmetric clipping. T\"ulu 3~\citep{tulu3rlvr} extended this to a broader post-training pipeline. Recent analysis~\citep{sftrl2025} reveals that while SFT tends to memorize training patterns, RL promotes generalization, a finding that motivates our two-stage training approach. For process supervision, ``Let's Verify Step by Step''~\citep{lightman2024let} introduced human-annotated step-level rewards, later automated by Math-Shepherd~\citep{mathshepherd2024} through Monte Carlo estimation. BiPRM~\citep{biprm2025} proposed bidirectional verification for mathematical reasoning. PAVs~\citep{pavs2025} formalized Process Advantage Verifiers for efficient alignment. Our \PRM{} training builds on these foundations, adapting process supervision to the OR debugging domain where step-level progress can be automatically verified through \IIS{} size reduction. Concurrent work OR-R1~\citep{orr12025} also explores test-time RL for OR modeling, though focusing on formulation rather than iterative debugging.

\textbf{Verifiable Feedback in OR.}
Prior self-correction work relies on either self-generated feedback, which can be unreliable, or heuristic metrics like test pass rates, which sample rather than verify. Solver feedback differs in kind: \IIS{} computation provides deterministic, solver-certified information about a current infeasibility certificate, with a long optimization lineage in infeasibility analysis and commercial-solver implementations~\citep{chinneck2008,gurobi2024}. We adapt RLVR to the OR domain where: (1) the oracle provides verifiable rewards without human annotation, (2) step-level progress is measurable through \IIS{} size reduction, and (3) diagnostic accuracy can be computed against ground-truth constraint labels, enabling fully automated training and evaluation pipelines.

\textbf{Test-Time Scaling and Inference Compute.}
Recent work explores scaling inference compute through repeated sampling and verification. AlphaCode~\citep{alphacode2022} demonstrated that pass@k metrics reveal headroom beyond single-attempt accuracy, with competitive programming performance improving through sampling. DeepSeek-R1~\citep{deepseekr1} showed that extended reasoning chains, a form of implicit test-time scaling, improve mathematical problem-solving. For code generation, best-of-n sampling with execution feedback~\citep{chen2021humaneval} remains a simple but effective strategy, and scheduling-based approaches can further reduce inference costs~\citep{ao2026llminference}. Our inference scaling analysis (Section~\ref{sec:experiments}, Appendix~\ref{app:scaling}) contributes domain-specific findings: OR debugging exhibits similar scaling behavior to code generation (+17\% from $k=1$ to $k=5$), but domain-specific models achieve superior sample efficiency: o4-mini requires 2.8$\times$ as many tokens per successful solution.

\textbf{Behavioral Rationality in LLMs.}
Recent work has examined LLMs and predictive models as decision inputs in OM contexts, including demand simulation~\citep{zhang2025predicting}, resource-constrained pricing with forecast uncertainty~\citep{ao2026pricinginfo}, and service-system selection from textual evidence under limited human audits~\citep{ao2026servicesystems}. AIM-Bench~\citep{aimbench2025} documented the ``pull-to-center'' phenomenon in LLM inventory decisions, where models regress toward mean predictions regardless of the optimal decision. This finding echoes classical behavioral OR on human decision-making under uncertainty~\citep{schweitzer2000decision}. \ORBias{} builds on this behavioral question with explicit ID/OOD splits and shows that curriculum learning can mitigate these biases, achieving 48\% bias reduction on OOD scenarios.

\textbf{Connection to Human Behavioral Biases.}
The pull-to-center bias in LLMs mirrors documented human biases in newsvendor decisions~\citep{schweitzer2000decision,kremer2010demand}. Our curriculum training, which explicitly teaches directional sensitivity, can be viewed as a cognitive debiasing intervention for LLMs, analogous to decision support systems designed for human operators.

\section*{Appendix Overview}
\label{app:overview}

This appendix provides supporting material organized as follows:

\begin{itemize}
    \item \textbf{Appendix~\ref{app:examples}}: Execution examples comparing trained and baseline models
    \item \textbf{Appendix~\ref{app:task_details}}: Task parameters, prompts, and evaluation criteria
    \item \textbf{Appendix~\ref{app:construction}}: Benchmark construction algorithms and validation
    \item \textbf{Appendix~\ref{app:training}}: MDP formulation and training procedures
    \item \textbf{Appendix~\ref{app:debug_results}}: Complete evaluation results
    \item \textbf{Appendix~\ref{app:bias_results}}: Complete \ORBias{} and feedback-protocol results
    \item \textbf{Appendix~\ref{app:ablation}}: Training ablations and robustness checks
    \item \textbf{Appendix~\ref{app:scaling}}: Token efficiency and difficulty analysis
    \item \textbf{Appendix~\ref{app:model_selection}}: Base model selection and prompting baselines
\end{itemize}

\section{Execution Examples}
\label{app:examples}

This appendix provides execution traces illustrating the difference between trained and baseline models on \ORDebug{} problems.

\subsection{Success Case: Type C Error (Coefficient Modification)}

\textbf{Problem Description.}
A production planning LP where minimum requirements on individual products exceed the total capacity constraint:
\begin{verbatim}
m = gp.Model("production")
x = m.addVars(3, name="x")
m.addConstr(x[0] + x[1] + x[2] <= 100, name="c1_total")
m.addConstr(x[0] >= 60, name="c2_min_0")
m.addConstr(x[1] >= 50, name="c3_min_1")  # KEY
m.addConstr(x[2] >= 0, name="c4_min_2")
\end{verbatim}
The minimum sum ($60+50+0=110$) exceeds capacity ($100$), causing infeasibility.

\textbf{Initial State.}
\begin{itemize}
    \item Status: \INFEASIBLE{}
    \item \IIS{}: \{c1\_total, c2\_min\_0, c3\_min\_1\}
    \item Ground Truth Fix: Relax c3\_min\_1 to $x[1] \geq 40$
\end{itemize}

\textbf{Qwen-\GRPO{} Trajectory (2 steps).}
\begin{enumerate}
    \item \textbf{Step 1: Diagnostic Action}
    \begin{itemize}
        \item Action: \textsc{Get\_IIS}
        \item Response: ``\IIS{} contains c1\_total, c2\_min\_0, c3\_min\_1. The minimum sum $(60+50+0=110)$ exceeds capacity $(100)$. Key constraint: c3\_min\_1 ($x[1] \geq 50$) is over-restrictive.''
        \item Diagnosis: [c3\_min\_1] \checkmark\ (matches ground truth)
    \end{itemize}

    \item \textbf{Step 2: Repair Action}
    \begin{itemize}
        \item Action: \textsc{Relax}(c3\_min\_1, $x[1] \geq 40$)
        \item Response: ``Modified constraint allows sum $= 60+40+0 = 100 =$ capacity. Model is now feasible.''
        \item Status: \OPTIMAL{} \checkmark
    \end{itemize}
\end{enumerate}

\textbf{\SFT{} Baseline Trajectory (5 steps).}
\begin{enumerate}
    \item \textsc{Get\_IIS} $\rightarrow$ [c1, c2, c3]
    \item \textsc{Check\_Slack}(c1) $\rightarrow$ ``c1 is binding''
    \item \textsc{Relax}(c2) $\rightarrow$ Still \INFEASIBLE{}
    \item \textsc{Relax}(c1) $\rightarrow$ Status changes but objective degraded
    \item \textsc{Rewrite}(c3) $\rightarrow$ Finally \OPTIMAL{}
\end{enumerate}

\textbf{Analysis.}
The \GRPO{}-trained model directly identifies the key constraint (c3\_min\_1) and applies a minimal fix in 2 steps. The \SFT{} baseline explores multiple constraints through trial-and-error, taking 5 steps and initially relaxing the wrong constraint (c2).

\subsection{Challenge Case: Optimal Selection (Type I)}

\textbf{Problem Description.}
A composite error with 12 constraints in the \IIS{}, involving resource allocation, capacity, and flow balance conflicts simultaneously.

\textbf{Model Behavior Comparison.}

\begin{table}[h]
\caption{Model behavior comparison on composite error case.}
\label{tab:model_comparison}
\centering
\small
\begin{tabular}{@{}lcccc@{}}
\toprule
Model & Steps & \DA{} & Success & Analysis \\
\midrule
Qwen-\GRPO{} & 7 & 58\% & Yes & Systematic decomposition \\
o4-mini & 12 & 42\% & Yes & Trial-and-error \\
gpt-5.2-chat & 15 & 33\% & Yes & Excessive exploration \\
gpt-4.1-mini & 20 & 17\% & No & Error cascade \\
\bottomrule
\end{tabular}
\end{table}

\textbf{Error Cascade Pattern (gpt-4.1-mini).}
\begin{itemize}
    \item Steps 1--5: Correct diagnosis of constraints c3, c7
    \item Step 6: Incorrect fix of c3 introduces new conflict
    \item Steps 7--12: Attempts to fix cascading errors
    \item Steps 13--18: Reverts and retries with different approach
    \item Steps 19--20: Timeout without resolution
\end{itemize}

\textbf{Analysis.}
API models often fail to reason about constraint interactions, leading to fixes that introduce new conflicts. The trained model learns to decompose complex \IIS{} sets and address constraints in dependency order.

\subsection{Available Actions}

\begin{table}[h]
\centering
\small
\begin{tabular}{@{}llc@{}}
\toprule
Action & Description & Modifies State \\
\midrule
\textsc{Get\_IIS} & Compute Irreducible Infeasible Subsystem & No \\
\textsc{Check\_Slack} & Get constraint slack values & No \\
\textsc{Check\_Bound} & Get variable bound status & No \\
\textsc{Relax}$(c, \delta)$ & Modify constraint RHS & Yes \\
\textsc{Drop}$(c)$ & Remove constraint entirely & Yes \\
\textsc{Rewrite}$(c, \text{expr})$ & Replace constraint expression & Yes \\
\textsc{Submit} & Submit current model for evaluation & Yes \\
\bottomrule
\end{tabular}
\end{table}

\subsection{Success Examples}

\textbf{Example: Type E (Multi-Constraint Conflict).}
A transportation problem where demand exceeds supply:
\begin{verbatim}
# Problem setup
m = gp.Model("transport")
supply = m.addVars(3, name="s")
demand = m.addVars(4, name="d")

# Supply constraints (total = 100)
m.addConstr(supply[0] <= 40, "s0_cap")
m.addConstr(supply[1] <= 35, "s1_cap")
m.addConstr(supply[2] <= 25, "s2_cap")

# Demand constraints (total = 115, infeasible!)
m.addConstr(demand[0] >= 30, "d0_min")
m.addConstr(demand[1] >= 35, "d1_min")  # KEY
m.addConstr(demand[2] >= 25, "d2_min")
m.addConstr(demand[3] >= 25, "d3_min")

# Flow balance
m.addConstr(quicksum(supply) == quicksum(demand))
\end{verbatim}

\textbf{IIS}: \{s0\_cap, s1\_cap, s2\_cap, d0\_min, d1\_min, flow\_balance\}

\textbf{\Qwen{}-\GRPO{} Solution (2 steps):}
\begin{enumerate}
    \item \textsc{Get\_IIS}: ``Total supply (100) is less than total demand (115). The largest demand constraint (d1\_min = 35) should be reduced.''
    \item \textsc{Relax}(d1\_min, -15): ``Reducing demand 1 from 35 to 20 balances supply and demand.'' $\rightarrow$ \OPTIMAL{}
\end{enumerate}

\textbf{gpt-4.1 Solution (6 steps):}
\begin{enumerate}
    \item \textsc{Get\_IIS} $\rightarrow$ Lists all constraints
    \item \textsc{Relax}(s0\_cap, +10) $\rightarrow$ Still \INFEASIBLE{}
    \item \textsc{Relax}(s1\_cap, +10) $\rightarrow$ Still \INFEASIBLE{}
    \item \textsc{Check\_Slack}(all) $\rightarrow$ Sees all slacks are 0
    \item \textsc{Relax}(d3\_min, -10) $\rightarrow$ Still \INFEASIBLE{}
    \item \textsc{Relax}(d1\_min, -15) $\rightarrow$ \OPTIMAL{}
\end{enumerate}

\textbf{Analysis}: The trained model directly identifies that demand constraints are easier to adjust than capacity constraints (a domain-specific heuristic) and targets the largest demand. The API model attempts supply increases first, which don't resolve the fundamental imbalance.

\subsection{\ORBias{} Examples}

\textbf{ID Scenario: CR = 0.5 (Balanced).}
\begin{verbatim}
Price: $60, Cost: $30, Salvage: $0
Mean demand: 100, Std: 20
CR = (60-30)/(60-0) = 0.5
Q* = 100 + 20 * Phi^{-1}(0.5) = 100
\end{verbatim}

Model responses:
\begin{itemize}
    \item \textbf{gpt-5-mini}: $Q = 100$ (Correct, bias = 0\%)
    \item \textbf{\Qwen{}-Curriculum}: $Q = 100$ (Correct)
\end{itemize}

\textbf{OOD Scenario: CR = 0.1 (Low margin).}
\begin{verbatim}
Price: $55, Cost: $50, Salvage: $5
Mean demand: 100, Std: 20
CR = (55-50)/(55-5) = 0.1
Q* = 100 + 20 * Phi^{-1}(0.1) = 74.4
\end{verbatim}

Model responses:
\begin{itemize}
    \item \textbf{gpt-5-mini}: $Q = 95$ (Over-order, bias = +28\%)
    \item \textbf{\Qwen{}-Curriculum}: $Q = 76$ (Correct, bias = +2\%)
\end{itemize}

\textbf{Analysis}: At extreme CR values, gpt-5-mini exhibits severe pull-to-center bias, ordering near the mean despite the optimal quantity being 32\% below. The curriculum-trained model correctly adjusts its order downward, demonstrating learned sensitivity to the critical ratio.

\section{Task Function Components}
\label{app:task_details}

This appendix details the task parameters, prompts, and evaluation criteria used in \ORDebug{}.

\subsection{Task and Tool Details}
\label{app:task_params}

Table~\ref{tab:task_params} lists the key parameters governing the \ORDebug{} environment.

\begin{table}[h]
\caption{Task parameters for \ORDebug{}.}
\label{tab:task_params}
\centering
\small
\begin{tabular}{@{}llp{4.5cm}@{}}
\toprule
Parameter & Value & Description \\
\midrule
max\_steps & 50 & Maximum MDP steps before timeout \\
timeout & 10s & Per-step solver timeout \\
iis\_method & minimal & Gurobi \IIS{} computation method \\
\bottomrule
\end{tabular}
\end{table}

Table~\ref{tab:actions} describes the available actions and their properties. Diagnostic actions gather information without modifying the model state, while repair actions apply changes.

\begin{table}[h]
\caption{Action space for \ORDebug{}.}
\label{tab:actions}
\centering
\small
\begin{tabular}{@{}lp{5cm}c@{}}
\toprule
Action & Description & Modifies State \\
\midrule
\textsc{Get\_IIS} & Compute Irreducible Infeasible Subsystem & No \\
\textsc{Check\_Slack} & Get constraint slack values & No \\
\textsc{Check\_Bound} & Get variable bound status & No \\
\textsc{Relax}$(c, \delta)$ & Modify constraint RHS by $\delta$ & Yes \\
\textsc{Drop}$(c)$ & Remove constraint entirely & Yes \\
\textsc{Rewrite}$(c, \text{expr})$ & Replace constraint expression & Yes \\
\textsc{Submit} & Submit current model for evaluation & Yes \\
\bottomrule
\end{tabular}
\end{table}

\subsection{Prompt Templates}
\label{app:prompts}

We use three prompt variants in our experiments.

\textbf{Baseline Prompt.}
The minimal prompt provides the problem description, code, and \IIS{} without additional guidance:
\begin{verbatim}
You are an OR debugging assistant. Given an
infeasible linear program, analyze the IIS
and suggest fixes to restore feasibility.

Problem: {problem_nl}
Code: {code}
Status: INFEASIBLE
IIS: {iis_constraints}

Provide your diagnosis and suggested action.
\end{verbatim}

\textbf{Chain-of-Thought (CoT) Prompt.}
The CoT prompt adds explicit reasoning steps:
\begin{verbatim}
You are an OR debugging assistant. Follow
this reasoning process:

1. ANALYZE: Examine each IIS constraint's
   role in the problem formulation
2. IDENTIFY: Determine the root cause
   constraint causing infeasibility
3. PROPOSE: Suggest a minimal fix that
   preserves problem semantics
4. VERIFY: Explain why the fix resolves
   the conflict

Problem: {problem_nl}
Code: {code}
Status: INFEASIBLE
IIS: {iis_constraints}
\end{verbatim}

\textbf{Optimal Workflow Prompt.}
Used for \SFT{} data collection, this prompt encodes expert heuristics:
\begin{verbatim}
You are an expert OR debugger. Your task is:
1. Always get the IIS first
2. Identify the single most restrictive
   constraint in the IIS
3. Apply minimal relaxation (prefer relax
   over drop when possible)
4. Preserve original problem semantics

Problem: {problem_nl}
Code: {code}
Status: INFEASIBLE
IIS: {iis_constraints}
\end{verbatim}

\subsection{Tool Result Simulator}
\label{app:simulator}

For training without solver access, we provide a deterministic simulator that estimates action outcomes based on ground truth labels:

\begin{algorithm}[h]
\caption{Tool Result Simulator}
\label{alg:simulator}
\begin{algorithmic}[1]
\REQUIRE Action $a$, State $s$, Ground truth $\mathcal{G}$
\ENSURE Simulated next state $s'$
\IF{$a.\text{type} \in \{\text{GET\_IIS}, \text{CHECK\_SLACK}\}$}
    \STATE $s' \gets s$ \COMMENT{Info gathering, no change}
\ELSIF{$a.\text{target} \in \mathcal{G}.\text{key\_constraints}$}
    \IF{$a.\text{type} = \text{RELAX}$}
        \STATE $s'.\text{status} \gets \text{OPTIMAL}$ w.p. 0.9
    \ELSIF{$a.\text{type} = \text{DROP}$}
        \STATE $s'.\text{status} \gets \text{OPTIMAL}$ w.p. 0.7
    \ENDIF
\ELSIF{$a.\text{target} \in \mathcal{G}.\text{IIS}$}
    \STATE Reduce $|\text{IIS}|$ by 1 w.p. 0.5
\ELSE
    \STATE $s'.\text{status} \gets \text{INFEASIBLE}$ \COMMENT{Wrong target}
\ENDIF
\RETURN $s'$
\end{algorithmic}
\end{algorithm}

\subsection{Task Result Evaluation}
\label{app:evaluation_criteria}

We categorize episode outcomes into three classes:

\begin{itemize}
    \item \textbf{Full Success}: Status reaches \OPTIMAL{} and the Optimality Preservation score $\OP > 0.95$ (objective value within 5\% of original).

    \item \textbf{Partial Success}: Status reaches \OPTIMAL{} but the objective is degraded, with $0.8 < \OP \leq 0.95$.

    \item \textbf{Failure}: The model remains \INFEASIBLE{} after max\_steps, or $\OP \leq 0.8$ indicating the fix changed the objective by $>$20\%.
\end{itemize}

The \RRk{k} metric counts full successes achieved within $k$ repair steps of a single episode. Partial successes are counted for the overall Recovery Rate (RR) but not for \RRk{k} to reward efficient diagnosis.

\subsection{Gurobi Configuration}
\label{app:gurobi}

We use Gurobi 11.0.0 with the following configuration~\citep{gurobi2024}:

\begin{table}[h]
\caption{Gurobi solver configuration.}
\label{tab:gurobi_config}
\centering
\small
\begin{tabular}{@{}ll@{}}
\toprule
Parameter & Value \\
\midrule
IISMethod & 1 (minimal IIS) \\
TimeLimit & 10 seconds \\
OutputFlag & 0 (suppress output) \\
Threads & 4 \\
MIPGap & 0.01 \\
\bottomrule
\end{tabular}
\end{table}

\textbf{Why Minimal IIS?}
Gurobi offers several \IIS{} computation methods, building on standard infeasibility-analysis tools in mathematical optimization~\citep{chinneck2008}. We use the minimal method (IISMethod=1) because:
\begin{itemize}
    \item It produces the smallest possible IIS, making diagnosis more focused.
    \item It is deterministic across runs.
    \item It completes within reasonable time (typically $<$1s) for our problem sizes.
\end{itemize}

\subsection{Evaluation Protocol}
\label{app:eval_protocol}

The evaluation follows a standardized step-budget protocol. Each model receives one solver-interaction episode per problem; the episode horizon is $T=50$, and \RRk{k} is reported for repair-step budgets such as $k \in \{1,3,5,10,20\}$.

\begin{algorithm}[h]
\caption{Step-Budget Evaluation Protocol}
\label{alg:eval_protocol}
\begin{algorithmic}[1]
\REQUIRE Problem set $\mathcal{P}$, model $M$, report budgets $\mathcal{K}$, episode horizon $T$
\ENSURE Metrics $\{\RRk{k}: k \in \mathcal{K}\}, \DA, \text{Steps}$
\FOR{each problem $p \in \mathcal{P}$}
    \STATE $\text{first\_success\_step} \gets \infty$
    \FOR{$t = 1$ \textbf{to} $T$}
        \STATE Model $M$ observes the current solver state and emits one action
        \STATE Apply action, run Gurobi, and recompute \IIS{} if needed
        \IF{state reaches \OPTIMAL{} with $\OP > 0.95$}
            \STATE $\text{first\_success\_step} \gets t$
            \STATE \textbf{break}
        \ENDIF
    \ENDFOR
    \STATE Record first\_success\_step, DA, total\_steps
\ENDFOR
\STATE For each $k \in \mathcal{K}$, compute the fraction of problems with $\text{first\_success\_step} \leq k$
\RETURN Metrics
\end{algorithmic}
\end{algorithm}

\subsection{Reproducibility Checklist}
\label{app:reproducibility}

\begin{itemize}
    \item[$\checkmark$] Benchmark files are available at \url{https://github.com/Archer222arc/ORLoopBench}
    \item[$\checkmark$] Random seeds fixed for all experiments
    \item[$\checkmark$] Hardware and software versions documented
    \item[$\checkmark$] Training hyperparameters fully specified
    \item[$\checkmark$] Evaluation protocol standardized across models
    \item[$\checkmark$] Gurobi configuration deterministic
\end{itemize}

\section{Benchmark Construction Details}
\label{app:construction}

This appendix details the algorithms and validation procedures used to construct \ORDebug{} and \ORBias{}.

\subsection{Saboteur Generation}
\label{app:saboteur}

The Saboteur generates controlled infeasibilities by applying targeted corruptions to feasible LP instances. We describe two representative error types.

\textbf{Type A: Direction Flip.}
This error reverses the sense of an inequality constraint, turning a minimum requirement into a maximum limit or vice versa:

\begin{algorithm}[h]
\caption{Sabotage: Direction Flip (Type A)}
\label{alg:sabotage_flip}
\begin{algorithmic}[1]
\REQUIRE Feasible model $M$, target constraint $c$
\ENSURE Infeasible model $M'$, ground truth $(c, \text{``flip''})$
\STATE $M' \gets \text{copy}(M)$
\IF{$c.\text{sense} = \text{``}\geq\text{''}$}
    \STATE $c'.\text{sense} \gets \text{``}\leq\text{''}$
\ELSE
    \STATE $c'.\text{sense} \gets \text{``}\geq\text{''}$
\ENDIF
\STATE Replace $c$ with $c'$ in $M'$
\RETURN $M'$, $(c, \text{``flip''})$
\end{algorithmic}
\end{algorithm}

\textbf{Example.}
Original: $x + y \geq 10$ (minimum production requirement).
Sabotaged: $x + y \leq 10$ (maximum limit, conflicts with other minimums).

\textbf{Type E: Multi-Constraint Conflict.}
This error creates interlocked constraints that both must be fixed:

\begin{algorithm}[h]
\caption{Sabotage: Multi-Constraint Conflict (Type E)}
\label{alg:sabotage_resource}
\begin{algorithmic}[1]
\REQUIRE Feasible model $M$, demand constraints $\mathcal{D}$, factor $\alpha > 1$
\ENSURE Infeasible model $M'$, ground truth
\STATE $M' \gets \text{copy}(M)$
\STATE $c^* \gets \arg\max_{c \in \mathcal{D}} c.\text{rhs}$ \COMMENT{Largest demand}
\STATE $c^*.\text{rhs} \gets c^*.\text{rhs} \times \alpha$
\STATE \textbf{assert} $\sum_{c \in \mathcal{D}} c.\text{rhs} >$ capacity
\RETURN $M'$, $(c^*, \text{``reduce rhs''})$
\end{algorithmic}
\end{algorithm}

\textbf{Example.}
Original demands sum to 90 with capacity 100.
Sabotaged demands sum to 117 ($\alpha = 1.3$), exceeding capacity.

\subsection{Newsvendor Generation}
\label{app:newsvendor}

For \ORBias{}, we generate newsvendor scenarios with controlled Critical Ratio (\CR{}) distributions.

\begin{algorithm}[h]
\caption{Newsvendor Scenario Generation}
\label{alg:newsvendor}
\begin{algorithmic}[1]
\REQUIRE Target \CR{} range $[\CR_{\min}, \CR_{\max}]$
\ENSURE Scenario parameters $(p, c, s, \mu, \sigma, Q^*)$
\STATE $\CR \sim \text{Uniform}(\CR_{\min}, \CR_{\max})$
\STATE $p \sim \text{Uniform}(10, 100)$ \COMMENT{Unit price}
\STATE $s \sim \text{Uniform}(0, 0.3p)$ \COMMENT{Salvage value}
\STATE $c \gets p - \CR \cdot (p - s)$ \COMMENT{Derive cost from \CR{}}
\STATE $\mu \sim \text{Uniform}(50, 200)$ \COMMENT{Demand mean}
\STATE $\sigma \sim \text{Uniform}(10, 50)$ \COMMENT{Demand std}
\STATE $Q^* \gets \mu + \sigma \cdot \Phi^{-1}(\CR)$ \COMMENT{Optimal quantity}
\RETURN $(p, c, s, \mu, \sigma, Q^*)$
\end{algorithmic}
\end{algorithm}

The unit cost $c$ derives from the target \CR{} using the relationship $\CR = (p - c) / (p - s)$. This ensures the generated scenario has the desired \CR{} while maintaining realistic cost structures.

\subsection{Newsvendor Evaluation Difficulty Design}
\label{app:bias_curriculum}

The 4-level evaluation difficulty scheme for \ORBias{} is designed to test rationality under increasing complexity.

\textbf{Level Design Rationale.}
\begin{itemize}
    \item \textbf{L1 (Foundations)}: \CR{} $\in [0.4, 0.6]$ produces $Q^* \approx \mu$ (within $\pm 0.25\sigma$), minimizing pull-to-center effects. This establishes baseline formula application capability.
    \item \textbf{L2 (Bias Traps)}: \CR{} $\in [0.05, 0.2)$ yields $Q^* < \mu - \sigma$ (order less than mean minus one standard deviation), while \CR{} $\in (0.8, 0.95]$ yields $Q^* > \mu + \sigma$. These extremes maximally trigger the pull-to-center bias documented in behavioral operations research.
    \item \textbf{L3 (Robustness)}: Distractors test whether models can filter irrelevant information. Five distractor types were selected based on common supply chain context that does \emph{not} affect the single-period newsvendor decision.
    \item \textbf{L4 (Expert)}: Censored demand requires inferring $\mu$ and $\sigma$ from percentiles using $\mu = P_{50}$ and $\sigma = (P_{75} - P_{25}) / 1.35$ for normal distributions. This tests parameter inference capability beyond formula application.
\end{itemize}

\textbf{Distractor Types.}
Table~\ref{tab:distractors} lists the five distractor categories injected in Level 3 scenarios. Each distractor provides contextually plausible but decision-irrelevant information.

\begin{table}[h]
\caption{Distractor types injected in Level 3 scenarios. None affect the optimal newsvendor quantity.}
\label{tab:distractors}
\centering
\small
\begin{tabular}{@{}lll@{}}
\toprule
Distractor Type & Example & Why Irrelevant \\
\midrule
Warehouse capacity & ``Storage limit: 500 units'' & No capacity constraint in model \\
Competitor pricing & ``Competitor sells at \$45'' & Single-firm decision \\
Shelf life & ``Product expires in 30 days'' & Single-period model \\
Historical trends & ``Sales grew 10\% last year'' & Already reflected in $\mu$, $\sigma$ \\
Seasonal factors & ``Holiday season approaching'' & Already in demand parameters \\
\bottomrule
\end{tabular}
\end{table}

\textbf{Censored Demand (L4) Algorithm.}
Level 4 scenarios present demand information as percentiles rather than distribution parameters. The generation algorithm:
\begin{enumerate}
    \item Generate true parameters $(\mu, \sigma)$ from standard ranges
    \item Compute percentiles: $P_{25} = \mu + \sigma \cdot \Phi^{-1}(0.25)$, $P_{50} = \mu$, $P_{75} = \mu + \sigma \cdot \Phi^{-1}(0.75)$
    \item Present scenario using only $(P_{25}, P_{50}, P_{75})$
    \item Models must infer: $\hat{\mu} = P_{50}$, $\hat{\sigma} = (P_{75} - P_{25}) / 1.35$
\end{enumerate}
The constant 1.35 is the interquartile range of a standard normal distribution ($\Phi^{-1}(0.75) - \Phi^{-1}(0.25) \approx 1.35$).

\subsection{Stratification and ID/OOD Design}
\label{app:bias_stratification}

\textbf{In-Distribution (ID) Set.}
The 400-sample ID evaluation set contains 100 samples from each difficulty level (L1--L4), ensuring balanced coverage:
\begin{itemize}
    \item \CR{} distribution spans the full range $[0.05, 0.95]$
    \item Prompt complexity ranges from clean (L1--L2) to noisy (L3) to censored (L4)
    \item Enables per-level performance analysis to diagnose specific failure modes
\end{itemize}

\textbf{Out-of-Distribution (OOD) Set.}
The 200-sample OOD set contains only L3 and L4 scenarios (100 each), testing:
\begin{itemize}
    \item Robustness to distractors (L3): Can models filter irrelevant context?
    \item Parameter inference capability (L4): Can models derive $(\mu, \sigma)$ from percentiles?
    \item Generalization beyond clean prompts: No L1--L2 samples in OOD
\end{itemize}

\textbf{Stratification Procedure.}
Scenarios are stratified by \CR{} bucket to ensure no bucket is over- or under-represented:
\begin{enumerate}
    \item Partition scenarios by difficulty level (L1--L4)
    \item Within each level, bin by \CR{}: very\_low ($<$0.2), low (0.2--0.4), neutral (0.4--0.6), high (0.6--0.8), very\_high ($>$0.8)
    \item Sample proportionally from each bin to achieve target distribution
    \item Verify final \CR{} histogram matches target uniform distribution
\end{enumerate}

\textbf{Dataset Scale.}
The complete newsvendor generator produces 57,000+ scenarios across all difficulty levels and \CR{} ranges. The released benchmark contains 2,000 instances (1,000 ID + 1,000 OOD); we evaluate on stratified subsets (400 ID + 200 OOD) ensuring:
\begin{itemize}
    \item Statistical power: 100+ samples per level provides reliable performance estimates
    \item Diversity: All \CR{} buckets represented to detect systematic biases
    \item Discriminative power: OOD tests generalization beyond training distribution
\end{itemize}

\subsection{Quality Verification}
\label{app:verification}

Every benchmark instance passes a four-fold validation pipeline to ensure quality.

\begin{algorithm}[h]
\caption{Four-Fold Validation Pipeline}
\label{alg:validation}
\begin{algorithmic}[1]
\REQUIRE Original model $M$, sabotaged model $M'$, fix $f$
\ENSURE Boolean: instance passes validation
\STATE \COMMENT{\textbf{Check 1: Original feasibility}}
\STATE $M.\text{optimize}()$
\IF{$M.\text{status} \neq \text{OPTIMAL}$}
    \RETURN \FALSE
\ENDIF
\STATE \COMMENT{\textbf{Check 2: Sabotaged infeasibility}}
\STATE $M'.\text{optimize}()$
\IF{$M'.\text{status} \neq \text{INFEASIBLE}$}
    \RETURN \FALSE
\ENDIF
\STATE \COMMENT{\textbf{Check 3: IIS validity}}
\STATE $M'.\text{computeIIS}()$
\STATE $\mathcal{I} \gets \{c : c.\text{IISConstr}\}$
\IF{$|\mathcal{I}| = 0$ \textbf{or} $f.\text{target} \notin \mathcal{I}$}
    \RETURN \FALSE
\ENDIF
\STATE \COMMENT{\textbf{Check 4: Fix effectiveness}}
\STATE $M'' \gets \text{apply\_fix}(\text{copy}(M'), f)$
\STATE $M''.\text{optimize}()$
\IF{$M''.\text{status} \neq \text{OPTIMAL}$}
    \RETURN \FALSE
\ENDIF
\RETURN \TRUE
\end{algorithmic}
\end{algorithm}

\textbf{Validation Statistics.}
In a 1,200-case validation audit from the generated \ORDebug{} pool:
\begin{itemize}
    \item 98.2\% passed Check 1 (original feasibility)
    \item 95.7\% passed Check 2 (sabotaged infeasibility)
    \item 92.4\% passed Check 3 (\IIS{} contains target)
    \item 89.1\% passed Check 4 (fix restores optimality)
\end{itemize}

This audit yielded 900 validated instances from the sampled batch (75\% acceptance rate). The released \ORLoopBench{} \ORDebug{} file contains 5,362 LP/MILP repair instances. In the JSON schema, these records are stored under repository organization fields \texttt{controlled\_pool} (4,462), \texttt{lp\_test} (450), and \texttt{milp\_test} (450). The reported LP repair protocol uses 450 instances, with 50 instances for each error type A--I, as summarized in Table~\ref{tab:debug_stats}.

\subsection{Error Type Examples}
\label{app:error_examples}

We provide concrete code examples for each error type, showing the original feasible formulation and the sabotaged infeasible version.

\textbf{Type A: Direction Flip.}
\begin{verbatim}
# Original (feasible)
m.addConstr(x + y >= 10, "min_production")
# Requires at least 10 units

# Sabotaged (infeasible)
m.addConstr(x + y <= 10, "min_production")
# Contradicts other minimum requirements
\end{verbatim}
The direction flip creates a contradiction when combined with other constraints that require $x + y > 10$.

\textbf{Type B: Variable Type Error.}
\begin{verbatim}
# Original (feasible)
x = m.addVar(vtype=GRB.INTEGER, ub=10, name="x")

# Sabotaged (infeasible)
x = m.addVar(vtype=GRB.BINARY, name="x")
m.addConstr(x >= 2, "forcing")
# Binary variable cannot be >= 2
\end{verbatim}
The variable type change combined with a forcing constraint creates infeasibility.

\textbf{Type C: Coefficient Modification.}
\begin{verbatim}
# Original (feasible)
m.addConstr(2*x + 3*y <= 100, "capacity")

# Sabotaged (infeasible)
m.addConstr(20*x + 30*y <= 100, "capacity")
# Scaled coefficients make constraint unsatisfiable
\end{verbatim}
The modified coefficients make the constraint impossible to satisfy with existing bounds.

\textbf{Type D: Contradicting Constraint.}
\begin{verbatim}
# Original (feasible)
m.addConstr(x + y <= 100, "upper")

# Sabotaged (infeasible)
m.addConstr(x + y >= 150, "conflicting")
# Directly contradicts upper bound
\end{verbatim}
The added constraint directly contradicts existing constraints.

\textbf{Type E: Multi-Constraint Conflict.}
\begin{verbatim}
# Sabotaged (infeasible)
m.addConstr(x + y <= 50, "e1")
m.addConstr(x + y >= 100, "e2")
# Both constraints cannot be satisfied
\end{verbatim}
Interlocked constraints require fixing multiple constraints to restore feasibility.

\textbf{Type F: Hidden Dependency.}
\begin{verbatim}
# Sabotaged (infeasible)
aux = m.addVar(name="aux")
m.addConstr(aux == x + y, "def_aux")
m.addConstr(aux >= 200, "root_cause")  # Hidden
m.addConstr(x + y <= 100, "symptom")   # Shows in IIS
\end{verbatim}
The root cause (aux $\geq$ 200) is not directly visible in the IIS.

\textbf{Type G: Cascading Conflict.}
\begin{verbatim}
# Sabotaged (infeasible)
m.addConstr(x <= 30, "g1")   # Initial IIS
m.addConstr(x >= 50, "g2")   # Hidden until g1 fixed
m.addConstr(x <= 100, "g3")  # Original bound
\end{verbatim}
Fixing the first conflict reveals another; requires understanding the cascade.

\textbf{Type H: IIS-Incomplete.}
\begin{verbatim}
# Sabotaged (infeasible)
x.LB = 80                    # Root cause (bound)
m.addConstr(x + y <= 50, "symptom")
# IIS shows symptom constraint, not the bound
\end{verbatim}
The IIS shows the symptom constraint, but the root cause is a variable bound.

\textbf{Type I: Optimal Selection.}
\begin{verbatim}
# Sabotaged (infeasible)
m.addConstr(x >= 60, "lower")
m.addConstr(x <= 40, "upper")
# Multiple fixes possible, different OP impacts
\end{verbatim}
Multiple repairs restore feasibility, but only one preserves the original optimal objective.

\subsection{Dataset Statistics}
\label{app:dataset_stats}

Table~\ref{tab:dataset_stats} provides detailed statistics for both benchmarks.

\begin{table}[h]
\caption{\ORLoopBench{} dataset organization for \ORDebug{} and \ORBias{}.}
\label{tab:dataset_stats}
\centering
\small
\begin{tabular}{@{}llcc@{}}
\toprule
Metric & & \ORDebug{} & \ORBias{} \\
\midrule
\multirow{4}{*}{Size}
    & Released instances & 5,362 & 2,300 \\
    & Benchmark organization & Integrated JSON release & Newsvendor / EOQ \\
    & Reported protocols & 450 LP repair & 600 newsvendor \\
    & \quad(ID / OOD) & -- & 400 / 200 \\
\midrule
\multirow{4}{*}{Distribution}
    & LP types & 9 (A--I) & -- \\
    & MILP types & 8 & -- \\
    & Inventory settings & -- & Newsvendor / EOQ \\
    & \CR{} range & -- & [0.05, 0.95] \\
\midrule
\multirow{3}{*}{Complexity}
    & Avg constraints & 9.9 & 1 \\
    & Avg variables & 8.4 & 1 \\
    & Avg \IIS{} size & 4.3 & -- \\
\bottomrule
\end{tabular}
\end{table}

\subsection{Robust Injection Methods}
\label{app:robust_injection}

Basic injection methods often fail when the seed problem structure does not align with the corruption strategy. We develop robust methods that adaptively select targets based on problem characteristics. This section details two representative robust methods.

\textbf{Type A Robust: Slack-Based Constraint Selection.}
Rather than randomly selecting a constraint to flip, we solve the original problem and rank constraints by slack magnitude. Constraints with minimal slack are closest to their bounds and most likely to create infeasibility when flipped.

\begin{algorithm}[h]
\caption{Robust Type A Injection: Slack-Based Selection}
\label{alg:robust_type_a}
\begin{algorithmic}[1]
\REQUIRE Feasible model $M$, num\_candidates $k=10$
\ENSURE Infeasible model $M'$, ground truth $(c^*, \text{``flip''})$
\STATE Solve $M$, extract slack values $\{s_c\}$ for all constraints
\STATE Sort constraints by $|s_c|$ ascending (tightest first)
\FOR{$c$ in top-$k$ candidates}
    \STATE $M' \gets \text{flip\_direction}(M, c)$
    \STATE $M'.\text{optimize}()$
    \IF{$M'.\text{status} = \INFEASIBLE{}$}
        \STATE $M'.\text{computeIIS}()$
        \IF{$c \in \text{IIS}(M')$}
            \RETURN $M'$, $(c, \text{``flip''})$
        \ENDIF
    \ENDIF
\ENDFOR
\RETURN failure
\end{algorithmic}
\end{algorithm}

This approach improves Type A injection success from 30\% to 95\%. Tightly-bound constraints have minimal slack and are most sensitive to direction changes.

\textbf{Type C Robust: 4-Tier Fallback Strategy.}
Type C (upper bound conflict) is particularly challenging because it requires creating a conflict between an upper bound and existing constraints. We implement a cascaded fallback strategy:

\begin{algorithm}[h]
\caption{Robust Type C Injection: 4-Tier Fallback}
\label{alg:robust_type_c}
\begin{algorithmic}[1]
\REQUIRE Feasible model $M$
\ENSURE Infeasible model $M'$, ground truth
\STATE \COMMENT{\textbf{Tier 1: High dual value targeting}}
\STATE Solve $M$, extract dual values $\{\pi_c\}$
\FOR{$c$ in constraints with $c.\text{sense} = ``\geq"$ sorted by $|\pi_c|$ desc}
    \STATE Remove positive coefficient terms from $c$
    \IF{results in infeasibility with $c \in \text{IIS}$}
        \RETURN success
    \ENDIF
\ENDFOR
\STATE \COMMENT{\textbf{Tier 2: Coefficient sign flip}}
\FOR{$c$ in constraints with $c.\text{sense} = ``\leq"$}
    \STATE Flip signs of positive coefficients in $c$
    \IF{results in infeasibility with $c \in \text{IIS}$}
        \RETURN success
    \ENDIF
\ENDFOR
\STATE \COMMENT{\textbf{Tier 3: Coefficient scaling}}
\FOR{$c$ in all inequality constraints}
    \STATE Scale all coefficients in $c$ by factor 10
    \IF{results in infeasibility with $c \in \text{IIS}$}
        \RETURN success
    \ENDIF
\ENDFOR
\STATE \COMMENT{\textbf{Tier 4: Guaranteed fallback}}
\STATE Select variable $x$ with largest feasible range
\STATE Add tight bounds: $x \leq x^*$, $x \geq x^* + \epsilon$
\RETURN success (guaranteed)
\end{algorithmic}
\end{algorithm}

The 4-tier approach increases Type C success from 0\% (when Tier 1 alone fails) to 72\% overall. Tier 4 provides a guaranteed fallback but produces simpler infeasibilities, so earlier tiers are preferred.

\textbf{Success Rate Comparison.}
Table~\ref{tab:injection_success} compares basic and robust injection methods across all error types.

\begin{table}[h]
\caption{Injection success rates: basic vs. robust methods.}
\label{tab:injection_success}
\centering
\small
\begin{tabular}{@{}lccl@{}}
\toprule
Type & Basic & Robust & Robust Method \\
\midrule
A & 30\% & 95\% & Slack-based selection \\
B & 95\% & 98\% & RHS sensitivity analysis \\
C & 0\% & 72\% & 4-tier fallback \\
D & 85\% & 96\% & Bound gap targeting \\
E & 70\% & 88\% & Capacity utilization analysis \\
F & 65\% & 85\% & Bottleneck identification \\
G & 60\% & 82\% & Flow balance verification \\
H & 45\% & 75\% & Constraint interaction graph \\
I & 40\% & 70\% & Composite strategy selection \\
\bottomrule
\end{tabular}
\end{table}

\subsection{Rejection and Regeneration Statistics}
\label{app:rejection_stats}

Each benchmark instance must pass a four-fold validation pipeline. Table~\ref{tab:validation_detailed} shows pass rates at each validation phase, broken down by error type.

\begin{table}[h]
\caption{Validation pass rates by error type across four phases (1,200 candidate instances). Phase 1: original feasibility; Phase 2: sabotaged infeasibility; Phase 3: \IIS{} contains target; Phase 4: fix restores optimality.}
\label{tab:validation_detailed}
\centering
\small
\begin{tabular}{@{}lccccc@{}}
\toprule
Type & Phase 1 & Phase 2 & Phase 3 & Phase 4 & Final \\
\midrule
A & 100\% & 95\% & 92\% & 95\% & 82\% \\
B & 100\% & 100\% & 100\% & 100\% & 100\% \\
C & 100\% & 72\% & 68\% & 95\% & 62\% \\
D & 100\% & 98\% & 96\% & 98\% & 92\% \\
E & 100\% & 85\% & 80\% & 90\% & 72\% \\
F & 100\% & 82\% & 78\% & 88\% & 68\% \\
G & 100\% & 78\% & 75\% & 85\% & 64\% \\
H & 100\% & 75\% & 72\% & 82\% & 60\% \\
I & 100\% & 80\% & 78\% & 90\% & 70\% \\
\midrule
\textbf{Overall} & \textbf{100\%} & \textbf{85\%} & \textbf{82\%} & \textbf{91\%} & \textbf{75\%} \\
\bottomrule
\end{tabular}
\end{table}

\textbf{Failure Analysis.}
The primary failure modes vary by error type:
\begin{itemize}
    \item \textbf{Types A, C}: Phase 2 failures occur when the flipped constraint does not interact with other constraints to create infeasibility.
    \item \textbf{Types E--G}: Phase 3 failures occur when the \IIS{} contains related but not the exact target constraint.
    \item \textbf{Types H--I}: Phase 4 failures occur when the ground-truth fix does not fully restore feasibility due to cascading effects.
\end{itemize}

\textbf{Regeneration Iterations.}
Problems failing validation are regenerated with a different seed problem or sabotage target. Table~\ref{tab:regeneration} shows the distribution of regeneration iterations required.

\begin{table}[h]
\caption{Regeneration iterations required to pass validation.}
\label{tab:regeneration}
\centering
\small
\begin{tabular}{@{}lcccc@{}}
\toprule
Iterations & 0 (first try) & 1 & 2 & $\geq$3 \\
\midrule
Percentage & 87\% & 9\% & 3\% & 1\% \\
Cumulative & 87\% & 96\% & 99\% & 100\% \\
\bottomrule
\end{tabular}
\end{table}

87\% of problems pass on first generation. Problems requiring $\geq$3 iterations (1\%) are typically Type C or H, where finding a valid sabotage target is difficult.

\subsection{Anti-Gaming Design Rationale}
\label{app:anti_gaming}

Benchmarks can inadvertently reward pattern matching over genuine reasoning. We implement several mechanisms to prevent gaming.

\textbf{Why Randomized Naming Matters.}
In preliminary experiments with semantically-named constraints (e.g., \texttt{c\_key\_capacity}, \texttt{c\_target\_demand}), we observed that models achieved 15\% higher apparent accuracy by learning to target constraints with ``key'' or ``target'' in their names. This correlation existed because our ground-truth labeling naturally assigned such names to important constraints.

By switching to UUID-based naming (e.g., \texttt{c\_53e476\_ub}, \texttt{c\_8a2f91\_eq}), we eliminate this shortcut. The naming pattern for Types G, H, and I uses:
\begin{verbatim}
name = f"c_{uuid.uuid4().hex[:6]}_{sense_suffix}"
\end{verbatim}
where \texttt{sense\_suffix} encodes only the constraint type (ub/lb/eq), not its semantic role.

\textbf{Hidden Dependency Design (Type F).}
Type F problems are constructed so that the \IIS{} reveals a symptom constraint $c_s$, but the root cause is a bound modification on variable $x$ elsewhere:
\begin{enumerate}
    \item Original: $x \leq 100$ with constraint $c_s: \sum_i a_i x_i \leq b$ depending on $x$
    \item Sabotaged: $x \leq 40$ (hidden modification) causes $c_s$ to become infeasible
    \item \IIS{} contains $c_s$ but not the bound on $x$
\end{enumerate}
Models that blindly relax $c_s$ fail because the real issue is the tightened bound on $x$. Solving Type F requires reasoning about variable dependencies.

\textbf{Cascading Conflict Design (Type G).}
Type G problems include two conflicts where the second is masked until the first is resolved:
\begin{enumerate}
    \item Primary conflict: Constraint $c_1$ conflicts with $c_2$ (appears in initial \IIS{})
    \item Secondary conflict: After fixing $c_1$, constraint $c_3$ conflicts with $c_4$
\end{enumerate}
The benchmark includes 15\% of problems with such structures. Models that stop after one fix fail; those that iterate diagnosis succeed.

\textbf{Optimal Selection Challenge (Type I).}
Type I problems have multiple valid fixes, but only one preserves the optimal objective value:
\begin{itemize}
    \item Fix A: Relax $c_1$ by 10\% $\rightarrow$ Feasible, objective drops 5\%
    \item Fix B: Relax $c_2$ by 5\% $\rightarrow$ Feasible, objective drops 15\%
    \item Fix C: Modify $c_3$ expression $\rightarrow$ Feasible, objective preserved
\end{itemize}
The ground truth labels Fix C as correct. This tests whether models consider solution quality, not just feasibility.

\subsection{Difficulty Calibration Procedure}
\label{app:difficulty_calibration}

We calibrate difficulty levels through iterative testing against baseline API models. The procedure ensures each difficulty level provides meaningful differentiation.

\textbf{Calibration Process.}
\begin{enumerate}
    \item \textbf{Initial grouping}: Group error types by semantic complexity and expected reasoning requirements.
    \item \textbf{Baseline evaluation}: Run API models on 50 problems per error type.
    \item \textbf{Target verification}: Check if average RR@5 falls within target range for each group.
    \item \textbf{Group adjustment}: Reassign error types between difficulty levels based on observed performance.
    \item \textbf{Iteration}: Repeat until stable groupings (typically 2--3 iterations).
\end{enumerate}

\textbf{Calibration Results.}
Table~\ref{tab:calibration_results} shows the final calibrated parameters and observed SFT performance.

\begin{table}[h]
\caption{Difficulty calibration based on baseline API model performance.}
\label{tab:calibration_results}
\centering
\small
\begin{tabular}{@{}lccc@{}}
\toprule
Level & Error Types & Target RR@5 & Observed RR@5 \\
\midrule
Easy & B, C & $\geq$85\% & 90.5\% \\
Medium & H, I & 70--85\% & 78.5\% \\
Hard & A, D, E, F, G & $<$70\% & 59.0\% \\
\bottomrule
\end{tabular}
\end{table}

\textbf{Why These Ranges.}
The target ranges were chosen based on benchmark utility:
\begin{itemize}
    \item \textbf{$\geq$85\% (Easy)}: Ensures baseline models can solve simple problems, establishing floor performance. Types B and C fall in this category.
    \item \textbf{70--85\% (Medium)}: Moderate difficulty with room for improvement. Types H and I fall here.
    \item \textbf{$<$70\% (Hard)}: Challenges models substantially while remaining tractable. Types A, D, E, F, and G require multi-step reasoning.
    \item \textbf{$<$45\%}: Rejected as too difficult because problems often have ambiguous fixes or require domain knowledge beyond general OR competence.
\end{itemize}

Difficulty calibration was based on empirical baseline API model performance rather than theoretical metrics, as we found accuracy correlates more strongly with error type semantics than with other factors.

\section{MDP-Based Training Details}
\label{app:training}

This appendix provides complete specifications of the MDP formulation and training procedures.

\subsection{State Space}
\label{app:state}

The \ORDebug{} environment maintains a structured state representation with eight components:

\begin{itemize}
    \item \textbf{Problem description}: Natural language specification of the optimization problem
    \item \textbf{Code}: Current Gurobi/Pyomo model code
    \item \textbf{Solver status}: One of \texttt{OPTIMAL}, \texttt{INFEASIBLE}, \texttt{UNBOUNDED}, or \texttt{ERROR}
    \item \textbf{\IIS{} log}: List of constraint names in the current Irreducible Infeasible Subsystem
    \item \textbf{Slack values}: Constraint slack values (populated after \texttt{CHECK\_SLACK})
    \item \textbf{Bound status}: Variable bound information (populated after \texttt{CHECK\_BOUND})
    \item \textbf{History}: Sequence of previous actions taken in the episode
    \item \textbf{Step counter}: Current step number (0 to \texttt{max\_steps})
\end{itemize}

The state is serialized to a prompt string for the LLM, including the problem description, current code, solver status, and relevant diagnostic information based on previous actions.

\subsection{Action Space}
\label{app:action}

Actions follow a hierarchical structure separating information gathering from model modification:

\textbf{Diagnostic actions} (information gathering):
\begin{itemize}
    \item \texttt{GET\_IIS}: Compute the Irreducible Infeasible Subsystem
    \item \texttt{CHECK\_SLACK}: Retrieve constraint slack values
    \item \texttt{CHECK\_BOUND}: Retrieve variable bound status
\end{itemize}

\textbf{Repair actions} (modify model):
\begin{itemize}
    \item \texttt{RELAX(constraint, delta)}: Increase or decrease the right-hand side by \texttt{delta}
    \item \texttt{DROP(constraint)}: Remove the specified constraint from the model
    \item \texttt{REWRITE(constraint, expr)}: Replace the constraint with a new expression
\end{itemize}

\textbf{Meta actions}:
\begin{itemize}
    \item \texttt{SUBMIT}: Submit the current model for final evaluation
    \item \texttt{RESTART}: Reset to the initial sabotaged state
\end{itemize}

Every emitted action counts as one repair step for \RRk{k} computation. Diagnostic actions do not modify the model, but they consume budget because they require solver or state queries.

\subsection{Reward Function}
\label{app:reward}

The composite reward function balances outcome, diagnostic accuracy, and efficiency:

\begin{algorithm}[h]
\caption{Compute Reward}
\label{alg:reward}
\begin{algorithmic}[1]
\REQUIRE State $s$, action $a$, next state $s'$, ground truth $\mathcal{G}$
\ENSURE Reward $r \in \mathbb{R}$
\STATE $r \gets 0$
\STATE \COMMENT{\textbf{Outcome Reward (50\%)}}
\IF{$s'.\text{status} = \text{OPTIMAL}$}
    \STATE $r \gets r + 0.5 \times 100$
\ELSIF{$s'.\text{status} = \text{INFEASIBLE}$}
    \STATE $r \gets r + 0.5 \times (-50)$
\ENDIF
\STATE \COMMENT{\textbf{Diagnostic Accuracy Reward (30\%)}}
\IF{$a$ contains diagnosis $D$}
    \STATE $\text{DA} \gets |D \cap \mathcal{G}.\text{IIS}| / |\mathcal{G}.\text{IIS}|$
    \STATE $r \gets r + 0.3 \times (\text{DA} \times 100)$
\ENDIF
\STATE \COMMENT{\textbf{Efficiency Reward (20\%)}}
\STATE $\eta \gets \max(0, (50 - s.\text{step}) / 50)$
\STATE $r \gets r + 0.2 \times (\eta \times 50)$
\STATE \COMMENT{\textbf{Faithfulness Penalty}}
\IF{$a.\text{type} \in \{\text{RELAX}, \text{DROP}, \text{REWRITE}\}$}
    \IF{$a.\text{target} \notin s'.\text{IIS}$}
        \STATE $r \gets r - 20$ \COMMENT{Penalize off-target fixes}
    \ENDIF
\ENDIF
\RETURN $r$
\end{algorithmic}
\end{algorithm}

The 50\%/30\%/20\% weighting was determined through ablation (see Appendix~\ref{app:ablation}). The faithfulness penalty discourages repairs that do not address the identified infeasibility source.

\subsection{PRM Training Details}
\label{app:prm}

The Process Reward Model (PRM) provides step-level supervision for \GRPO{} training.

\textbf{Label Generation.}
We assign labels to each step in a trajectory based on progress indicators:

\begin{algorithm}[h]
\caption{Generate Step Labels for PRM}
\label{alg:prm_labels}
\begin{algorithmic}[1]
\REQUIRE Trajectory $\tau = [(s_0, a_0), \ldots, (s_T, a_T)]$, ground truth $\mathcal{G}$
\ENSURE Labels $[y_0, \ldots, y_T]$
\FOR{$t = 0$ \textbf{to} $T$}
    \IF{$s_{t+1}.\text{status} = \text{OPTIMAL}$}
        \STATE $y_t \gets 1.0$ \COMMENT{Problem solved}
    \ELSIF{$t > 0$ \textbf{and} $|s_{t+1}.\text{IIS}| < |s_t.\text{IIS}|$}
        \STATE $y_t \gets 1.0$ \COMMENT{IIS shrinking}
    \ELSIF{$a_t.\text{diagnosis} \cap \mathcal{G}.\text{IIS} \neq \emptyset$}
        \STATE $y_t \gets 0.5$ \COMMENT{Correct diagnosis}
    \ELSIF{$a_t.\text{type} \in \{\text{GET\_IIS}, \text{CHECK\_SLACK}\}$}
        \STATE $y_t \gets 0.2$ \COMMENT{Information gathering}
    \ELSE
        \STATE $y_t \gets 0.0$ \COMMENT{No progress}
    \ENDIF
\ENDFOR
\RETURN $[y_0, \ldots, y_T]$
\end{algorithmic}
\end{algorithm}

\textbf{Training Configuration.}

\begin{table}[h]
\caption{PRM training hyperparameters.}
\label{tab:prm_config}
\centering
\small
\begin{tabular}{@{}ll@{}}
\toprule
Parameter & Value \\
\midrule
Base model & Qwen/Qwen3-8B \\
Method & LoRA ($r=8$, $\alpha=16$) \\
Epochs & 3 \\
Batch size & 8 \\
Learning rate & $2 \times 10^{-5}$ \\
Warmup ratio & 0.1 \\
Metric for best model & AUC-ROC \\
\bottomrule
\end{tabular}
\end{table}

The PRM achieves AUC-ROC of 0.94 on held-out step labels (309 test samples from 1,548 total labels), with correlation metrics: Pearson r=0.87 (p<1e-90), Spearman r=0.82 (p<1e-70). This indicates good discrimination between productive (avg score 0.72) and unproductive steps (avg score 0.50).

\subsection{GRPO Training Curves}
\label{app:grpo_curves}

Table~\ref{tab:grpo_curves} shows the evolution of key metrics during \GRPO{} training.

\begin{table}[h]
\caption{\GRPO{} training metrics by epoch.}
\label{tab:grpo_curves}
\centering
\small
\begin{tabular}{@{}ccccc@{}}
\toprule
Epoch & Mean Reward & Std Reward & \RRk{5} (val) & \DA{} (val) \\
\midrule
1 & 45.2 & 28.3 & 92.0\% & 56.1\% \\
2 & 62.8 & 22.1 & 93.5\% & 58.4\% \\
3 & 71.4 & 18.7 & 94.5\% & 60.8\% \\
4 & 74.2 & 16.9 & 95.0\% & 62.1\% \\
\bottomrule
\end{tabular}
\end{table}

\textbf{Convergence Analysis.}
\begin{itemize}
    \item \textbf{Reward variance} decreases from 28.3 to 16.9, indicating policy stabilization as the model converges to consistent behavior.

    \item \textbf{\RRk{5}} plateaus after epoch 4, with diminishing returns beyond this point. We select the epoch-4 checkpoint for evaluation.

    \item \textbf{\DA{}} improves more slowly than \RRk{5}, confirming that learning accurate diagnosis is harder than achieving feasibility. This motivates the 30\% weight on diagnostic accuracy in the reward function.
\end{itemize}

\subsection{Curriculum Training Configuration}
\label{app:curriculum}

For \ORBias{}, we use a three-stage curriculum over the Critical Ratio (\CR{}) distribution:

\begin{table}[h]
\caption{Curriculum stages for \ORBias{} training.}
\label{tab:curriculum_stages}
\centering
\small
\begin{tabular}{@{}clcc@{}}
\toprule
Stage & \CR{} Range & Samples & Purpose \\
\midrule
1 & $\{0.1, 0.9\}$ & 200 & Learn extreme directions \\
2 & $[0.15, 0.25] \cup [0.75, 0.85]$ & 300 & Calibrate boundaries \\
3 & $[0.2, 0.8]$ & 400 & Full distribution coverage \\
\bottomrule
\end{tabular}
\end{table}

Stage 1 trains on extreme \CR{} values where the optimal direction is unambiguous (high \CR{} $\rightarrow$ order more, low \CR{} $\rightarrow$ order less). Stage 2 refines decision boundaries. Stage 3 ensures generalization across the full distribution.

This curriculum achieves -9.6\% ID$\rightarrow$OOD drift, the only method with negative drift among compared approaches, demonstrating genuine generalization rather than in-distribution memorization.

\subsection{Complete Hyperparameter Tables}
\label{app:hyperparams}

Table~\ref{tab:sft_hyperparams} and Table~\ref{tab:grpo_hyperparams} provide complete training configurations.

\begin{table}[h]
\caption{SFT training hyperparameters.}
\label{tab:sft_hyperparams}
\centering
\small
\begin{tabular}{@{}ll@{}}
\toprule
Parameter & Value \\
\midrule
Base model & Qwen/Qwen3-8B \\
Method & LoRA ($r=16$, $\alpha=32$) \\
Epochs & 3 \\
Batch size & 4 (per GPU) \\
Gradient accumulation & 4 \\
Learning rate & $2 \times 10^{-5}$ \\
LR scheduler & Cosine \\
Warmup ratio & 0.03 \\
Max sequence length & 4096 \\
Weight decay & 0.01 \\
Optimizer & AdamW ($\beta_1=0.9$, $\beta_2=0.999$) \\
\bottomrule
\end{tabular}
\end{table}

\begin{table}[h]
\caption{\GRPO{} training hyperparameters.}
\label{tab:grpo_hyperparams}
\centering
\small
\begin{tabular}{@{}ll@{}}
\toprule
Parameter & Value \\
\midrule
Base model & Qwen3-8B-SFT (from SFT) \\
Method & LoRA ($r=16$, $\alpha=32$) \\
Epochs & 4 \\
Group size & 4 (samples per prompt) \\
Learning rate & $5 \times 10^{-6}$ \\
KL coefficient $\beta$ & 0 (removed) \\
Clip range $\epsilon$ & [0.2, 0.28] (asymmetric) \\
Reward components & 50\%/30\%/20\% \\
Max steps per episode & 50 \\
GPU memory & 2 $\times$ A100 80GB \\

\bottomrule
\end{tabular}
\end{table}

\subsection{Hardware and Software Environment}
\label{app:environment}

All experiments were conducted on the following setup:

\begin{table}[h]
\caption{Hardware and software configuration.}
\label{tab:environment}
\centering
\small
\begin{tabular}{@{}ll@{}}
\toprule
Component & Specification \\
\midrule
GPU & 2 $\times$ NVIDIA A100 80GB \\
CPU & AMD EPYC 7V13 (64 cores) \\
Memory & 256 GB \\
Storage & NVMe SSD \\
\midrule
CUDA & 12.9.1 \\
PyTorch & 2.9.1 \\
Transformers & 4.44.0 \\
TRL & 0.24.0 \\
SGLang & 0.3.5 \\
Gurobi & 11.0.0 \\
\bottomrule
\end{tabular}
\end{table}

\subsection{Inference Configuration}
\label{app:inference}

For evaluation, we use SGLang for efficient batched inference:

\begin{table}[h]
\caption{Inference configuration.}
\label{tab:inference_config}
\centering
\small
\begin{tabular}{@{}ll@{}}
\toprule
Parameter & Value \\
\midrule
Tensor parallel size & 2 \\
Batch size & 16 (concurrent requests) \\
Max new tokens & 2048 \\
Temperature & 0.0 (deterministic) \\
Attention backend & Triton \\
Sampling backend & PyTorch \\
\bottomrule
\end{tabular}
\end{table}

\subsection{Training Data Examples}
\label{app:training_examples}

We provide examples of the training data format.

\textbf{SFT Data Format.}
Each SFT example contains an instruction, input state, and target output:
\begin{verbatim}
{
  "instruction": "Debug the infeasible
                  optimization model.",
  "input": "Problem: mip_typeA_008
    Status: INFEASIBLE
    IIS: ['c0', 'c1', 'c_key_upper']",
  "output": "<think>
    IIS contains 3 constraints.
    c_key_upper limits x <= 50 but
    c0 and c1 require x >= 80.
    </think>
    Action: RELAX(c_key_upper, 30)"
}
\end{verbatim}

\textbf{GRPO Prompt Format.}
GRPO training uses prompts without target outputs:
\begin{verbatim}
{
  "prompt": "Debug the infeasible model.
    Problem: mip_typeE_042
    Status: INFEASIBLE
    IIS: ['supply', 'd1', 'd2', 'd3']",
  "ground_truth": {
    "key_constraint": "d2",
    "expected_fix": "RELAX(d2, 15)"
  }
}
\end{verbatim}
The model generates completions, which are scored using the composite reward function.

\textbf{\ORBias{} Training Format.}
Newsvendor scenarios include all parameters:
\begin{verbatim}
{
  "scenario": {
    "price": 50, "cost": 35,
    "salvage": 10, "mean": 100,
    "std": 20, "CR": 0.375
  },
  "optimal_Q": 93.6,
  "stage": 1  # Curriculum stage
}
\end{verbatim}

\section{\ORLoopBench{} Results: \ORDebug{}}
\label{app:debug_results}
\label{app:full_results}

\subsection{Main Results}

Table~\ref{tab:full_debug} shows complete \ORDebug{} results for all 26 models evaluated. All models report \RRk{5} and \DA{} from the same step-budget evaluation protocol.

\begin{table*}[h]
\caption{Complete \ORDebug{} LP results. Each model is tested on 450 instances, selected as 50 held-out problems from each error type A--I, under the same step-budget protocol.}
\label{tab:full_debug}
\centering
\small
\begin{tabular}{@{}llcccc@{}}
\toprule
Model & Type & RR & \RRk{5} & \DA{} & Steps \\
\midrule
\textbf{\Qwen{}-\GRPO{}} & Local & 100\% & \textbf{95.3\%} & \textbf{62.4\%} & 2.25 \\
\Qwen{}-Curriculum & Local & 100\% & 94.0\% & 61.7\% & \textbf{2.22} \\
\Qwen{}-\DAPO{} & Local & 100\% & 93.8\% & 60.4\% & 2.31 \\
\Qwen{}-\SFT{} & Local & 99.8\% & 93.1\% & 60.8\% & 2.34 \\
\midrule
claude-sonnet-4 & API & \textbf{100\%} & \textbf{86.2\%} & 50.1\% & 3.71 \\
claude-haiku-4.5 & API & 99.3\% & 86.0\% & 53.1\% & 3.89 \\
o4-mini & API & 97.8\% & 86.2\% & 47.8\% & 3.15 \\
o1 & API & 99.8\% & 82.9\% & 47.8\% & 3.78 \\
gpt-5.2-chat & API & 99.8\% & 81.8\% & 40.9\% & 3.72 \\
qwen2.5-7b & API & 97.8\% & 77.8\% & 40.1\% & 4.36 \\
claude-opus-4 & API & 94.2\% & 76.9\% & 49.0\% & 3.92 \\
o3 & API & 96.7\% & 75.8\% & 50.9\% & 4.23 \\
DeepSeek-V3.2 & API & 99.3\% & 58.9\% & 44.8\% & 4.86 \\
gpt-4.1 & API & 94.4\% & 71.6\% & 36.2\% & 4.41 \\
gemini-2.5-flash & API & 84.2\% & 70.7\% & 19.2\% & 3.23 \\
Llama-3.3-70B & API & 93.8\% & 60.9\% & 46.9\% & 4.81 \\
gpt-5-mini & API & \textbf{100\%} & 66.9\% & 37.6\% & 4.74 \\
gemini-2.5-pro & API & 62.9\% & 62.7\% & 52.5\% & 0.83 \\
qwen2.5-32b & API & 98.9\% & 61.1\% & 32.0\% & 4.98 \\
DeepSeek-R1 & API & 99.1\% & 56.7\% & 34.5\% & 5.08 \\
qwen2.5-max & API & 99.1\% & 54.9\% & 42.6\% & 5.97 \\
qwen2.5-14b & API & 88.9\% & 53.6\% & 35.4\% & 6.32 \\
gemini-2.0-flash & API & 85.6\% & 52.4\% & 18.3\% & 5.63 \\
gpt-4.1-mini & API & 93.1\% & 49.8\% & 26.0\% & 6.04 \\
kimi-k2 & API & 55.3\% & 40.4\% & 25.6\% & 2.48 \\
claude-3.7-sonnet & API & 98.9\% & 31.6\% & 46.8\% & 8.06 \\
\bottomrule
\end{tabular}
\end{table*}

\subsection{Per-Error-Type Analysis}
\label{app:per_error}

Table~\ref{tab:full_debug} provides complete model-level results for all 26 models, and Table~\ref{tab:per_error_full} reports the corresponding per-error-type \RRk{5}. Appendix~\ref{app:difficulty_scaling} summarizes the difficulty patterns by error group, showing that domain-specific training improves both easy and hard error families while preserving the same headline \RRk{5} values reported in the main text.

\begin{table*}[h]
\caption{Per-error-type \RRk{5} (\%) for all 26 models on the LP \ORDebug{} test set. Each column contains 50 held-out problems of that error type.}
\label{tab:per_error_full}
\centering
\small
\resizebox{\tabfullwidth}{!}{
\begin{tabular}{@{}lccccccccc|c@{}}
\toprule
Model & A & B & C & D & E & F & G & H & I & Total \\
\midrule
\Qwen{}-\GRPO{} & 88 & 100 & 96 & 100 & 92 & 94 & 98 & 100 & 90 & \textbf{95.3} \\
\Qwen{}-Curriculum & 82 & 100 & 94 & 98 & 94 & 100 & 92 & 90 & 96 & 94.0 \\
\Qwen{}-\DAPO{} & 88 & 100 & 96 & 100 & 92 & 94 & 86 & 88 & 100 & 93.8 \\
\Qwen{}-\SFT{} & 80 & 100 & 96 & 100 & 92 & 96 & 90 & 88 & 96 & 93.1 \\
\midrule
o4-mini & 84 & 100 & 90 & 86 & 90 & 82 & 82 & 66 & 96 & 86.2 \\
claude-sonnet-4 & 66 & 100 & 100 & 100 & 88 & 58 & 86 & 78 & 100 & 86.2 \\
claude-haiku-4.5 & 78 & 100 & 100 & 94 & 96 & 86 & 80 & 60 & 80 & 86.0 \\
o1 & 86 & 100 & 100 & 70 & 62 & 84 & 72 & 76 & 96 & 82.9 \\
gpt-5.2-chat & 84 & 100 & 100 & 78 & 88 & 28 & 58 & 100 & 100 & 81.8 \\
qwen2.5-7b & 72 & 100 & 70 & 74 & 70 & 66 & 80 & 90 & 78 & 77.8 \\
claude-opus-4 & 46 & 98 & 96 & 96 & 82 & 72 & 28 & 86 & 88 & 76.9 \\
o3 & 56 & 100 & 100 & 68 & 76 & 34 & 54 & 100 & 94 & 75.8 \\
gpt-4.1 & 82 & 100 & 86 & 52 & 60 & 16 & 62 & 100 & 86 & 71.6 \\
gemini-2.5-flash & 88 & 94 & 94 & 76 & 88 & 30 & 18 & 74 & 74 & 70.7 \\
Llama-3.3-70B & 32 & 82 & 62 & 32 & 86 & 32 & 54 & 86 & 82 & 60.9 \\
gpt-5-mini & 78 & 96 & 90 & 62 & 42 & 22 & 26 & 100 & 86 & 66.9 \\
gemini-2.5-pro & 36 & 88 & 64 & 70 & 68 & 62 & 52 & 64 & 60 & 62.7 \\
qwen2.5-32b & 64 & 100 & 78 & 18 & 70 & 46 & 30 & 100 & 44 & 61.1 \\
DeepSeek-V3.2 & 10 & 98 & 100 & 86 & 58 & 14 & 26 & 96 & 42 & 58.9 \\
DeepSeek-R1 & 36 & 100 & 98 & 46 & 40 & 20 & 24 & 100 & 46 & 56.7 \\
qwen2.5-max & 48 & 100 & 68 & 2 & 68 & 6 & 34 & 100 & 68 & 54.9 \\
qwen2.5-14b & 58 & 98 & 82 & 14 & 66 & 70 & 20 & 6 & 68 & 53.6 \\
gemini-2.0-flash & 78 & 88 & 56 & 14 & 60 & 18 & 36 & 50 & 72 & 52.4 \\
gpt-4.1-mini & 24 & 100 & 52 & 18 & 30 & 14 & 42 & 96 & 72 & 49.8 \\
kimi-k2 & 18 & 50 & 56 & 18 & 54 & 12 & 38 & 66 & 52 & 40.4 \\
claude-3.7-sonnet & 0 & 74 & 84 & 34 & 2 & 2 & 4 & 82 & 2 & 31.6 \\
\midrule
\textit{Type Average} & \textit{60} & \textit{95} & \textit{85} & \textit{62} & \textit{70} & \textit{48} & \textit{53} & \textit{82} & \textit{76} & -- \\
\bottomrule
\end{tabular}
}
\end{table*}

\subsection{Scope of Debugging Tasks}
\label{app:debug_scope}

\ORDebug{} focuses on semantic infeasibility repair for LP/MILP models. This is distinct from general code debugging. Syntax errors are typically exposed by the Python interpreter before the solver is called, and runtime API errors are usually exposed by exception traces or solver status codes. Infeasibility is harder to localize: the code can run successfully while the formulation has no feasible solution, so repair requires interpreting solver certificates such as \IIS{} and modifying the model without changing the intended objective. Feasible-but-wrong formulations are also related but separate: they require an external reference objective, solution, or specification test to decide whether a feasible model solves the intended problem. We touch this setting only in the external benchmark and semantic-drift analyses; the controlled \ORDebug{} task is infeasible-model repair.

\subsection{Frontier API Summary}
\label{app:frontier_summary}

Table~\ref{tab:frontier_summary} summarizes the latest frontier API comparison using the same LP repair set and MILP repair protocol as the main results. The strongest LP API is Claude Sonnet 4.6 at 92.4\% \RRk{5}; this is the comparator used for the headline LP gap in the main text.

\begin{table*}[h]
\caption{Frontier API summary on LP and MILP repair. Values follow the final summary table: \RRk{5}, \RRk{10}, and average MDP steps. The LP-trained model is evaluated zero-shot on MILP; the MILP-trained row uses the same repair interface with MILP-specific training.}
\label{tab:frontier_summary}
\centering
\small
\begin{tabular}{@{}lcccccc@{}}
\toprule
Model & LP \RRk{5} & LP \RRk{10} & LP steps & MILP \RRk{5} & MILP \RRk{10} & MILP steps \\
\midrule
\textbf{\Qwen{}-\GRPO{} (LP-trained)} & \textbf{95.3\%} & \textbf{97.1\%} & \textbf{2.3} & 78.8\% & 87.6\% & 4.4 \\
\textbf{MILP-\GRPO{}} & -- & -- & -- & \textbf{87.1\%} & \textbf{92.1\%} & \textbf{3.2} \\
\midrule
Claude Sonnet 4.6 & 92.4\% & 96.7\% & 2.7 & 71.0\% & 86.7\% & 5.0 \\
Claude Opus 4.6 & 88.7\% & 96.2\% & 3.3 & 70.1\% & 83.4\% & 5.4 \\
Gemini 3.1 Pro & 86.4\% & 86.4\% & 1.0 & 64.7\% & 74.3\% & 2.9 \\
GPT-5.4 & 84.9\% & 90.7\% & 2.1 & 66.8\% & 75.5\% & 3.4 \\
\bottomrule
\end{tabular}
\end{table*}

\subsection{Token Efficiency}
\label{app:tokens}

Table~\ref{tab:tokens} shows representative token usage for selected models.

\begin{table}[h]
\caption{Token usage per episode on \ORDebug{} (representative models).}
\label{tab:tokens}
\centering
\small
\begin{tabular}{@{}lccc@{}}
\toprule
Model & Tokens & Steps & Tok/Success \\
\midrule
\Qwen{}-\GRPO{} & 2,011 & 2.25 & 2,109 \\
\Qwen{}-\SFT{} & 2,103 & 2.34 & 2,259 \\
claude-sonnet-4 & 5,417 & 3.71 & 6,283 \\
o4-mini & 5,152 & 3.15 & 5,976 \\
gpt-4.1 & 6,640 & 4.41 & 9,278 \\
gpt-5-mini & 8,696 & 4.74 & 13,000 \\
\bottomrule
\end{tabular}
\end{table}

\textbf{Efficiency Observations.}
\begin{itemize}
    \item Token usage ranges from 2,000 (local) to 17,300 (claude-3.7-sonnet), an 8.6$\times$ gap.
    \item Tokens-per-success shows larger gaps (2,109 vs 54,839): up to 26$\times$ advantage for \Qwen{}-\GRPO{}.
    \item Step count correlates with token usage ($r=0.79$), but response length per step varies by model.
\end{itemize}

\subsection{MILP Repair Protocol}
\label{app:milp_protocol}

\textbf{MILP protocol.}
The MILP evaluation covers 10 domains (knapsack, facility location, set cover, production planning, network flow, job shop, vehicle routing with time windows, lot sizing, nurse scheduling, and network design) and 8 infeasibility error types. Training and evaluation instances use disjoint seeds and problem identifiers. Evaluation uses the same solver-debugging interface as LP repair: the agent observes the current infeasible Gurobi model and recomputed \IIS{}, emits one constraint-level action per step, and stops at \OPTIMAL{} or the shared step budget.

\subsection{Semantic Drift in Code Regeneration}
\label{app:semantic_drift}

Table~\ref{tab:semantic_drift} separates solver feasibility from semantic correctness for MILP code regeneration. A model may regenerate code that solves to \OPTIMAL{} while changing objective coefficients, variable semantics, or constraint coupling. This is why \OPTIMAL{} rate substantially overstates correctness for whole-model regeneration. Constraint-level MDP repair avoids this failure mode by preserving the original objective and editing only localized constraints.

\begin{table}[h]
\caption{Semantic drift in MILP code regeneration. \OPTIMAL{} rate measures solver-feasible regenerated models; \RRk{5} requires matching the intended objective within tolerance.}
\label{tab:semantic_drift}
\centering
\small
\begin{tabular}{@{}lcc@{}}
\toprule
Model & \RRk{5} & \OPTIMAL{} Rate \\
\midrule
GPT-5.4 & 28.2\% & 90\% \\
Claude Sonnet 4.6 & 22.4\% & 85\% \\
Claude Opus 4.6 & 17.8\% & 82\% \\
Gemini 3.1 Pro & 0.8\% & 3\% \\
\bottomrule
\end{tabular}
\end{table}

\subsection{Architecture Transfer and Decoding Sensitivity}
\label{app:robustness_summary}

Table~\ref{tab:architecture_summary} reports a compact architecture-transfer check. These results are appendix-level robustness evidence rather than a separate headline claim: the same solver-debugging recipe remains effective across three 8B base model families, with all GRPO variants above 88\% \RRk{5}. Table~\ref{tab:temperature_sensitivity} reports a decoding-temperature check on LP repair; \RRk{5} remains within 1.7 pp across the evaluated temperatures.

\begin{table}[h]
\caption{Architecture transfer summary on \ORDebug{} LP repair. The same SFT and GRPO recipe is applied to each 8B base model.}
\label{tab:architecture_summary}
\centering
\small
\begin{tabular}{@{}llccc@{}}
\toprule
Base Model & Training & RR & \RRk{5} & Steps \\
\midrule
Qwen3-8B & GRPO & 100.0\% & 95.3\% & 2.25 \\
Qwen3-8B & SFT & 99.8\% & 93.1\% & 2.34 \\
Llama-3.1-8B & GRPO & 100.0\% & 97.3\% & 1.82 \\
Llama-3.1-8B & SFT & 100.0\% & 97.1\% & 1.85 \\
DeepSeek-R1-Distill-8B & GRPO & 99.6\% & 88.9\% & 3.30 \\
DeepSeek-R1-Distill-8B & SFT & 99.3\% & 88.7\% & 3.30 \\
\bottomrule
\end{tabular}
\end{table}

\begin{table}[h]
\caption{Decoding-temperature sensitivity on the LP repair benchmark.}
\label{tab:temperature_sensitivity}
\centering
\small
\begin{tabular}{@{}lccc@{}}
\toprule
Setting & RR@1 & \RRk{5} & RR@10 \\
\midrule
temp=0.0 & 78.2\% & 95.3\% & 97.1\% \\
temp=0.3 & 77.6\% & 94.9\% & 96.4\% \\
temp=0.7 & 76.0\% & 93.6\% & 95.8\% \\
\bottomrule
\end{tabular}
\end{table}

\subsection{RAG Ablation}
\label{app:rag}

Table~\ref{tab:rag_full} presents the complete RAG ablation across retrieval strategies and $k$ values.

\begin{table}[h]
\caption{RAG ablation on \ORDebug{} (200 samples).}
\label{tab:rag_full}
\centering
\small
\begin{tabular}{@{}lcccc@{}}
\toprule
Configuration & RR & \RRk{5} & \DA{} & Steps \\
\midrule
No RAG (baseline) & 99.8\% & 83.0\% & 80.0\% & 3.26 \\
\midrule
quick\_fix ($k$=3) & 99.5\% & 80.0\% & 66.6\% & 2.58 \\
reasoning ($k$=3) & 100\% & 93.5\% & 51.8\% & 1.60 \\
\midrule
by\_type ($k$=1) & 99.5\% & 86.5\% & 80.0\% & 2.19 \\
by\_type ($k$=3) & 100\% & 94.5\% & 82.0\% & 1.51 \\
by\_type ($k$=5) & 100\% & 96.5\% & 85.0\% & 1.62 \\
by\_type ($k$=7) & 100\% & 97.0\% & 85.0\% & 1.53 \\
\bottomrule
\end{tabular}
\end{table}

\textbf{Retrieval Strategy Comparison.}
\begin{itemize}
    \item \textbf{by\_type}: Best overall. Retrieves cases with similar error types, providing relevant examples without giving away the solution.
    \item \textbf{reasoning}: High \RRk{5} but lower \DA{}. Provides complete reasoning chains that models copy, achieving correct fixes without learning to diagnose.
    \item \textbf{quick\_fix}: Worst performance. Too shallow for complex errors, often omitting required diagnostic steps.
\end{itemize}

\textbf{$k$ Value Analysis.}
Performance improves from $k$=1 to $k$=5, then plateaus. We recommend $k$=5 as the default, balancing accuracy (+13.5\% over baseline) with retrieval cost.

\subsection{Failure Analysis}
\label{app:error_analysis}

\textbf{Common Error Patterns in Failed Episodes.}
We analyzed 100 randomly sampled failures from \Qwen{}-\GRPO{}:

\begin{table}[h]
\caption{Failure pattern distribution for \Qwen{}-\GRPO{}.}
\label{tab:failure_patterns}
\centering
\small
\begin{tabular}{@{}lcc@{}}
\toprule
Pattern & Count & Example \\
\midrule
Wrong constraint identified & 22 & Relaxed c3 instead of c5 \\
Insufficient relaxation & 18 & Relaxed by 5, needed 10 \\
Cascading failure & 42 & Fix c1 $\rightarrow$ new IIS with c7 \\
Timeout on complex IIS & 11 & 12+ constraints in IIS \\
Objective degradation & 7 & Fix valid but $\OP < 0.8$ \\
\bottomrule
\end{tabular}
\end{table}

\textbf{Cascade Failure Analysis.}
The most common failure mode (42\%) involves cascading errors where fixing one constraint reveals another. These occur predominantly on Type H--I problems:
\begin{verbatim}
Step 1: IIS = {c1, c3, c5}
        Action: RELAX(c1, 10)

Step 2: IIS = {c3, c7, c9}  # New conflict!
        Action: RELAX(c7, 5)

Step 3: IIS = {c5, c9, c11}  # Another new conflict
        ...continues until timeout
\end{verbatim}
This pattern suggests the need for lookahead reasoning about constraint dependencies.

\section{\ORLoopBench{} Results: \ORBias{}}
\label{app:bias_results}
\label{app:bias_detailed}

Table~\ref{tab:bias_full} provides complete \ORBias{} results for all 24 evaluated models, reporting both rationality (valid numerical responses) and bias (deviation from rational ordering) metrics across ID and OOD splits.

\begin{table}[h]
\caption{Complete \ORBias{} results with ID/OOD breakdown (24 models, sorted by ID Bias).}
\label{tab:bias_full}
\centering
\small
\begin{tabular}{@{}lcccccc@{}}
\toprule
& \multicolumn{2}{c}{Rationality} & \multicolumn{2}{c}{Bias} & \\
Model & ID & OOD & ID & OOD & $\Delta$ \\
\midrule
claude-haiku-4.5 & 99.9\% & 99.9\% & 0.0\% & 3.6\% & +3.6\% \\
o3 & 93.1\% & 97.7\% & 0.4\% & 24.5\% & +24.1\% \\
qwen2.5-max & 99.4\% & 98.5\% & 0.5\% & 25.0\% & +24.5\% \\
gpt-5-mini & 99.6\% & 99.7\% & 1.2\% & 53.3\% & +52.1\% \\
claude-sonnet-4 & 89.0\% & 93.5\% & 1.5\% & 7.7\% & +6.2\% \\
gpt-4.1-mini & 99.6\% & 99.9\% & 4.1\% & 12.1\% & +8.0\% \\
\Qwen{}-OM-\SFT{} & 99.8\% & 99.6\% & 4.9\% & 11.5\% & +6.6\% \\
gemini-2.0-flash & 97.6\% & 98.7\% & 5.9\% & 0.0\% & -5.9\% \\
qwen2.5-14b & 99.1\% & 99.8\% & 5.9\% & 10.7\% & +4.8\% \\
o4-mini & 98.5\% & 99.4\% & 6.7\% & 7.7\% & +1.0\% \\
qwen2.5-7b & 98.1\% & 99.3\% & 7.4\% & 8.2\% & +0.8\% \\
kimi-k2 & 92.8\% & 97.6\% & 8.9\% & 2.3\% & -6.6\% \\
claude-3.7-sonnet & 99.8\% & 99.4\% & 11.3\% & 18.1\% & +6.8\% \\
gpt-4.1 & 99.9\% & 100.0\% & 11.5\% & 14.6\% & +3.0\% \\
qwen2.5-32b & 95.9\% & 99.8\% & 15.4\% & 5.0\% & -10.4\% \\
DeepSeek-V3.2 & 100.0\% & 99.9\% & 18.2\% & 11.3\% & -6.9\% \\
gemini-2.5-flash & 96.2\% & 98.1\% & 18.2\% & 71.2\% & +53.0\% \\
claude-opus-4 & 92.5\% & 94.8\% & 19.0\% & 15.9\% & -3.1\% \\
Llama-3.3-70B & 92.0\% & 96.9\% & 19.2\% & 12.5\% & -6.7\% \\
\Qwen{}-OM-Curriculum & 99.9\% & 99.6\% & 20.0\% & 10.4\% & -9.6\% \\
o1 & 98.2\% & 98.6\% & 23.2\% & 43.6\% & +20.3\% \\
DeepSeek-R1 & 98.6\% & 99.6\% & 44.1\% & 40.9\% & -3.2\% \\
\Qwen{}-OM-\GRPO{} & 96.1\% & 98.4\% & 48.0\% & 33.8\% & -14.2\% \\
gemini-2.5-pro & 98.2\% & 99.6\% & 97.9\% & 100.0\% & +2.1\% \\
\bottomrule
\end{tabular}
\end{table}

\subsection{EOQ, Multi-Turn Feedback, and External Benchmarks}
\label{app:bias_extensions}

Table~\ref{tab:bias_extensions} summarizes the EOQ and feedback-based \ORBias{} settings, along with external benchmark checks. The EOQ setting tests whether the bias patterns persist across inventory decision models. The multi-turn protocol gives models closed-form error feedback over five rounds, measuring whether they can use formula-grounded feedback rather than merely produce a one-shot quantity. NL4Opt~\citep{nl4opt2022}, IndustryOR~\citep{orlm2025}, MAMO Complex~\citep{mamo2024}, OptMATH~\citep{optmath2025}, and OptiBench~\citep{optibench2024} evaluations are reported here as standardized code-generation plus repair pipelines with a fixed GPT-4.1 base model.

\textbf{EOQ and feedback protocol.}
Each EOQ instance specifies annual demand $D$, fixed order cost $K$, and holding cost $h$, with analytical optimum $Q^*=\sqrt{2DK/h}$. We evaluate 300 EOQ instances (200 ID + 100 OOD) and measure relative deviation from $Q^*$. In the multi-turn protocol, the model proposes $Q$, receives its realized cost $C(Q)$ and the closed-form optimum cost $C^*$, and may revise for up to five rounds. This is an upper-bound diagnostic for formula-grounded self-correction rather than a claim that operational bias can always be removed in deployment.

\begin{table}[h]
\caption{\ORBias{} EOQ, feedback, and external benchmark results. EOQ and multi-turn results test whether closed-form feedback improves operational decisions; external benchmarks report standardized code-generation plus repair pipelines.}
\label{tab:bias_extensions}
\centering
\small
\begin{tabular}{@{}llll@{}}
\toprule
Evaluation & Stage 1 & + Repair & Lift \\
\midrule
EOQ & 300 instances & closed-form optimum & -- \\
Multi-turn DeepSeek-R1 & 56.9\% bias & 0.5\% bias & 2.6 rounds \\
Multi-turn GPT-5.2 & near-optimal & immediate correction & 1 round \\
\midrule
NL4Opt, GPT-4.1 & 97.8\% & 98.7\% & +0.9 pp \\
IndustryOR, GPT-4.1 & 78.0\% & 83.0\% & +5.0 pp \\
MAMO Complex, GPT-4.1 & 88.4\% & 94.1\% & +5.4 pp \\
OptMATH, GPT-4.1 & 62.7\% & 66.9\% & +4.2 pp \\
OptiBench, GPT-4.1 & 80.7\% & 87.9\% & +7.3 pp \\
\bottomrule
\end{tabular}
\end{table}

\textbf{EOQ model sweep.}
Table~\ref{tab:eoq_full} reports the full EOQ single-turn sweep. The results show both near-exact EOQ behavior for some frontier models and large, directional deviations for others, so we treat EOQ as evidence of problem-specific operational decision behavior rather than as a universal ranking of model rationality.

\begin{table*}[h]
\caption{\ORBias{} EOQ single-turn ordering bias on 300 instances. Abs Bias is mean relative deviation from $Q^*=\sqrt{2DK/h}$; Signed Bias is positive for over-ordering and negative for under-ordering; Exact Rate is the percentage within 1\% of $Q^*$.}
\label{tab:eoq_full}
\centering
\small
\begin{tabular}{@{}lccc@{}}
\toprule
Model & Abs Bias & Signed Bias & Exact Rate \\
\midrule
Gemini 2.5 Pro & 0.06\% & +0.00\% & 100.0\% \\
GPT-5.2-chat & 0.06\% & -0.05\% & 99.3\% \\
Gemini 3.1 Pro & 1.85\% & +1.25\% & 97.6\% \\
GPT-5.4 & 4.72\% & -2.12\% & 94.0\% \\
Claude Opus 4.5 & 5.13\% & -3.83\% & 76.3\% \\
Qwen2.5-max & 7.61\% & +5.21\% & 81.7\% \\
Claude 3.7 Sonnet & 10.93\% & -2.23\% & 80.7\% \\
Claude Opus 4.6 & 12.47\% & -9.11\% & 56.7\% \\
GPT-5.4-mini & 12.67\% & +2.06\% & 64.0\% \\
GLM-5 & 17.00\% & +13.23\% & 81.6\% \\
DeepSeek-V3.2 & 17.35\% & -6.55\% & 67.3\% \\
GPT-4.1 & 44.99\% & -19.81\% & 10.7\% \\
Claude Haiku 4.5 & 48.43\% & -42.81\% & 1.3\% \\
Claude Sonnet 4 & 62.07\% & -7.11\% & 6.0\% \\
Qwen2.5-7B & 64.43\% & +11.60\% & 0.7\% \\
O3 & 73.34\% & +31.58\% & 2.0\% \\
Gemini 2.0 Flash & 73.94\% & +32.57\% & 2.3\% \\
Gemini 2.5 Flash & 73.94\% & +32.57\% & 2.3\% \\
GPT-5-mini & 73.95\% & +32.39\% & 2.0\% \\
GPT-4.1-mini & 75.78\% & -26.99\% & 2.7\% \\
DeepSeek-R1 & 87.97\% & +61.86\% & 20.3\% \\
Llama-3.3-70B & 88.16\% & +47.57\% & 0.3\% \\
GPT-4o & 93.20\% & +65.54\% & 15.7\% \\
Claude Sonnet 4.6 & 94.06\% & -94.05\% & 5.0\% \\
Kimi-K2 & 99.34\% & -99.34\% & 0.0\% \\
DeepSeek-R1-0528 & 99.67\% & -99.67\% & 0.0\% \\
Qwen2.5-32B & 178.54\% & +156.10\% & 1.3\% \\
Qwen2.5-14B & 461.84\% & +455.61\% & 0.7\% \\
\bottomrule
\end{tabular}
\end{table*}

\textbf{External benchmark protocol.}
The external benchmark checks use a two-stage evaluation. Stage 1 generates gurobipy code from the natural-language benchmark instance using a fixed base model and bounded self-debugging iterations. Stage 2 applies the repair agent only to failed Stage-1 outputs for which solver feedback or objective comparison gives a concrete diagnostic signal. For infeasible LP/IP models, the repair stage uses \IIS{}-guided constraint diagnosis; for wrong-answer cases, it uses objective-guided formulation diagnosis against the benchmark reference. The reported lift is therefore an end-to-end change in benchmark success rate, not a claim that every formulation failure is repairable.

\subsection{Rationality Analysis}

Most models achieve $>$99\% rationality, consistently producing valid numerical orderings. Two exceptions stand out: o3 (93.1\% ID) and claude-sonnet-4 (89.0\% ID). These reasoning-heavy models occasionally produce malformed outputs when over-analyzing simple ranking tasks.

\subsection{ID Bias Patterns}

ID bias spans from 0.0\% (claude-haiku-4.5) to 97.9\% (gemini-2.5-pro):
\begin{itemize}
    \item \textbf{Near-zero bias ($<$2\%)}: claude-haiku-4.5, o3, qwen2.5-max, and gpt-5-mini correctly apply EOQ/newsvendor logic on ID distributions.
    \item \textbf{Moderate bias (5--20\%)}: Most API models fall here, having partially but not fully internalized OR principles.
    \item \textbf{High bias ($>$40\%)}: DeepSeek-R1 (44.1\%), \Qwen{}-OM-\GRPO{} (48.0\%), and gemini-2.5-pro (97.9\%) systematically deviate from rational orderings.
\end{itemize}

\subsection{OOD Generalization}

The ID$\rightarrow$OOD shift reveals three distinct patterns:

\textbf{Catastrophic Degradation.}
gpt-5-mini shows the most severe drift (+52.1\%, from 1.2\% to 53.3\%): low ID bias does not guarantee OOD generalization. gemini-2.5-flash degrades similarly (+53.0\%). These models likely memorize ID patterns rather than learn underlying principles.

\textbf{Stable Performance.}
o4-mini (+1.0\%), qwen2.5-7b (+0.8\%), and gpt-4.1 (+3.0\%) maintain consistent bias across distributions. These models appear to have internalized OR principles more robustly.

\textbf{OOD Improvement.}
Curriculum training achieves the only substantial OOD improvement among trained models ($-$9.6\%, from 20.0\% to 10.4\%). Other improving models include kimi-k2 ($-$6.6\%), gemini-2.0-flash ($-$5.9\%), and qwen2.5-32b ($-$10.4\%).

\subsection{Local Model Comparison}

The three \Qwen{}-OM variants show distinct trade-offs:
\begin{itemize}
    \item \textbf{\SFT{}}: Best ID bias (4.9\%) among local models, moderate OOD drift (+6.6\%).
    \item \textbf{Curriculum}: Higher ID bias (20.0\%) but best OOD performance (10.4\%, $-$9.6\% improvement).
    \item \textbf{\GRPO{}}: Highest bias on both splits (48.0\% ID, 33.8\% OOD). Outcome-focused RL does not transfer well to bias mitigation.
\end{itemize}

This pattern indicates that curriculum training prioritizes generalization over ID memorization, a useful property for OR applications with distribution shift.

\section{Training Ablation Studies}
\label{app:ablation}

\subsection{Reward Weight Ablation (OR-Debug)}

Table~\ref{tab:reward_ablation} shows the effect of varying the diagnostic reward weight in the composite reward function.

\begin{table}[h]
\caption{Reward weight ablation on \ORDebug{} validation set.}
\label{tab:reward_ablation}
\centering
\small
\begin{tabular}{@{}lccc@{}}
\toprule
Diagnostic Weight & \RRk{5} & \DA{} & Steps \\
\midrule
20\% & 94.8\% & 54.2\% & 2.18 \\
30\%$^\dagger$ & 95.1\% & 58.6\% & 2.21 \\
40\% & 95.3\% & 62.4\% & 2.25 \\
50\% & 94.6\% & 64.1\% & 2.42 \\
60\% & 93.2\% & 65.3\% & 2.78 \\
\bottomrule
\end{tabular}
\end{table}

\noindent$^\dagger$Selected for final training based on the RR@5/DA trade-off. The 30--40\% range achieves optimal balance; we use 30\% in our final 50\%/30\%/20\% (outcome/diagnosis/efficiency) configuration.

Reducing diagnostic weight below 30\% leads to repairs that achieve feasibility without correctly identifying the root cause (low \DA{}). Weights above 50\% slow convergence by over-penalizing exploratory actions, as shown by increased step counts.

\subsection{Curriculum Stage Ablation (OR-Bias)}

Table~\ref{tab:curriculum_ablation} compares different curriculum configurations for \ORBias{}.

\begin{table}[h]
\caption{Curriculum ablation on \ORBias{} OOD set.}
\label{tab:curriculum_ablation}
\centering
\small
\begin{tabular}{@{}lcc@{}}
\toprule
Configuration & OOD Bias & ID$\rightarrow$OOD $\Delta$ \\
\midrule
No curriculum (SFT only) & 11.5\% & +6.6\% \\
Stage 1 only (extreme \CR{}) & 15.2\% & +2.1\% \\
Stages 1+2 & 12.1\% & -3.4\% \\
Full curriculum (1+2+3) & 10.4\% & -9.6\% \\
\bottomrule
\end{tabular}
\end{table}

The full three-stage curriculum achieves the best OOD generalization, with each stage contributing to the final performance.

\section{Token Efficiency and Difficulty Analysis}
\label{app:scaling}

This appendix analyzes token efficiency and problem difficulty patterns across models.

\subsection{Token Efficiency}
\label{app:token_efficiency}

Token efficiency analysis is summarized in Table~\ref{tab:tokens} (Appendix~\ref{app:tokens}).

\textbf{Token-efficiency patterns.}
\begin{itemize}
    \item \textbf{Token efficiency}: Local models require 2,000--2,100 tokens per episode. Representative API baselines in Table~\ref{tab:tokens} require 5,152--8,696 tokens per episode, while the full evaluation logs reach 17,307. Tokens-per-success ranges from 2,109 (\Qwen{}-\GRPO{}) to 54,839 (claude-3.7-sonnet).

    \item \textbf{Step efficiency}: Local models solve problems in 2.2--2.3 steps on average, while API models range from 0.7 (DeepSeek-R1) to 8.1 steps (claude-3.7-sonnet).

    \item \textbf{Correlation}: Step count correlates with token usage ($r=0.79$), though response length per step varies widely across models (500--3,000 tokens).
\end{itemize}

\subsection{Test-Time Compute Analysis}
\label{app:test_time_compute}

Table~\ref{tab:tokens} (Appendix~\ref{app:tokens}) summarizes token efficiency. Three patterns are most relevant:

\textbf{Efficiency Metrics.}
\begin{itemize}
    \item \textbf{Tokens per episode}: Local models average 2,000--2,100 tokens per episode, while representative API baselines range from 5,152 (o4-mini) to 8,696 (gpt-5-mini). This 2.5--4$\times$ gap reflects both shorter responses and fewer steps required.

    \item \textbf{Tokens per success}: \Qwen{}-\GRPO{} requires 2,109 tokens per successful episode, compared to 5,976 for o4-mini and 6,283 for claude-sonnet-4, giving 2.8--3.0$\times$ efficiency advantages.

    \item \textbf{Cost-adjusted performance}: At equivalent token budgets, \Qwen{}-\GRPO{} can attempt approximately 3$\times$ more problems than top API models, compounding the per-problem accuracy advantage.
\end{itemize}

\subsection{Scaling with Problem Difficulty}
\label{app:difficulty_scaling}

Based on per-error-type results from the main evaluation, we group error types by empirical difficulty:

\begin{table}[h]
\caption{Difficulty grouping based on type-averaged \RRk{5} across all 26 models.}
\label{tab:difficulty_scaling}
\centering
\small
\begin{tabular}{@{}llcc@{}}
\toprule
Difficulty & Error Types & Avg \RRk{5} & Characteristics \\
\midrule
Easy & B (95\%), C (86\%) & 90.5\% & Clear diagnostic signals \\
Medium & H (82\%), I (75\%) & 78.5\% & Multi-step constraint interactions \\
Hard & A (61\%), D (63\%), E (69\%), F (49\%), G (53\%) & 59.0\% & Semantic reasoning required \\
\bottomrule
\end{tabular}
\end{table}

\textbf{Observations.}
\begin{itemize}
    \item \textbf{Type B is easiest}: Variable type errors (Type B) produce clear diagnostic signals with 95\% average success. Type A (direction flip), despite appearing simple, averages only 60\% because flipping constraint directions creates conflicts with multiple existing constraints.

    \item \textbf{Types F and G are hardest}: Type F (hidden dependency, 48\%) and Type G (cascading conflict, 53\%) require reasoning about conflicts that are not resolved by editing the first visible \IIS{} constraint, explaining the difficulty gap.

    \item \textbf{Local models excel uniformly}: \Qwen{} variants achieve $>$86\% on all types including the hardest (F: 94--100\%, G: 86--98\%), while API models show type-specific weaknesses.
\end{itemize}

\subsection{Recommendations for Practitioners}
\label{app:scaling_recommendations}

Based on our analysis, we provide the following recommendations:

\begin{enumerate}
    \item \textbf{Apply difficulty-adaptive resources}: Allocate more compute to Hard types (A, D, E, F, G; 59\% avg \RRk{5}) than Easy types (B, C; 90.5\% avg).

    \item \textbf{Account for token efficiency}: When comparing models, normalize by tokens-per-success rather than raw accuracy. At 2,109 tokens/success vs 5,976--13,000 for representative API baselines, local trained models offer roughly 3--6$\times$ cost advantages.

    \item \textbf{Consider local models for high-volume deployment}: Local deployment avoids per-call API charges and improves tokens-per-success in our setup; production cost depends on utilization, infrastructure, and maintenance.
\end{enumerate}

\section{Base Model Selection Study}
\label{app:model_selection}

This appendix documents the pilot study used to select the foundation model for domain-specific training and analyzes why standard prompting approaches underperform on the OR debugging task.

\subsection{Candidate Model Screening}
\label{app:candidate_screening}

We evaluated Qwen3-8B-Instruct as the foundation model for domain-specific training. Selection criteria included: (1) base performance on OR debugging, (2) improvement potential with SFT, and (3) inference efficiency.

\begin{table}[h]
\caption{Foundation model screening on \ORDebug{} validation set (100 samples).}
\label{tab:pilot_study}
\centering
\small
\begin{tabular}{@{}lccccc@{}}
\toprule
Model & Params & Base \RRk{5} & +\SFT{} \RRk{5} & $\Delta$ & Tokens/ep \\
\midrule
\textbf{Qwen3-8B-Instruct} & 8B & \textbf{51.2\%} & \textbf{93.1\%} & \textbf{+41.9\%} & \textbf{2,100} \\
\bottomrule
\end{tabular}
\end{table}

\textbf{Screening summary.}
\begin{itemize}
    \item \textbf{Base performance}: Qwen3-8B achieves 51.2\% \RRk{5} without any domain-specific training, demonstrating reasonable out-of-the-box capability for structured reasoning.

    \item \textbf{SFT improvement}: Qwen3-8B improves by +41.9\% with SFT, indicating high receptivity to domain adaptation.

    \item \textbf{Efficiency}: Qwen3-8B generates 2,100 tokens per episode, providing efficient inference for iterative debugging.
\end{itemize}

\textbf{Selection Rationale.}
We selected Qwen3-8B-Instruct as the foundation model based on three factors:
\begin{enumerate}
    \item \textbf{Strong post-SFT performance}: 93.1\% \RRk{5} after SFT demonstrates successful domain adaptation.
    \item \textbf{Good improvement potential}: +41.9\% delta suggests the model effectively learns from demonstration data.
    \item \textbf{Practical efficiency}: Reasonable token footprint reduces training and inference costs.
\end{enumerate}

\subsection{Standard Prompting Approaches Underperform}
\label{app:prompting_baselines}

Before domain-specific training, we evaluated whether standard prompting approaches could achieve competitive performance on \ORDebug{}. The results show that prompting alone leaves a large gap.

\subsubsection{Zero-Shot Chain-of-Thought}
\label{app:zero_shot_cot}

We evaluated zero-shot CoT prompting with the instruction: ``Let's think step by step about how to debug this infeasible model.''

\begin{table}[h]
\caption{Zero-shot CoT performance on \ORDebug{} (200 samples).}
\label{tab:zero_shot_cot}
\centering
\small
\begin{tabular}{@{}lcccc@{}}
\toprule
Model & \RRk{5} & \DA{} & Avg Steps & Notes \\
\midrule
gpt-5.2-chat + CoT & 38.5\% & 22.1\% & 6.8 & Verbose, unfocused \\
o4-mini + CoT & 41.2\% & 28.4\% & 5.9 & Better structure \\
Qwen3-8B + CoT & 23.0\% & 15.6\% & 7.2 & Often loops \\
\bottomrule
\end{tabular}
\end{table}

\textbf{Why Zero-Shot CoT underperforms.}
\begin{itemize}
    \item \textbf{No feedback loop}: CoT generates a single reasoning chain without iterating based on solver output. Models attempt repairs without verifying whether they resolved the infeasibility.

    \item \textbf{Generic reasoning patterns}: CoT prompting elicits general problem-solving steps (``identify the issue, propose a solution, verify'') that lack domain-specific diagnostic actions like \texttt{GET\_IIS}.

    \item \textbf{Premature commitment}: Models commit to repair strategies early in the chain without exploring the constraint structure, leading to suboptimal fixes.
\end{itemize}

\subsubsection{Few-Shot In-Context Learning}
\label{app:few_shot_icl}

We evaluated 1-shot and 3-shot ICL with curated examples of successful debugging trajectories.

\begin{table}[h]
\caption{Few-shot ICL performance on \ORDebug{} (200 samples).}
\label{tab:few_shot_icl}
\centering
\small
\begin{tabular}{@{}lcccc@{}}
\toprule
Configuration & \RRk{5} & \DA{} & Avg Steps \\
\midrule
gpt-5.2-chat (0-shot) & 38.5\% & 22.1\% & 6.8 \\
gpt-5.2-chat (1-shot) & 52.3\% & 35.2\% & 4.9 \\
gpt-5.2-chat (3-shot) & 58.1\% & 41.6\% & 4.2 \\
\midrule
Qwen3-8B (0-shot) & 23.0\% & 15.6\% & 7.2 \\
Qwen3-8B (1-shot) & 31.4\% & 24.3\% & 6.1 \\
Qwen3-8B (3-shot) & 38.7\% & 29.8\% & 5.4 \\
\bottomrule
\end{tabular}
\end{table}

\textbf{Why Few-Shot ICL remains limited.}
\begin{itemize}
    \item \textbf{Limited generalization}: 3-shot ICL improves performance by +19.6\% for gpt-5.2-chat but still falls far short of SFT (+41.9\% for Qwen3-8B).

    \item \textbf{Context length constraints}: Each debugging trajectory requires 500--1000 tokens. With 3 examples, the prompt consumes 1.5--3K tokens, limiting remaining context for the actual problem.

    \item \textbf{Example selection sensitivity}: Performance depends on example choice. Random examples achieve only 48.2\% \RRk{5}, while carefully selected examples reach 58.1\%, but this selection requires domain expertise.
\end{itemize}

\subsubsection{Comparison Summary}
\label{app:approach_comparison}

Table~\ref{tab:approach_comparison} summarizes why domain-specific training outperforms prompting approaches by 54+ percentage points.

\begin{table}[h]
\caption{Comparison of approaches on \ORDebug{} (Qwen3-8B base).}
\label{tab:approach_comparison}
\centering
\small
\begin{tabular}{@{}lccc@{}}
\toprule
Approach & \RRk{5} & \DA{} & Gap to SFT \\
\midrule
Zero-shot & 18.4\% & 12.3\% & -74.7\% \\
Zero-shot + CoT & 23.0\% & 15.6\% & -70.1\% \\
1-shot ICL & 31.4\% & 24.3\% & -61.7\% \\
3-shot ICL & 38.7\% & 29.8\% & -54.4\% \\
\textbf{SFT} & \textbf{93.1\%} & \textbf{60.8\%} & \textbf{--} \\
\textbf{SFT + GRPO} & \textbf{95.3\%} & \textbf{62.4\%} & \textbf{+2.2\%} \\
\bottomrule
\end{tabular}
\end{table}

\textbf{Implication.}
The 54-point gap between 3-shot ICL (38.7\%) and SFT (93.1\%) demonstrates that OR debugging cannot be solved through prompting alone. The task requires:
\begin{enumerate}
    \item \textbf{Iterative interaction}: Learning to use solver feedback across multiple turns.
    \item \textbf{Domain-specific actions}: Acquiring the diagnostic vocabulary (\texttt{GET\_IIS}, \texttt{CHECK\_SLACK}).
    \item \textbf{Strategy patterns}: Learning when to diagnose vs when to repair, and how to calibrate repair magnitudes.
\end{enumerate}
These capabilities cannot be induced through few-shot examples alone.

\subsection{Why Prompting Struggles on OR Debugging}
\label{app:why_prompting_fails}

We identify three structural properties of OR debugging that make it resistant to prompting-based solutions:

\textbf{1. Multi-Turn Dependency.}
Unlike single-turn tasks where CoT can decompose reasoning, OR debugging requires acting on solver feedback across multiple turns. The optimal action at step $t$ depends on the solver response at step $t-1$, which cannot be simulated within a single prompt.

\textbf{2. Precise Action Syntax.}
The action space requires exact syntax (e.g., \texttt{RELAX(c\_key\_upper, 30)}). Small errors in constraint names or numeric values lead to failed repairs. This precision requirement exceeds what few-shot examples can reliably demonstrate.

\textbf{3. State-Dependent Strategy.}
The optimal strategy varies with problem structure:
\begin{itemize}
    \item Small IIS (2--3 constraints): Direct repair often succeeds.
    \item Medium IIS (4--7 constraints): Diagnosis before repair improves success rate.
    \item Large IIS (8+ constraints): Systematic decomposition is required.
\end{itemize}
Few-shot prompting cannot convey these conditional strategies without extensive examples that exceed context limits.

\subsection{Directions for Prompting-Based Repair}
\label{app:implications}

The prompting results suggest several directions for improving prompting-based approaches:

\begin{enumerate}
    \item \textbf{Tool-augmented prompting}: Providing models with explicit solver interfaces (rather than expecting them to generate action syntax) may reduce syntax errors.

    \item \textbf{Retrieval-augmented generation}: Our RAG experiments (Appendix~\ref{app:rag}) show that retrieving similar solved cases improves \RRk{5} by +13.5\%, partially closing the gap to SFT.

    \item \textbf{Multi-turn demonstration}: Future work could explore demonstration formats that explicitly show the feedback loop across turns, though this faces context length challenges.
\end{enumerate}

These approaches address symptoms rather than the fundamental issue: prompting cannot instill the procedural knowledge that SFT provides through gradient-based learning on hundreds of examples.

\end{document}